\documentclass{article}
\newif\ifemail
\newif\ifchecklist
\newif\ifappendix

\usepackage[final,nonatbib]{neurips_2021}\emailfalse\checklistfalse\appendixtrue %

\newcommand{\confnotice}{}

\usepackage[utf8]{inputenc}   %
\usepackage[OT2,T1]{fontenc}  %
\usepackage{hyperref}         %
\usepackage{url}              %
\usepackage{booktabs}         %
\usepackage{amsfonts}         %
\usepackage{nicefrac}         %
\usepackage{microtype}        %
\usepackage{graphicx}
\usepackage{amsmath}
\usepackage{amssymb}
\usepackage[noend]{algpseudocode}
\usepackage{algorithmicx}
\usepackage{algorithm}
\usepackage{subcaption}
\usepackage{calc}
\usepackage{tikz}
\usepackage{array}
\usepackage{multirow}
\usepackage{tabu}
\usepackage{bm}
\usepackage{xfrac}

\graphicspath{{./figures/}}

\usepackage{color}
\usepackage[normalem]{ulem}  %
\usepackage[outline]{contour}
\usepackage{afterpage}
\usepackage{pifont}
\usepackage{geometry}
\definecolor{olive}{rgb}{0.5, 0.5, 0.0}
\definecolor{maroon}{rgb}{0.69, 0.19, 0.38}
\definecolor{celestialblue}{rgb}{0.29, 0.59, 0.82}
\definecolor{darkgreen}{rgb}{0.0, 0.5, 0.0}
\definecolor{grey}{rgb}{0.5,0.5,0.5}
\definecolor{darkblue}{rgb}{0.19, 0.19, 0.62}
\definecolor{silver}{rgb}{0.7,0.7,0.7}

\newcommand{\FINAL}[1]{#1}

\def\clap#1{\hbox to 0pt{\hss #1\hss}}%

\newcommand{\integer}{\mathbb{Z}}
\newcommand{\low}[1]{\raisebox{0pt}[0pt][0pt]{#1}}
\newcommand{\lowsqrt}[1]{\low{$\sqrt{#1}$}}

\newcommand{\Synthesis}{\mathbf{G}}
\newcommand{\synthesis}{\mathbf{g}}
\newcommand{\layer}{\mathbf{f}}
\newcommand{\Layer}{\mathbf{F}}
\newcommand{\layern}[1]{\mathbf{f}_{\!\:\text{#1}}}			%
\newcommand{\Layern}[1]{\mathbf{F}_{\!\!\:\text{#1}}}
\newcommand{\layernm}[1]{\mathbf{f}_{\!\:\mathrm{#1}}}		%
\newcommand{\Layernm}[1]{\mathbf{F}_{\!\!\:\mathrm{#1}}}
\newcommand{\feat}{z}
\newcommand{\Feat}{Z}

\newcommand{\Kernel}{K}

\newcommand{\sfreq}{s}   %
\newcommand{\freq}{f}
\newcommand{\freqc}{\freq_c} %
\newcommand{\freqh}{\freq_h} %
\newcommand{\freqt}{\freq_t} %
\newcommand{\freqtzero}{\freq_{t,0}} %

\newcommand{\xx}{\boldsymbol{x}}

\newcommand{\conv}{\ast}
\newcommand{\mult}{\odot}

\DeclareSymbolFont{cyrletters}{OT2}{wncyr}{m}{n}
\DeclareMathSymbol{\Sha}{\mathalpha}{cyrletters}{"58}
\newcommand{\comb}{\Sha}
\newcommand{\combs}{\comb_{\sfreq\!\!\!\:\:}}
\newcommand{\combsp}{\comb_{\sfreq'\!\!\:}}

\newcommand{\rawnonlin}{\sigma}

\newcommand{\Trans}{\mathbf{T}}
\newcommand{\trans}{\mathbf{t}}
\newcommand{\Rot}{\mathbf{R}}
\newcommand{\rot}{\mathbf{r}}

\newcommand{\idealsq}{\phi}
\newcommand{\idealsqs}{\idealsq_\sfreq}
\newcommand{\idealsqsp}{\idealsq_{\sfreq'}}
\newcommand{\ideallow}{\psi}
\newcommand{\idealdiscs}{\idealsq_\sfreq^\circ}

\newcommand{\ww}{{\bf w}}

\newcommand{\WW}{\mathcal{W}}

\DeclareMathOperator{\sinc}{sinc}
\DeclareMathOperator{\jinc}{jinc}
\DeclareMathOperator{\round}{round}

\newcommand{\metrict}{\text{EQ-T}}
\newcommand{\metricr}{\text{EQ-R}}
\newcommand{\expectation}{\mathbb{E}}

\newcommand{\h}{0mm}
\newcommand{\hh}{0mm}
\newcommand{\hhh}{0mm}

\newcommand{\vv}{0mm}

\newcommand{\s}{\hphantom{0}}
\newcommand{\cs}{\hspace{0}}

\newcommand{\projectpage}{{\small\url{https://nvlabs.github.io/stylegan3}}}
\newcommand{\codepage}{{\small\url{https://github.com/NVlabs/stylegan3}}}

\title{Alias-Free Generative Adversarial Networks}

\author{%
	Tero Karras\\NVIDIA\\
	\ifemail\texttt{tkarras@nvidia.com}\else\fi
	\And
	Miika Aittala\\NVIDIA\\
	\ifemail\texttt{maittala@nvidia.com}\else\fi
	\And
	Samuli Laine\\NVIDIA\\
	\ifemail\texttt{slaine@nvidia.com}\else\fi
	\AND
	Erik H\"ark\"onen\thanks{This work was done during an internship at NVIDIA.}\\Aalto University and NVIDIA\\
	\ifemail\texttt{erik.harkonen@aalto.fi}\else\fi
	\And
	Janne Hellsten\\NVIDIA\\
	\ifemail\texttt{jhellsten@nvidia.com}\else\fi
	\AND
	Jaakko Lehtinen\\NVIDIA and Aalto University\\
	\ifemail\texttt{jlehtinen@nvidia.com}\else\fi
	\And
	Timo Aila\\NVIDIA\\
	\ifemail\texttt{taila@nvidia.com}\else\fi
}

\newcommand{\figIntro}[1]{
\renewcommand{\h}{0.225\linewidth}%
\renewcommand{\hh}{0.45\linewidth}%
\renewcommand{\hhh}{0.11\linewidth}%
\renewcommand{\vv}{31mm}%
\begin{figure}[t]
\rotatebox{90}{\phantom{Bl}}\hfill%
\makebox[\h][c]{\small StyleGAN2}\hfill
\makebox[\h][c]{\small \FINAL{Ours}}\hfill%
\hspace{8mm}%
\parbox[c][5mm][c]{\h}{\centering \small StyleGAN2\\\tiny $\leftarrow$ latent interpolation $\rightarrow$}\hfill
\parbox[c][5mm][c]{\h}{\centering \small \FINAL{Ours}\\\tiny $\leftarrow$ latent interpolation $\rightarrow$}\\%
\rotatebox{90}{\small\makebox[\hhh]{Averaged}\hfill\makebox[\hhh]{Central}}
\includegraphics[width=\h]{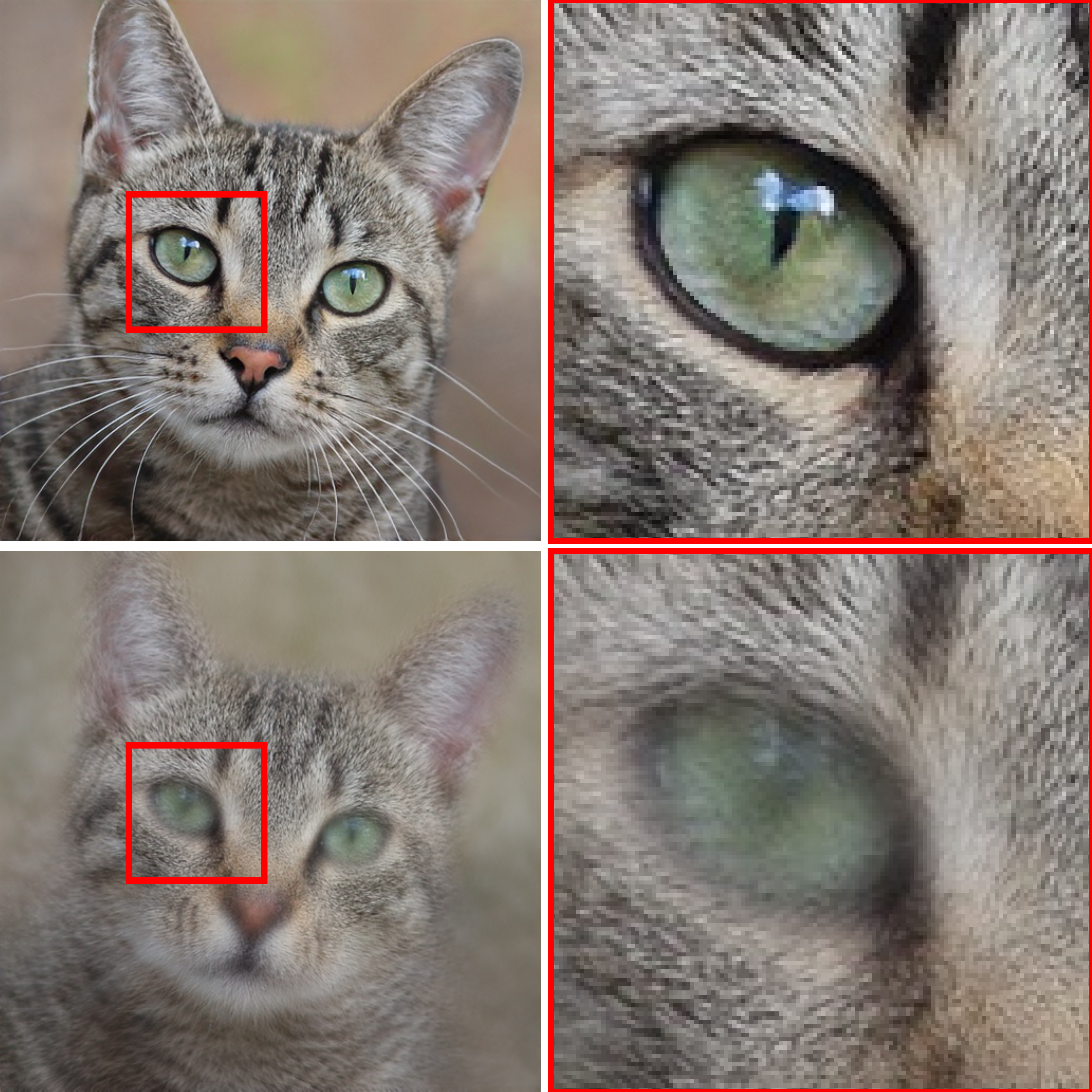}\hfill
\includegraphics[width=\h]{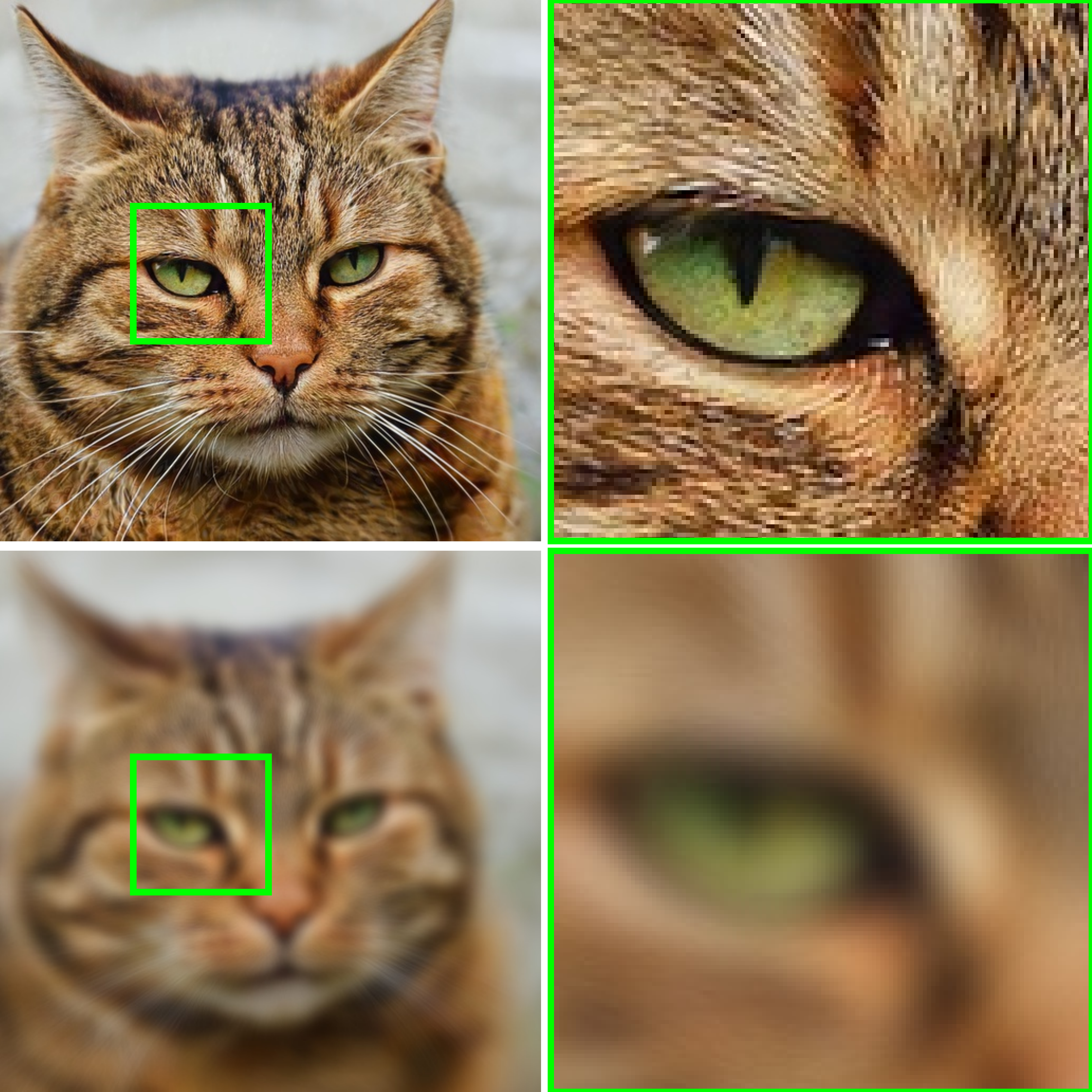}%
\hspace{4mm}%
\begin{tikzpicture}\draw (0,0) -- (0,\vv);\end{tikzpicture}%
\hspace{4mm}%
\includegraphics[width=\h]{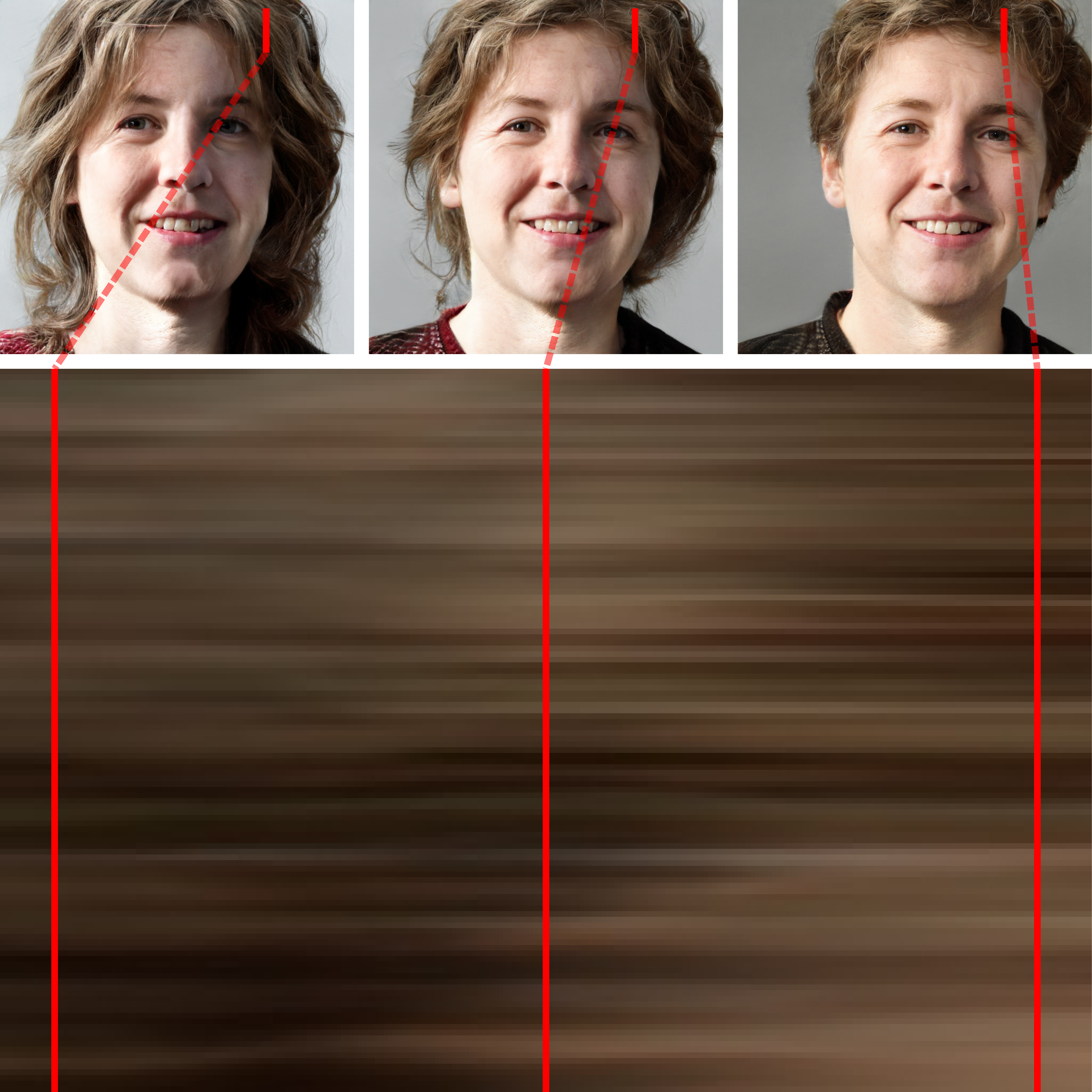}\hfill
\includegraphics[width=\h]{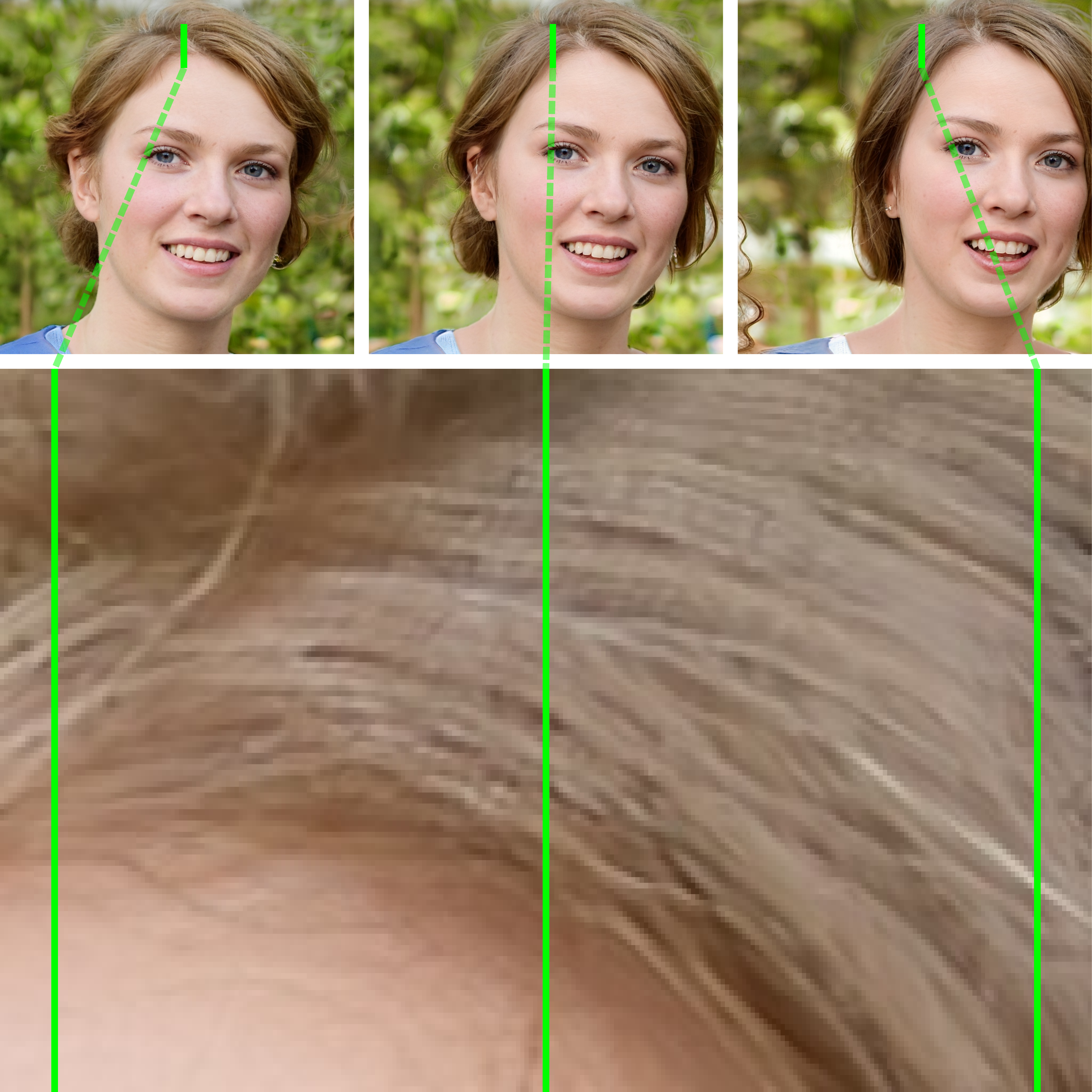}%
\caption{
Examples of ``texture sticking''.
\textbf{Left:} The average of images generated from a small neighborhood around a central latent (top row). 
The intended result is uniformly blurry because all details should move together.
However, with StyleGAN2 many details (e.g.,~fur) stick to the same pixel coordinates, showing unwanted sharpness.
\textbf{Right:} 
From a latent space interpolation (top row), we extract a short vertical segment of pixels from each generated image and stack them horizontally (bottom).
The desired result is hairs moving in animation, creating a time-varying field.
With StyleGAN2 the hairs mostly stick to the same coordinates, creating horizontal streaks instead.
}
\label{#1}
\end{figure}
}

\newcommand{\figTheory}[1]{
\renewcommand{\h}{0.55\linewidth}%
\renewcommand{\hh}{0.44\linewidth}%
\renewcommand{\vv}{32mm}%
\begin{figure}[t]
\includegraphics[width=\linewidth]{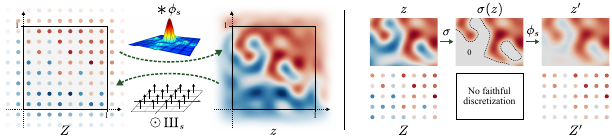}
\caption{
\textbf{Left:}
Discrete representation $\Feat$ and continuous representation $\feat$ are related to each other via convolution with ideal interpolation filter $\idealsqs$ and pointwise multiplication with Dirac comb $\combs$.
\textbf{Right:}
Nonlinearity $\rawnonlin$, ReLU in this example, may produce arbitrarily high frequencies in the continuous-domain $\rawnonlin(\feat)$. Low-pass filtering via $\idealsqs$ is necessary to ensure that $\Feat'$ captures the result.
}
\label{#1}
\end{figure}
}

\newcommand{\genBridgeUnalignedFfhqSmall}{%
         & {\bf Configuration}                            & FID\,$\downarrow$  & EQ-T\,$\uparrow$  & EQ-R\,$\uparrow$  \\
\tabucline{-}
{\sc a}  & StyleGAN2                                      & 5.14               & --                & --                \\
{\sc b}  & + Fourier features                             & 4.79               & 16.23             & 10.81             \\
{\sc c}  & + No noise inputs                              & 4.54               & 15.81             & 10.84             \\
{\sc d}  & + Simplified generator                         & 5.21               & 19.47             & 10.41             \\
\tabucline{-}
{\sc e}  & + Boundaries \& upsampling                     & 6.02               & 24.62             & 10.97             \\
{\sc f}  & + Filtered nonlinearities                      & 6.35               & 30.60             & 10.81             \\
{\sc g}  & + Non-critical sampling                        & 4.78               & 43.90             & 10.84             \\
{\sc h}  & + Transformed Fourier features                 & 4.64               & 45.20             & 10.61             \\
\tabucline{-}
{\sc t}  & + Flexible layers \hfill(\FINAL{StyleGAN3-T})  & 4.62               & 63.01             & 13.12             \\
{\sc r}  & + Rotation equiv. \hfill(\FINAL{StyleGAN3-R})  & {\bf 4.50}         & {\bf 66.65}       & {\bf 40.48}       \\
}

\newcommand{\genSweepsCfgR}{%
         & {\bf Parameter}                  & FID\,$\downarrow$  & EQ-T\,$\uparrow$  & EQ-R\,$\uparrow$  & Time                & Mem.                \\
\tabucline{-}
{\sc }   & Filter size $n = 4$              & 4.72               & 57.49             & 39.70             & {\bf 0.84$\times$}  & {\bf 0.99$\times$}  \\
{\sc *}  & Filter size $n = 6$              & {\bf 4.50}         & {\bf 66.65}       & 40.48             & 1.00$\times$        & 1.00$\times$        \\
{\sc }   & Filter size $n = 8$              & 4.66               & 65.57             & {\bf 42.09}       & 1.18$\times$        & 1.01$\times$        \\
\tabucline{-}
{\sc }   & Upsampling $m = 1$               & {\bf 4.38}         & 39.96             & 36.42             & {\bf 0.65$\times$}  & {\bf 0.87$\times$}  \\
{\sc *}  & Upsampling $m = 2$               & 4.50               & 66.65             & 40.48             & 1.00$\times$        & 1.00$\times$        \\
{\sc }   & Upsampling $m = 4$               & 4.57               & {\bf 74.21}       & {\bf 40.97}       & 2.31$\times$        & 1.62$\times$        \\
\tabucline{-}
{\sc }   & Stopband $\freqtzero = 2^{1.5}$  & 4.62               & 51.10             & 29.14             & {\bf 0.86$\times$}  & {\bf 0.90$\times$}  \\
{\sc *}  & Stopband $\freqtzero = 2^{2.1}$  & {\bf 4.50}         & 66.65             & 40.48             & 1.00$\times$        & 1.00$\times$        \\
{\sc }   & Stopband $\freqtzero = 2^{3.1}$  & 4.68               & {\bf 73.13}       & {\bf 41.63}       & 1.36$\times$        & 1.25$\times$        \\
}

\newcommand{\tabBridgeUnalignedFfhqSmall}{%
\newcolumntype{x}{>{\centering\arraybackslash\hspace{0pt}}p{7.3mm}}%
\newcolumntype{y}{>{\centering\arraybackslash\hspace{0pt}}p{9.6mm}}%
\renewcommand{\cs}{\hspace{2.5mm}}%
\tabulinesep=0.50mm%
\tabulinestyle{0.17mm}%
\begin{tabu}{|c@{\hspace{2.4mm}}l|x@{\cs}y@{\cs}y|}
\tabucline{-}
\genBridgeUnalignedFfhqSmall
\tabucline{-}
\end{tabu}%
}

\newcommand{\tabSweeps}{%
\newcolumntype{x}{>{\centering\arraybackslash\hspace{0pt}}p{7.3mm}}%
\newcolumntype{y}{>{\centering\arraybackslash\hspace{0pt}}p{9.6mm}}%
\renewcommand{\cs}{\hspace{2.5mm}}%
\tabulinesep=0.605mm%
\tabulinestyle{0.17mm}%
\begin{tabu}{|c@{\hspace{1.4mm}}l|x@{\cs}y@{\cs}y|y@{\hspace{2mm}}y|}
\tabucline{-}
\genSweepsCfgR
\tabucline{-}
\end{tabu}%
}

\newcommand{\figBridgeTables}[1]{
\begin{figure}[t]
\centering%
\footnotesize%
\scalebox{0.75}{\tabBridgeUnalignedFfhqSmall}\hfill%
\scalebox{0.75}{\tabSweeps}%
\caption{
Results for FFHQ-U (unaligned FFHQ) at $256^2$.
\textbf{Left:} Training configurations.
FID is computed between 50k generated images and all training images~\cite{Heusel2017,Karras2020}; lower is better.
\metrict{} and \metricr{} are our equivariance metrics in decibels (dB); higher is better.
\textbf{Right:} Parameter ablations using our final configuration (\textsc{r}) for the filter's support, magnification around nonlinearities, and the minimum stopband frequency at the first layer.
* indicates our default choices.
}
\label{#1}
\end{figure}
}

\newcommand{\figPractice}[1]{
\renewcommand{\h}{1.0\linewidth}%
\begin{figure}[t]
\includegraphics[width=\linewidth]{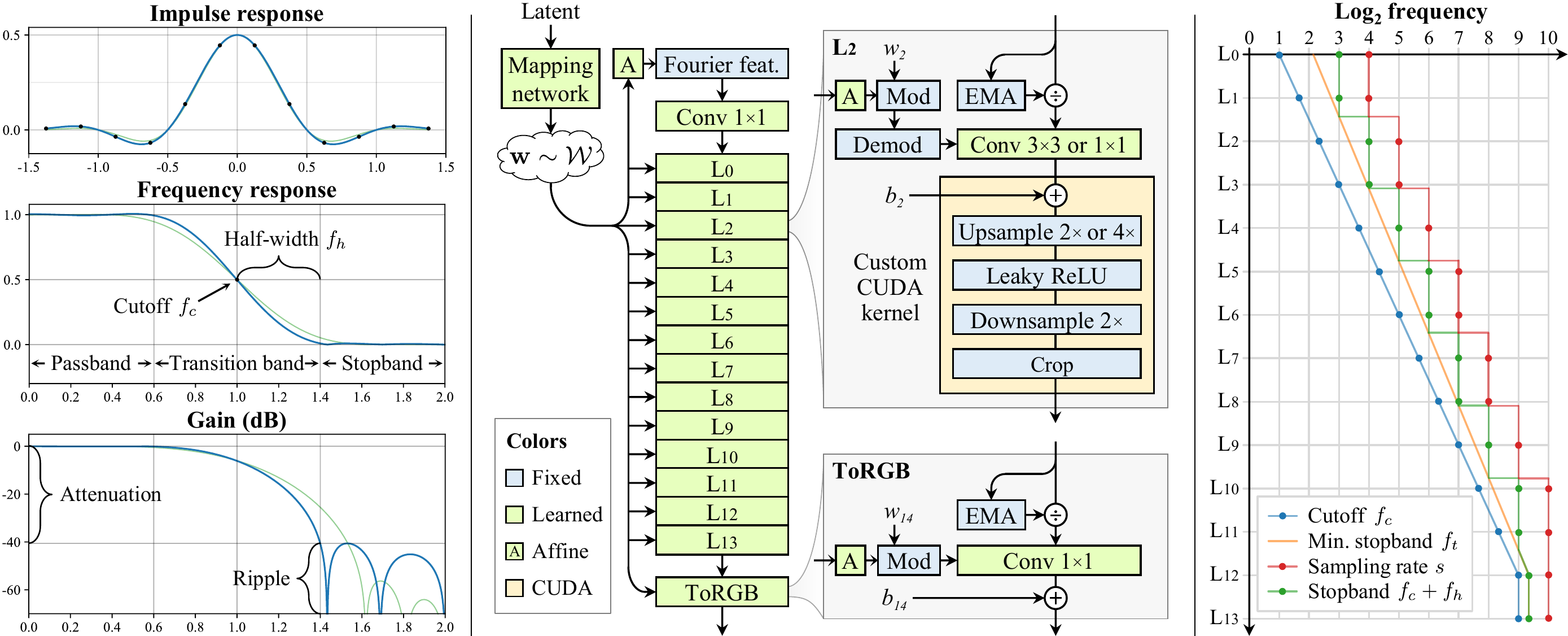}\\
\footnotesize%
\hspace*{2.5mm}%
\makebox[37.2mm]{(a) Filter design concepts}%
\hspace*{4.4mm}%
\makebox[60.0mm]{(b) Our alias-free \FINAL{StyleGAN3} generator architecture}%
\hfill%
\makebox[27mm]{(c) Flexible layers}%
\hspace*{1.5mm}%
\caption{
\textbf{(a)} 1D example of a 2$\times$ upsampling filter with $n=6$, $\sfreq=2$, $\freqc=1$, and $\freqh=0.4$ (blue). Setting $\freqh=0.6$ makes the transition band wider (green), which reduces the unwanted stopband ripple and thus leads to stronger attenuation.
\textbf{(b)} Our alias-free generator, corresponding to configs \textsc{t} and \textsc{r} in Figure~\ref{fig:BridgeTables}. The main datapath consists of Fourier features and normalization (Section~\ref{sec:practice_removals}), modulated convolutions~\cite{Karras2019}, and filtered nonlinearities (Section~\ref{sec:practice_additions}).
\textbf{(c)} Flexible layer specifications (config~\textsc{t}) with $N = 14$ and $\sfreq_N = 1024$. Cutoff $\freqc$ (blue) and minimum acceptable stopband frequency $\freqt$ (orange) obey geometric progression over the layers; sampling rate $\sfreq$ (red) and actual stopband $\freqc + \freqh$ (green) are computed according to our design constraints.
}
\label{#1}
\end{figure}
}

\newcommand{\genDatasets}{%
{\bf Dataset}                                                                                                                                             & {\bf Config}                & FID\,$\downarrow$  & EQ-T\,$\uparrow$  & EQ-R\,$\uparrow$  \\
\tabucline{-}
\multirow{3}{*}{\begin{tabular}{@{}l@{}}{\sc FFHQ-U}\\[-0.4mm]{\scriptsize 70000 img, 1024$^2$}\\[-0.4mm]{\scriptsize Train from scratch}\end{tabular}}   & StyleGAN2                   & \s3.79             & 15.89             & 10.79             \\
                                                                                                                                                          & \FINAL{StyleGAN3-T (ours)}  & \s3.67             & 61.69             & 13.95             \\
                                                                                                                                                          & \FINAL{StyleGAN3-R (ours)}  & \s{\bf 3.66}       & {\bf 64.78}       & {\bf 47.64}       \\
\tabucline{-}
\multirow{3}{*}{\begin{tabular}{@{}l@{}}{\sc FFHQ}\\[-0.4mm]{\scriptsize 70000 img, 1024$^2$}\\[-0.4mm]{\scriptsize Train from scratch}\end{tabular}}     & StyleGAN2                   & \s{\bf 2.70}       & 13.58             & 10.22             \\
                                                                                                                                                          & \FINAL{StyleGAN3-T (ours)}  & \s2.79             & 61.21             & 13.82             \\
                                                                                                                                                          & \FINAL{StyleGAN3-R (ours)}  & \s3.07             & {\bf 64.76}       & {\bf 46.62}       \\
\tabucline{-}
\multirow{3}{*}{\begin{tabular}{@{}l@{}}{\sc MetFaces-U}\\[-0.4mm]{\scriptsize 1336 img, 1024$^2$}\\[-0.4mm]{\scriptsize ADA, from FFHQ-U}\end{tabular}}  & StyleGAN2                   & 18.98              & 18.77             & 13.19             \\
                                                                                                                                                          & \FINAL{StyleGAN3-T (ours)}  & {\bf 18.75}        & 64.11             & 16.63             \\
                                                                                                                                                          & \FINAL{StyleGAN3-R (ours)}  & {\bf 18.75}        & {\bf 66.34}       & {\bf 48.57}       \\
\tabucline{-}
\multirow{3}{*}{\begin{tabular}{@{}l@{}}{\sc MetFaces}\\[-0.4mm]{\scriptsize 1336 img, 1024$^2$}\\[-0.4mm]{\scriptsize ADA, from FFHQ}\end{tabular}}      & StyleGAN2                   & 15.22              & 16.39             & 12.89             \\
                                                                                                                                                          & \FINAL{StyleGAN3-T (ours)}  & {\bf 15.11}        & {\bf 65.23}       & 16.82             \\
                                                                                                                                                          & \FINAL{StyleGAN3-R (ours)}  & 15.33              & 64.86             & {\bf 46.81}       \\
\tabucline{-}
\multirow{3}{*}{\begin{tabular}{@{}l@{}}{\sc AFHQv2}\\[-0.4mm]{\scriptsize 15803 img, 512$^2$}\\[-0.4mm]{\scriptsize ADA, from scratch}\end{tabular}}     & StyleGAN2                   & \s4.62             & 13.83             & 11.50             \\
                                                                                                                                                          & \FINAL{StyleGAN3-T (ours)}  & \s{\bf 4.04}       & 60.15             & 13.51             \\
                                                                                                                                                          & \FINAL{StyleGAN3-R (ours)}  & \s4.40             & {\bf 64.89}       & {\bf 40.34}       \\
\tabucline{-}
\multirow{3}{*}{\begin{tabular}{@{}l@{}}{\sc Beaches}\\[-0.4mm]{\scriptsize 20155 img, 512$^2$}\\[-0.4mm]{\scriptsize ADA, from scratch}\end{tabular}}    & StyleGAN2                   & \s5.03             & 15.73             & 12.69             \\
                                                                                                                                                          & \FINAL{StyleGAN3-T (ours)}  & \s{\bf 4.32}       & 59.33             & 15.88             \\
                                                                                                                                                          & \FINAL{StyleGAN3-R (ours)}  & \s4.57             & {\bf 63.66}       & {\bf 37.42}       \\
}

\newcommand{\genAblations}{%
   & \multirow{2}{*}{\bf Ablation}    & \multicolumn{2}{c|}{\bf Translation eq.}  & \multicolumn{3}{c|}{\bf + Rotation eq.}                    \\
&                                     & FID\,$\downarrow$  & EQ-T\,$\uparrow$     & FID\,$\downarrow$  & EQ-T\,$\uparrow$  & EQ-R\,$\uparrow$  \\
\tabucline{-}
*  & Main configuration               & 4.62               & 63.01                & {\bf 4.50}         & 66.65             & 40.48             \\
   & With mixing reg.                 & {\bf 4.60}         & 63.48                & 4.67               & 63.59             & 40.90             \\
   & With noise inputs                & 4.96               & 24.46                & 5.79               & 26.71             & 26.80             \\
   & Without flexible layers          & 4.64               & 45.20                & 4.65               & 44.74             & 22.52             \\
   & Fixed Fourier features           & 5.93               & 64.57                & 6.48               & 66.20             & 41.77             \\
   & With path length reg.            & 5.00               & {\bf 68.36}          & 5.98               & {\bf 71.64}       & {\bf 42.18}       \\
\tabucline{-}
   & 0.5$\times$ capacity             & 7.43               & 63.14                & 6.52               & 63.08             & 39.89             \\
*  & 1.0$\times$ capacity             & 4.62               & 63.01                & 4.50               & 66.65             & 40.48             \\
   & 2.0$\times$ capacity             & {\bf 3.80}         & {\bf 66.61}          & {\bf 4.18}         & {\bf 70.06}       & {\bf 42.51}       \\
\tabucline{-}
*  & Kaiser filter, $n = 6$           & {\bf 4.62}         & {\bf 63.01}          & 4.50               & {\bf 66.65}       & {\bf 40.48}       \\
   & Lanczos filter, $a = 2$          & 4.69               & 51.93                & {\bf 4.44}         & 57.70             & 25.25             \\
   & Gaussian filter, $\sigma = 0.4$  & 5.91               & 56.89                & 5.73               & 59.53             & 39.43             \\
}

\newcommand{\genGroupConv}{%
            & {\bf G-CNN comparison}            & FID\,$\downarrow$  & EQ-T\,$\uparrow$  & EQ-R\,$\uparrow$  & Params  & Time                \\
\tabucline{-}
\textsc{*}  & \FINAL{StyleGAN3-T (ours)}        & 4.62               & 63.01             & 13.12             & 23.3M   & {\bf 1.00$\times$}  \\
\textsc{}   & + $p4$ symmetry \cite{Cohen2016}  & 4.69               & 61.90             & 17.07             & 21.8M   & 2.48$\times$        \\
\textsc{*}  & \FINAL{StyleGAN3-R (ours)}        & {\bf 4.50}         & {\bf 66.65}       & {\bf 40.48}       & 15.8M   & 1.37$\times$        \\
}

\newcommand{\tabDatasets}{%
\newcolumntype{x}{>{\centering\arraybackslash\hspace{0pt}}p{8.0mm}}%
\newcolumntype{y}{>{\centering\arraybackslash\hspace{0pt}}p{9.6mm}}%
\renewcommand{\cs}{\hspace{2mm}}%
\tabulinesep=0.50mm%
\tabulinestyle{0.17mm}%
\begin{tabu}{|l@{\hspace{1.4mm}}|l|x@{\cs}y@{\cs}y|}
\tabucline{-}
\genDatasets
\tabucline{-}
\end{tabu}%
}

\newcommand{\tabAblations}{%
\newcolumntype{x}{>{\centering\arraybackslash\hspace{0pt}}p{7.3mm}}%
\newcolumntype{y}{>{\centering\arraybackslash\hspace{0pt}}p{9.6mm}}%
\renewcommand{\cs}{\hspace{2.5mm}}%
\tabulinesep=0.55mm%
\tabulinestyle{0.17mm}%
\begin{tabu}{|c@{\hspace{1.4mm}}l|x@{\cs}y|x@{\cs}y@{\cs}y|}
\tabucline{-}
\genAblations
\tabucline{-}
\end{tabu}%
}

\newcommand{\tabGroupConv}{%
\newcolumntype{x}{>{\centering\arraybackslash\hspace{0pt}}p{7.3mm}}%
\newcolumntype{y}{>{\centering\arraybackslash\hspace{0pt}}p{9.6mm}}%
\newcolumntype{z}{>{\centering\arraybackslash\hspace{0pt}}p{8.4mm}}%
\renewcommand{\cs}{\hspace{2.5mm}}%
\tabulinesep=0.55mm%
\tabulinestyle{0.17mm}%
\begin{tabu}{|c@{\hspace{1.4mm}}l@{\hspace{5.1mm}}|x@{\cs}y@{\cs}y|y@{\cs}z|}
\tabucline{-}
\genGroupConv
\tabucline{-}
\end{tabu}%
}

\newcommand{\figResultTables}[1]{
\renewcommand{\vv}{54.2mm}%
\begin{figure}[t]
\centering%
\footnotesize%
\parbox[b][\vv]{0.45\linewidth}{%
    \scalebox{0.72}{\tabDatasets}\hfill\\%
}%
\hfill%
\parbox[b][\vv]{0.53\linewidth}{%
    \hfill\scalebox{0.73}{\tabAblations}\vfill%
    \hfill\scalebox{0.73}{\tabGroupConv}\\%
}%
\caption{
\textbf{Left:} Results for six datasets. We use adaptive discriminator augmentation (ADA)~\cite{Karras2020} for the smaller datasets. 
``StyleGAN2'' corresponds to our baseline config~\textsc{b} with Fourier features.
\textbf{Right:} Ablations and comparisons for FFHQ-U (unaligned FFHQ) at 256$^2$.
* indicates our default choices.
}
\label{#1}
\end{figure}
}

\newcommand{\figResults}[1]{
\begin{figure}[t]
\includegraphics[width=\linewidth,trim={2.7 8.5 8.5 2.4},clip]{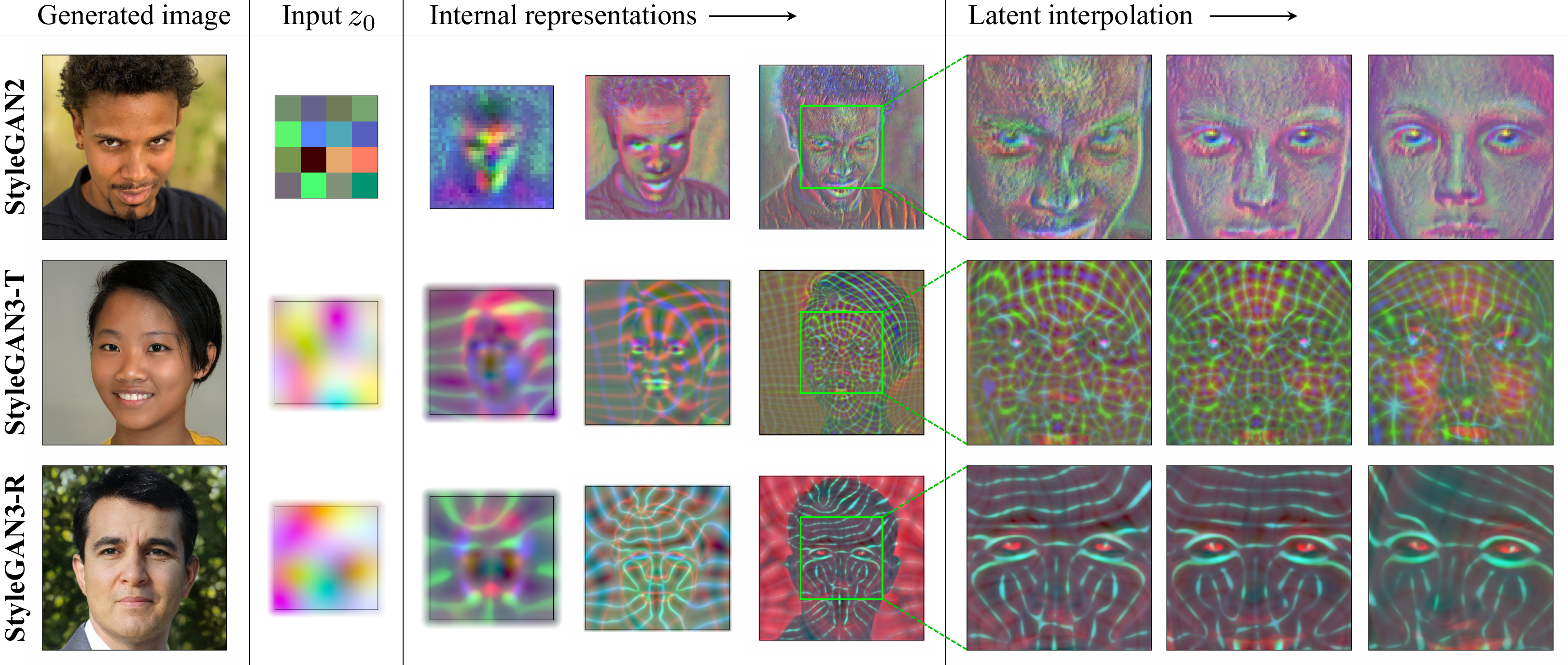}
\caption{
Example internal representations (3 feature maps as RGB) in StyleGAN2 and our generators.
}
\label{#1}
\end{figure}
}

\begin{document}
\maketitle
\confnotice

\ifappendix
  \newcommand{\refappResults}{Appendix~\ref{app:results}}
  \newcommand{\refappDatasets}{Appendix~\ref{app:datasets}}
  \newcommand{\refappFilter}{Appendix~\ref{app:filter}}
  \newcommand{\refappKernel}{Appendix~\ref{app:kernel}}
  \newcommand{\refappMetrics}{Appendix~\ref{app:metrics}}
  \newcommand{\refappMetricsFirstContact}{Appendix~\ref{app:metrics}}
  \newcommand{\refappImplementation}{Appendix~\ref{app:implementation}}
\else
  \newcommand{\refappResults}{Appendix~A}
  \newcommand{\refappDatasets}{Appendix~B}
  \newcommand{\refappFilter}{Appendix~C}
  \newcommand{\refappKernel}{Appendix~D}
  \newcommand{\refappMetrics}{Appendix~E}
  \newcommand{\refappMetricsFirstContact}{Appendix~E in the Supplement}
  \newcommand{\refappImplementation}{Appendix~F}
\fi

\begin{abstract}
We observe that despite their hierarchical convolutional nature, the synthesis process of typical generative adversarial networks depends on absolute pixel coordinates in an unhealthy manner. 
This manifests itself as, e.g., detail appearing to be glued to image coordinates instead of the surfaces of depicted objects. 
We trace the root cause to careless signal processing that causes aliasing in the generator network. 
Interpreting all signals in the network as continuous, we derive generally applicable, small architectural changes that guarantee that unwanted information cannot leak into the hierarchical synthesis process. 
The resulting networks match the FID of StyleGAN2 but differ dramatically in their internal representations, and they are fully equivariant to translation and rotation even at subpixel scales. 
Our results pave the way for generative models better suited for video and animation.
\end{abstract}

\section{Introduction}

The resolution and quality of images produced by generative adversarial networks (GAN) \cite{Goodfellow2014} have seen rapid improvement recently \cite{Karras2017,Brock2018,Karras2018,Karras2019}. 
They have been used for a variety of applications, including
image editing \cite{Richardson2020,Suzuki2018,Park2020,Gu2019,Menon2020,Alaluf2021},
domain translation \cite{Zhu2017,Liu2017B,Wang2018,Park2019},
and video generation \cite{Tulyakov2017,Chu2018,Hao2021}. 
While several ways of controlling the generative process have been found \cite{JunYan2018,Jahanian2019,Broad2020,Park2019,Harkonen2020,Abdal2020,Bau2020,Ramesh2020,Bau2021},
the foundations of the synthesis process remain only partially understood.%

In the real world, details of different scale tend to transform hierarchically. For instance, moving a head causes the nose to move, which in turn moves the skin pores on it. The structure of a typical GAN generator is analogous: coarse, low-resolution features are hierarchically refined by upsampling layers, locally mixed by convolutions, and new detail is introduced through nonlinearities. %
We observe that despite this superficial similarity, current GAN architectures do not synthesize images in a natural hierarchical manner:
the coarse features mainly control the \emph{presence} of finer features, but not their precise positions. Instead, much of the fine detail appears to be fixed in pixel coordinates. 
This disturbing ``texture sticking'' is clearly visible in latent interpolations (see Figure~\ref{fig:intro} and our accompanying videos on the project page 
\FINAL{\projectpage}),
breaking the illusion of a solid and coherent object moving in space.
Our goal is an architecture that exhibits a more natural transformation hierarchy, where the exact sub-pixel position of each feature is exclusively inherited from the underlying coarse features.

It turns out that current networks can partially bypass the ideal hierarchical construction by drawing on unintentional positional references available to the intermediate layers through image borders \cite{Islam2020,Kayhan2020,Xu2020}, per-pixel noise inputs~\cite{Karras2018} and positional encodings, and aliasing~\cite{Azulay2018,Zhang2019}. Aliasing, despite being a subtle and critical issue~\cite{Parmar2021}, has received little attention in the GAN literature. We identify two sources for it: 1) faint after-images of the pixel grid resulting from non-ideal upsampling filters%
\footnote{\FINAL{Consider nearest neighbor upsampling. If we upsample a 4$\times$4 image to 8$\times$8, the original pixels will be clearly visible, allowing one to reliably distinguish between even and odd pixels. Since the same is true on all scales, this (leaked) information makes it possible to reconstruct even the absolute pixel coordinates. With better filters such as bilinear or bicubic, the clues get less pronounced, but are nevertheless evident for the generator.}}
such as nearest, bilinear, or strided convolutions, and 2) the pointwise application of nonlinearities such as ReLU~\cite{ReLU} or swish~\cite{swish}. We find that the network has the means and motivation to amplify even the slightest amount of aliasing and combining it over multiple scales allows it to build a basis for texture motifs that are fixed in screen coordinates. This holds for all filters commonly used in deep learning \cite{Zhang2019,Vasconcelos2021}, and even high-quality filters used in image processing.

How, then, do we eliminate the unwanted side information and thereby stop the network from using it? 
While borders can be solved by simply operating on slightly larger images, %
aliasing is much harder.
We begin by noting that aliasing is most naturally treated in the classical Shannon-Nyquist signal processing framework, and switch focus to bandlimited functions on a continuous domain that are merely represented by discrete sample grids. 
Now, successful elimination of all sources of positional references means that details can be generated equally well regardless of pixel coordinates, which in turn is equivalent to enforcing continuous \emph{equivariance} to sub-pixel translation (and optionally rotation) in all layers.
To achieve this, we describe a comprehensive overhaul of all signal processing aspects of the StyleGAN2 generator~\cite{Karras2019}.
Our contributions include the surprising finding that current upsampling filters are simply not aggressive enough in suppressing aliasing, and that extremely high-quality filters with over 100dB attenuation are required.
Further, we present a principled solution to aliasing caused by pointwise nonlinearities \cite{Azulay2018} by considering their effect in the continuous domain and appropriately low-pass filtering the results.
We also show that after the overhaul, a model based on 1$\times$1 convolutions yields a strong, rotation equivariant generator.

\figIntro{fig:intro} %

Once aliasing is adequately suppressed to force the model to implement more natural hierarchical refinement, its mode of operation changes drastically: \FINAL{the emergent internal representations now include coordinate systems that allow 
details to be correctly attached to the underlying surfaces.}
This promises significant improvements to models that generate video and animation. The new \FINAL{StyleGAN3} generator matches StyleGAN2 in terms of FID~\cite{Heusel2017}, while being slightly heavier computationally.
\FINAL{Our implementation and pre-trained models are available at \codepage}

Several recent works have studied the lack of translation equivariance
in CNNs, mainly in the context of classification~\cite{Islam2020,Kayhan2020,Xu2020,Azulay2018,Manfredi2020,Zhang2019,Chaman2020,Zou2020,Vasconcelos2021}. We significantly expand upon the antialiasing measures in this literature and show that doing so induces a fundamentally altered image generation behavior.
Group-equivariant CNNs aim to generalize the efficiency benefits of translational weight sharing to, e.g., rotation~\cite{Cohen2016,Worrall2017,Weiler2017,Weiler2021} and scale~\cite{Worrall2019}. Our 1$\times$1 convolutions can be seen an instance of a continuously $\textrm{E}(2)$-equivariant model~\cite{Weiler2021} that remains compatible with, e.g., channel-wise ReLU nonlinearities and modulation.
Dey et al.~\cite{Dey2021} apply $90^\circ$ rotation-and-flip equivariant CNNs~\cite{Cohen2016} to GANs and show improved data efficiency. Our work is complementary, and not motivated by efficiency. 
Recent implicit network~\cite{Sitzmann2019,Tancik2020,Chen2020} based GANs~\cite{Anokhin2020,Skorokhodov2020} generate each pixel independently via similar 1$\times$1 convolutions. While equivariant, these models do not help with texture sticking, as they do not use an upsampling hierarchy or implement a shallow non-antialiased one.

\section{Equivariance via continuous signal interpretation}
\label{sec:theory}

To begin our analysis of equivariance in CNNs, we shall first rethink our view of what exactly is the signal that flows through a network.
Even though data may be stored as values in a pixel grid, we cannot na\"ively hold these values to directly represent the signal.
Doing so would prevent us from considering operations as trivial as translating the contents of a feature map by half a pixel.

According to the Nyquist--Shannon sampling theorem \cite{Shannon1949}, a regularly sampled signal can represent any continuous signal containing frequencies between zero and half of the sampling rate.
Let us consider a two-dimensional, discretely sampled feature map $\Feat[\xx]$ that consists of a regular grid of Dirac impulses of varying magnitudes, spaced $1/\sfreq$ units apart where $\sfreq$ is the sampling rate.
This is analogous to an infinite two-dimensional grid of values.

Given $\Feat[\xx]$ and $\sfreq$, the Whittaker--Shannon interpolation formula \cite{Shannon1949} states that the corresponding continuous representation $\feat(\xx)$ is obtained by convolving the discretely sampled Dirac grid $\Feat[\xx]$ with an ideal interpolation filter $\idealsqs$, i.e.,
$\feat(\xx) = \big(\idealsqs\conv\Feat\big)(\xx)$,
where $\conv$ denotes continuous convolution and $\idealsqs(\xx) = \sinc(\sfreq x_0) \cdot \sinc(\sfreq x_1)$ using the signal processing convention of defining $\sinc(x) = \sin(\pi x)/(\pi x)$.
$\idealsqs$ has a bandlimit of $\sfreq/2$ along the horizontal and vertical dimensions, ensuring that the resulting continuous signal captures all frequencies that can be represented with sampling rate $\sfreq$.

Conversion from the continuous to the discrete domain corresponds to sampling the continuous signal $\feat(\xx)$ at the sampling points of $\Feat[\xx]$ that we define to be offset by half the sample spacing to lie at the ``pixel centers'', see Figure~\ref{fig:theory}, left.
This can be expressed as a pointwise multiplication with a two-dimensional Dirac comb $\combs(\xx)=\sum_{X\!\in\integer^2}\delta\big(\xx - (X + \frac{1}{2})/s\big)$.

We earmark the unit square $\xx \in [0,1]^2$ in $\feat(\xx)$ as our canvas for the signal of interest.
In $\Feat[\xx]$ there are $\sfreq^2$ discrete samples in this region, but the above convolution with $\idealsqs$ means that values of $\Feat[\xx]$ outside the unit square also influence $\feat(\xx)$ inside it.
Thus storing an \mbox{$\sfreq\times\sfreq$} -pixel feature map is not sufficient; in theory, we would need to store the entire infinite $\Feat[\xx]$.
As a practical solution, we store $\Feat[\xx]$ as a two-dimensional array that covers a region slightly larger than the unit square (Section~\ref{sec:practice_additions}). %

Having established correspondence between bandlimited, continuous feature maps $\feat(\xx)$ and discretely sampled feature maps $\Feat[\xx]$, we can shift our focus away from the usual pixel-centric view of the signal.
In the remainder of this paper, we shall interpret $\feat(\xx)$ as being the actual signal being operated on, and the discretely sampled feature map $\Feat[\xx]$ as merely a convenient encoding for it.

\figTheory{fig:theory} %

\paragraph{Discrete and continuous representation of network layers}

Practical neural networks operate on the discretely sampled feature maps.
Consider operation $\Layer$ (convolution, nonlinearity, etc.) operating on a discrete feature map: $\Feat'=\Layer(\Feat)$.
The feature map has a corresponding continuous counterpart, so we also have a corresponding mapping in the continuous domain: $\feat'=\layer(\feat)$.
Now, an operation specified in one domain can be seen to perform a corresponding operation in the other domain:
\begin{align}
\label{eq:continuous_discrete}
\layer(\feat) &= \idealsqsp \conv \Layer(\combs\mult\feat), &
\Layer(\Feat) &= \combsp\mult\layer(\idealsqs \conv \Feat),
\end{align}
where $\mult$ denotes pointwise multiplication and $\sfreq$ and $\sfreq'$ are the input and output sampling rates.
Note that in the latter case $\layer$ must not introduce frequency content beyond the output bandlimit $\sfreq'/2$.

\subsection{Equivariant network layers}
\label{sec:theory_layers}

Operation $\layer$ is equivariant with respect to a spatial transformation $\trans$ of the 2D plane if it commutes with it in the continuous domain: $\trans \circ \layer = \layer \circ \trans$.
We note that when inputs are bandlimited to $\sfreq/2$, an equivariant operation must not generate frequency content above the output bandlimit of $\sfreq'/2$, as otherwise no faithful discrete output representation exists.

We focus on two types of equivariance in this paper: translation and rotation.
In the case of rotation the spectral constraint is somewhat stricter\,---\,%
rotating an image corresponds to rotating the spectrum, and in order to guarantee the bandlimit in both horizontal and vertical direction, the spectrum must be limited to a disc with radius $\sfreq/2$.
This applies to both the initial network input as well as the bandlimiting filters used for downsampling, as will be described later.

We now consider the primitive operations in a typical generator network: convolution, upsampling, downsampling, and nonlinearity.
Without loss of generality, we discuss the operations acting on a single feature map: pointwise linear combination of features has no effect on the analysis.

\paragraph{Convolution}

Consider a standard convolution with a discrete kernel $\Kernel$.
We can interpret $\Kernel$ as living in the same grid as the input feature map, with sampling rate $\sfreq$.
The discrete-domain operation is simply $\Layern{conv}(\Feat) = \Kernel \conv \Feat$, and we obtain the corresponding continuous operation from Eq.~\ref{eq:continuous_discrete}:
\begin{equation}
\layern{conv}(\feat)
= \idealsqs \conv \big( \Kernel \conv (\combs \mult \feat) \big)
= \Kernel \conv \big( \idealsqs \conv (\combs \mult \feat) \big)
= \Kernel \conv \feat
\end{equation}
due to commutativity of convolution and the fact that discretization followed by convolution with ideal low-pass filter, both with same sampling rate $\sfreq$, is an identity operation, i.e., $\idealsqs\conv(\combs\mult\feat)=\feat$.
In other words, the convolution operates by continuously sliding the discretized kernel over the continuous representation of the feature map.
This convolution introduces no new frequencies, so the bandlimit requirements for both translation and rotation equivariance are trivially fulfilled.

Convolution also commutes with translation in the continuous domain, and thus the operation is equivariant to translation.
For rotation equivariance, the discrete kernel $\Kernel$ needs to be radially symmetric.
We later show in Section~\ref{sec:practice_additions} that trivially symmetric 1$\times$1 convolution kernels are, despite their simplicity, a viable choice for rotation equivariant generative networks.%

\paragraph{Upsampling and downsampling}

Ideal upsampling does not modify the continuous representation. %
Its only purpose is to increase the output sampling rate ($\sfreq' > \sfreq$) to add headroom in the spectrum where subsequent layers may introduce additional content.
Translation and rotation equivariance follow directly from upsampling being an identity operation in the continuous domain.
With $\layern{up}(\feat) = \feat$, the discrete operation according to Eq.~\ref{eq:continuous_discrete} is $\Layern{up}(\Feat) = \combsp\mult(\idealsqs \conv \Feat)$.
If we choose $\sfreq' = n\sfreq$ with integer $n$, this operation can be implemented %
by first interleaving $\Feat$ with zeros to increase its sampling rate and then convolving it with a discretized filter $\combsp\mult\idealsqs$.

In downsampling, we must low-pass filter $\feat$ to remove frequencies above the output bandlimit, so that the signal can be represented faithfully in the coarser discretization. 
The operation in continuous domain is $\layern{down}(\feat) = \ideallow_{\sfreq'}\conv\feat$, where an ideal low-pass filter $\ideallow_{\sfreq} := {\sfreq}^2 \cdot \idealsqs$ is simply the corresponding interpolation filter normalized to unit mass.
The discrete counterpart is $\Layern{down}(\Feat) = \combsp\mult\big(\ideallow_{\sfreq'}\conv(\idealsqs\conv\Feat)\big) = 1/\sfreq^2 \cdot \combsp\mult(\ideallow_{\sfreq'} \conv \ideallow_{\sfreq} \conv\Feat) = (\sfreq' / \sfreq)^2 \cdot \combsp\mult(\idealsqsp \conv\Feat)$.
The latter equality follows from $\ideallow_{\sfreq} \conv \ideallow_{\sfreq'} = \ideallow_{\text{min}(\sfreq, \sfreq')}$. %
Similar to upsampling, downsampling by an integer fraction can be implemented with a discrete convolution followed by dropping sample points.
Translation equivariance follows automatically from the commutativity of $\layern{down}(\feat)$ with translation, but for rotation equivariance we must replace $\idealsqsp$ with a radially symmetric filter with disc-shaped frequency response.
The ideal such filter~\cite{Blahut2004} is given by $\idealdiscs(\xx) = \jinc(s\lVert\xx\rVert) = 2J_1(\pi\sfreq\lVert\xx\rVert)/(\pi\sfreq\lVert\xx\rVert)$, where $J_1$ is the first order Bessel function of the first kind.

\paragraph{Nonlinearity}

Applying a pointwise nonlinearity $\rawnonlin$ in the discrete domain does not commute with fractional translation or rotation.
However, in the continuous domain, any pointwise function commutes trivially with geometric transformations and is thus equivariant to translation and rotation.
Fulfilling the bandlimit constraint is another question\,---\,applying, e.g., ReLU in the continuous domain may introduce arbitrarily high frequencies that cannot be represented in the output.

A natural solution is to eliminate the offending high-frequency content by convolving the continuous result with the ideal low-pass filter $\ideallow_{\sfreq}$. %
Then, the continuous representation of the nonlinearity becomes $\layernm{\rawnonlin}(\feat) = \ideallow_{\sfreq}\conv\rawnonlin(\feat) = \sfreq^2 \cdot \idealsqs\conv\rawnonlin(\feat)$ and the discrete counterpart is $\Layernm{\rawnonlin}(\Feat) = \sfreq^2 \cdot \combs\mult(\idealsqs\conv\rawnonlin(\idealsqs\conv\Feat))$ (see Figure~\ref{fig:theory}, right).
This discrete operation cannot be realized without temporarily entering the continuous representation. %
We approximate this by upsampling the signal, applying the nonlinearity in the higher resolution, and downsampling it afterwards. Even though the nonlinearity is still performed in the discrete domain,
we have found that only a 2$\times$ temporary resolution increase is sufficient for high-quality equivariance.
For rotation equivariance, we must use the radially symmetric interpolation filter $\idealdiscs$ in the downsampling step, as discussed above.

Note that nonlinearity is the only operation capable of generating novel frequencies in our formulation, and that we can limit the range of these novel frequencies by applying a reconstruction filter with a lower cutoff than $\sfreq/2$ before the final discretization operation.
This gives us precise control over how much new information is introduced by each layer of a generator network (Section~\ref{sec:practice_additions}).

\section{Practical application to generator network}
\label{sec:practice}

We will now apply the theoretical ideas from the previous section in practice, by converting the well-established StyleGAN2~\cite{Karras2019} generator to be fully equivariant to translation and rotation.
We will introduce the necessary changes step-by-step, evaluating their impact in Figure~\ref{fig:BridgeTables}.
The discriminator remains unchanged in our experiments.

The StyleGAN2 generator consists of two parts.
First, a \emph{mapping network} transforms an initial, normally distributed latent to an intermediate latent code $\ww \sim \WW$.
Then, a \emph{synthesis network} $\Synthesis$ starts from a learned 4$\times$4$\times$512 constant $\Feat_0$ and applies a sequence of $N$ layers\,---\,consisting of convolutions, nonlinearities, upsampling, and per-pixel noise\,---\,to produce an output image $\Feat_N = \Synthesis(\Feat_0; \ww)$.
The intermediate latent code $\ww$ controls the modulation of the convolution kernels in $\Synthesis$.
The layers follow a rigid 2$\times$ upsampling schedule, where two layers are executed at each resolution and the number of feature maps is halved after each upsampling.
Additionally, StyleGAN2 employs skip connections, mixing regularization~\cite{Karras2018}, and path length regularization.

\figBridgeTables{fig:BridgeTables} %

Our goal is to make every layer of $\Synthesis$ equivariant w.r.t.~the continuous signal, so that all finer details transform together with the coarser features of a local neighborhood.
If this succeeds, the entire network becomes similarly equivariant.
In other words, we aim to make the \emph{continuous} operation $\synthesis$ of the synthesis network equivariant w.r.t.~transformations $\trans$ (translations and rotations) applied on the continuous input $\feat_0$:
$\synthesis(\trans[\feat_0]; \ww) = \trans[\synthesis(\feat_0; \ww)]$.
To evaluate the impact of various architectural changes and practical approximations, we need a way to measure how well the network implements the equivariances. For translation equivariance, we report the peak signal-to-noise ratio (PSNR) in decibels (dB) between two sets of images, obtained by translating the input and output of the synthesis network by a random amount, resembling the definition by Zhang~\cite{Zhang2019}:
\begin{equation}
	\metrict{} = 10 \cdot \log_{10} \left( I^2_\mathit{max} \big/ \expectation_{\ww \sim \WW, x \sim \mathcal{X}^2, p \sim \mathcal{V}, c \sim \mathcal{C}} \left[ \big( \synthesis(\trans_x[\feat_0]; \ww)_c(p) - \trans_x [\synthesis(\feat_0; \ww)]_c(p) \big)^2 \right] \right)
\end{equation}
Each pair of images, corresponding to a different random choice of $\ww$, is sampled at integer pixel locations $p$ within their mutually valid region $\mathcal{V}$.
Color channels $c$ are processed independently, and the intended dynamic range of generated images $-1\ldots{+}1$ gives $I_\mathit{max} = 2$. %
Operator $\trans_x$ implements spatial translation with 2D offset $x$, here drawn from distribution $\mathcal{X}^2$ of integer offsets.
We define an analogous metric \metricr{} for rotations, with the rotation angles drawn from $\mathcal{U}(0^{\circ},360^{\circ})$.
\refappMetricsFirstContact{} gives implementation details and our accompanying videos highlight the practical relevance of different dB values.

\vspace{-1mm}
\subsection{Fourier features and baseline simplifications (configs \textsc{b}--\textsc{d})}
\label{sec:practice_removals}

To facilitate exact continuous translation and rotation of the input $\feat_0$, we replace the learned input constant in StyleGAN2 with Fourier features \cite{Tancik2020,Xu2020}, which also has the advantage of naturally defining a spatially infinite map.
We sample the frequencies uniformly within the circular frequency band $\freqc = 2$, matching the original 4$\times$4 input resolution, and keep them fixed over the course of training.
This change (configs~\textsc{a} and~\textsc{b} in Figure~\ref{fig:BridgeTables}, left) slightly improves FID and, crucially, allows us to compute the equivariance metrics without having to approximate the operator $\trans$.
This baseline architecture is far from being equivariant; our accompanying videos show that the output images deteriorate drastically when the input features are translated or rotated from their original position.

Next, we remove the per-pixel noise inputs \FINAL{because they are strongly at odds with our goal of a natural transformation hierarchy, i.e., that the exact sub-pixel position of each feature is exclusively inherited from the underlying coarse features.}
While this change (config~\textsc{c}) is approximately FID-neutral, it fails to improve the equivariance metrics when considered in isolation.

To further simplify the setup, we decrease the mapping network depth as recommended by Karras~et~al.~\cite{Karras2020} and disable mixing regularization and path length regularization~\cite{Karras2019}.
Finally, we also eliminate the output skip connections. 
We hypothesize that their benefit is mostly related to gradient magnitude dynamics during training and address the underlying issue more directly using a simple normalization
before each convolution. We track the exponential moving average $\sigma^2 = \expectation[x^2]$ over all pixels and feature maps during training, and divide the feature maps by $\lowsqrt{\sigma^2}$.
In practice, we bake the division into the convolution weights to improve efficiency.
These changes (config~\textsc{d}) bring FID back to the level of original StyleGAN2, while leading to a slight improvement in translation equivariance.

\figPractice{fig:practice} %

\vspace{-1mm}
\subsection{Step-by-step redesign motivated by continuous interpretation}
\paragraph{Boundaries and upsampling (config \textsc{e})}
\label{sec:practice_additions}
\hspace*{-1mm}%
Our theory assumes an infinite spatial extent for the feature maps, which we approximate by maintaining a fixed-size margin around the target canvas, cropping to this extended canvas after each layer.
This explicit extension is necessary as border padding is known to leak absolute image coordinates into the internal representations \cite{Islam2020,Kayhan2020,Xu2020}.
In practice, we have found a $10$-pixel margin to be enough; further increase has no noticeable effect on the results.

Motivated by our theoretical model, we replace the bilinear 2$\times$ upsampling filter with a better approximation of the ideal low-pass filter.
We use a windowed $\sinc$ filter with a relatively large Kaiser window~\cite{Oppenheim2009} of size $n = 6$, meaning that each output pixel is affected by 6 input pixels in upsampling and each input pixel affects 6 output pixels in downsampling.
Kaiser window is a particularly good choice for our purposes, because it offers explicit control over the transition band and attenuation (Figure~\ref{fig:practice}a).
In the remainder of this section, we specify the transition band explicitly and compute the remaining parameters using Kaiser's original formulas (\refappFilter{}).
For now, we choose to employ \emph{critical sampling} and set the filter cutoff $\freqc = \sfreq/2$, i.e., exactly at the bandlimit, %
and transition band half-width $\freqh=(\lowsqrt{2}-1)(\sfreq/2)$.
Recall that sampling rate $\sfreq$ equals the width of the canvas in pixels, given our definitions in Section~\ref{sec:theory}.

The improved handling of boundaries and upsampling (config~\textsc{e}) leads to better translation equivariance. 
\FINAL{However, FID is compromised by 16\%, probably because we started to constrain what the feature maps can contain.}
In a further ablation (Figure~\ref{fig:BridgeTables}, right), smaller resampling filters ($n=4$) hurt translation equivariance, while larger filters ($n=8$) mainly increase training time.

\vspace{-1mm}
\paragraph{Filtered nonlinearities (config \textsc{f})}

Our theoretical treatment of nonlinearities calls for wrapping each leaky ReLU \FINAL{(or any other commonly used non-linearity)} between $m\times$ upsampling and $m\times$ downsampling, for some magnification factor $m$.
We further note that the order of upsampling and convolution can be switched by virtue of the signal being bandlimited, allowing us to fuse the regular 2$\times$ upsampling and a subsequent $m \times$ upsampling related to the nonlinearity into a single $2m \times$ upsampling.
In practice, we find $m=2$ to be sufficient (Figure~\ref{fig:BridgeTables}, right), again improving \metrict{} (config~\textsc{f}).
Implementing the upsample-LReLU-downsample sequence is not efficient using the primitives available in current deep learning frameworks \cite{Tensorflow,pyTorch}, and thus we implement a custom CUDA kernel (\refappKernel{}) that combines these operations (Figure~\ref{fig:practice}b), leading to 10$\times$ faster training and considerable memory savings.

\vspace{-0.5\baselineskip}
\paragraph{Non-critical sampling (config \textsc{g})}

The critical sampling scheme\,---\,where filter cutoff is set exactly at the bandlimit\,---\,is ideal for many image processing applications as it strikes a good balance between antialiasing and the retention of high-frequency detail \cite{Turkowski1990}.
However, our goals are markedly different because aliasing is highly detrimental for the equivariance of the generator.
While high-frequency detail is important in the output image and thus in the highest-resolution layers, it is less important in the earlier ones given that their exact resolutions are somewhat arbitrary to begin with.

To suppress aliasing, we can simply lower the cutoff frequency to $\freqc = \sfreq/2 - \freqh$, which ensures that all alias frequencies (above $\sfreq/2$) are in the stopband.%
\footnote{Here, $\freqc$ and $\freqh$ correspond to the output (downsampling) filter of each layer. The input (upsampling) filters are based on the properties of the incoming signal, i.e., the output filter parameters of the previous layer.}
For example, lowering the cutoff of the blue filter in Figure~\ref{fig:practice}a would move its frequency response left so that the the worst-case attenuation of alias frequencies improves from $6$\,dB to $40$\,dB.
This \emph{oversampling} can be seen as a computational cost of better antialiasing, as we now use the same number of samples to express a slower-varying signal than before.
In practice, we choose to lower $\freqc$ on all layers except the highest-resolution ones, because in the end the generator must be able to produce crisp images to match the training data.
As the signals now contain less spatial information, we modify the heuristic used for determining the number of feature maps to be inversely proportional to $\freqc$ instead of the sampling rate $\sfreq$.
These changes (config~\textsc{g}) further improve translation equivariance and push FID below the original StyleGAN2.

\vspace{-0.5\baselineskip}
\paragraph{Transformed Fourier features (config \textsc{h})}

Equivariant generator layers are well suited for modeling unaligned and arbitrarily oriented datasets, because any geometric transformation introduced to the intermediate features $\feat_i$ will directly carry over to the final image $\feat_N$.
Due to the limited capability of the layers themselves to introduce global transformations, however, the input features $\feat_0$ play a crucial role in defining the global orientation of $\feat_N$.
To let the orientation vary on a per-image basis, the generator should have the ability to transform $\feat_0$ based on $\ww$.
This motivates us to introduce a learned affine layer that outputs global translation and rotation parameters for the input Fourier features (Figure~\ref{fig:practice}b and \refappImplementation{}).
The layer is initialized to perform an identity transformation%
, but learns to use the mechanism over time when beneficial; in config~\textsc{h} this improves the FID slightly.

\vspace{-0.5\baselineskip}
\paragraph{Flexible layer specifications (config \textsc{t})}

Our changes have improved the equivariance quality considerably, but some visible artifacts still remain as our accompanying videos demonstrate.
On closer inspection, it turns out that the attenuation of our filters (as defined for config \textsc{g}) is still insufficient for the lowest-resolution layers.
These layers tend to have rich frequency content near their bandlimit, which calls for extremely strong attenuation to completely eliminate aliasing.

So far, we have used the rigid sampling rate progression from StyleGAN2, coupled with simplistic choices for filter cutoff $\freqc$ and half-width $\freqh$, but this need not be the case; we are free to specialize these parameters on a per-layer basis.
In particular, we would like $\freqh$ to be high in the lowest-resolution layers to maximize attenuation in the stopband, but low in the highest-resolution layers to allow matching high-frequency details of the training data.

Figure~\ref{fig:practice}c illustrates an example progression of filter parameters in a 14-layer generator with two critically sampled full-resolution layers at the end.
The cutoff frequency grows geometrically from $\freqc=2$ in the first layer to $\freqc=\sfreq_N/2$ in the first critically sampled layer.
We choose the minimum acceptable stopband frequency to start at \low{$\freqtzero=2^{2.1}$}, and it grows geometrically but slower than the cutoff frequency.
In our tests, the stopband target at the last layer is \low{$\freqt=\freqc\cdot2^{0.3}$}, but the progression is halted at the first critically sampled layer.
Next, we set the sampling rate $\sfreq$ for each layer so that it accommodates frequencies up to $\freqt$, rounding up to the next power of two without exceeding the output resolution.
Finally, to maximize the attenuation of aliasing frequencies, we set the transition band half-width to $\freqh=\max(\sfreq/2,\freqt)-\freqc$, i.e., making it as wide as possible within the limits of the sampling rate, but at least wide enough to reach $\freqt$.
The resulting improvement depends on how much slack is left between $\freqt$ and $\sfreq/2$; as an extreme example, the first layer stopband attenuation improves from $42$\,dB to $480$\,dB using this scheme.

The new layer specifications again improve translation equivariance (config~\textsc{t}), eliminating the remaining artifacts.
A further ablation (Figure~\ref{fig:BridgeTables}, right) shows that $\freqtzero$ provides an effective way to trade training speed for equivariance quality.
Note that the number of layers is now a free parameter that does not directly depend on the output resolution.
In fact, we have found that a fixed choice of $N$ works consistently across multiple output resolutions and makes other hyperparameters such as learning rate behave more predictably.
We use $N=14$ in the remainder of this paper.

\vspace{-0.25\baselineskip}
\paragraph{Rotation equivariance (config \textsc{r})}

We obtain a rotation equivariant version of the network with two changes.
First, we replace the 3$\times$3 convolutions with 1$\times$1 on all layers and compensate for the reduced capacity by doubling the number of feature maps.
\FINAL{Only the upsampling and downsampling operations spread information between pixels in this config.}
Second, we replace the $\sinc$-based downsampling filter with a radially symmetric $\jinc$-based one that we construct using the same Kaiser scheme (\refappFilter{}).
We do this for all layers except the two critically sampled ones, where it is important to match the potentially non-radial spectrum of the training data.
These changes (config \textsc{r}) improve EQ-R without harming FID, even though each layer has 56\% fewer trainable parameters.

\FINAL{We also employ an additional stabilization trick in this configuration. Early on in the training, we blur all images the discriminator sees using a Gaussian filter. We start with $\sigma=10$ pixels, which we ramp to zero over the first 200k images. This prevents the discriminator from focusing too heavily on high frequencies early on. Without this trick, config \textsc{r} is prone to early collapses because the generator sometimes learns to produce
high frequencies with a small delay, trivializing the discriminator's task.}

\vspace{-0.4\baselineskip}
\section{Results}
\vspace{-0.4\baselineskip}
\label{sec:results}

\figResultTables{fig:ResultTables} %

Figure~\ref{fig:ResultTables} gives results for six datasets using StyleGAN2 \cite{Karras2019} as well as our alias-free \FINAL{StyleGAN3-T and StyleGAN3-R} generators.
In addition to the standard \textsc{FFHQ}~\cite{Karras2018} and \textsc{Metfaces}~\cite{Karras2020}, we created unaligned versions of them.
We also created  a properly resampled version of \textsc{AFHQ}~\cite{AFHQ} and collected a new \textsc{Beaches} dataset.
\refappDatasets{} describes the datasets in detail.
The results show that our FID remains competitive with StyleGAN2.
\FINAL{StyleGAN3-T and StyleGAN3-R} perform equally well in terms of FID, and both show a very high level of translation equivariance.
As expected, only the latter provides rotation equivariance. 
In FFHQ (1024$\times$1024) the three generators had 30.0M, 22.3M and 15.8M parameters, while the training times were 1106, 1576 (+42\%) and 2248 (+103\%) GPU hours. %
Our accompanying videos show side-by-side comparisons with StyleGAN2, demonstrating visually that the texture sticking problem has been solved. The resulting motion is much more natural, better sustaining an illusion that there is a coherent 3D scene being imaged.

\vspace{-0.25\baselineskip}
\paragraph{Ablations and comparisons}

In Section~\ref{sec:practice_removals} we disabled a number of StyleGAN2 features.
We can now turn them on one by one to gauge their effect on our generators (Figure~\ref{fig:ResultTables}, right).
While mixing regularization can be re-enabled without any ill effects, we also find that styles can be mixed quite reliably even without this explicit regularization (\refappResults{}).
Re-enabling noise inputs or relying on StyleGAN2's original layer specifications compromises equivariances significantly, and using fixed Fourier features or re-enabling path length regularization harms FID.
Path length regularization is in principle at odds with translation equivariance, as it penalizes image changes upon latent space walk and thus encourages texture sticking.
We suspect that the counterintuitive improvement in equivariance may come from slightly blurrier generated images, at a cost of poor FID.

In a scaling test we tried changing the number of feature maps, observing that equivariances remain at a high level, but FID suffers considerably when the capacity is halved. Doubling the capacity improves result quality in terms of FID, at the cost of almost 4$\times$ training time.
Finally, we consider alternatives for our windowed Kaiser filter. 
Lanczos is competitive in terms of FID, but as a separable filter it compromises rotation equivariance in particular.
Gaussian leads to clearly worse FIDs.

We compare \FINAL{StyleGAN3-R} to an alternative where the rotation part is implemented using $p4$ symmetric G-CNN \cite{Cohen2016,Dey2021} on top of our \FINAL{StyleGAN3-T}. This approach provides only modest rotation equivariance while being slower to train.
Steerable filters \cite{Weiler2017} could theoretically provide competitive EQ-R, but the memory and training time requirements proved infeasible with generator networks of this size.

\FINAL{\refappResults{} demonstrates that the spectral properties of generated images closely match training data, comparing favorably to several earlier architectures. %
}

\figResults{fig:results} %

\vspace{-0.4\baselineskip}
\paragraph{Internal representations}
Figure~\ref{fig:results} visualizes typical internal representations from the networks. 
While in StyleGAN2 all feature maps seem to encode signal magnitudes, in our networks some of the maps take a different role and encode phase information instead.
Clearly this is something that is needed when the network synthesizes detail \emph{on the surfaces}; it needs to invent a coordinate system. 
In \FINAL{StyleGAN3-R}, the emergent positional encoding patterns appear to be somewhat more well-defined.
We believe that the existence of a coordinate system that allows precise localization on the surfaces of objects will prove useful in various applications, including advanced image and video editing. 

\vspace{-0.4\baselineskip}
\section{Limitations, discussion, and future work}
\label{sec:limitations}
\vspace{-0.4\baselineskip}

In this work we modified only the generator, but it seems likely that further benefits would be available by making the discriminator equivariant as well.
For example, in our FFHQ results the teeth do not move correctly when the head turns, and we suspect that this is caused by the discriminator accidentally preferring to see the front teeth at certain pixel locations.
\FINAL{Concurrent work has identified that aliasing is detrimental for such generalization \cite{Vasconcelos2021}}.

\FINAL{Our alias-free generator architecture contains implicit assumptions about the nature of the training data, and violating these may cause training difficulties. Let us consider an example. Suppose we have black-and-white cartoons as training data that we (incorrectly) pre-process using point sampling \cite{Parmar2021}, leading to training images where almost all pixels are either black or white and the edges are jagged. This kind of badly aliased training data is difficult for GANs in general, but it is especially at odds with equivariance: on the one hand, we are asking the generator to be able to translate the output smoothly by subpixel amounts, but on the other hand, edges must still remain jagged and pixels only black/white, to remain faithful to the training data. 
The same issue can also arise with letterboxing of training images, low-quality JPEGs, or retro pixel graphics, where the jagged stair-step edges are a defining feature of the aesthetic. In such cases it may be beneficial for the generator to be aware of the pixel grid.
}

In future, it might be interesting to re-introduce noise inputs (stochastic variation) in a way \FINAL{that is consistent with hierarchical synthesis}. 
A better path length regularization would encourage neighboring features to move together, not discourage them from moving at all.
It might be beneficial to try to extend our approach to equivariance w.r.t.~scaling, anisotropic scaling, or even arbitrary homeomorphisms.
Finally, it is well known that antialiasing should be done before tone mapping. So far, all GANs\,---\,including ours\,---\,have operated in the sRGB color space (after tone mapping).

\FINAL{Attention layers in the middle of a generator~\cite{Zhang2018sagan} could likely be dealt with similarly to non-linearities by temporarily switching to higher resolution -- although the time complexity of attention layers may make this somewhat challenging in practice. Recent attention-based GANs that start with a tokenizing transformer (e.g., VQGAN~\cite{Esser2020}) may be at odds with equivariance. Whether it is possible to make them equivariant is an important open question.}

\vspace{-0.5\baselineskip}
\paragraph{Potential negative societal impacts} of (image-producing) GANs include many forms of disinformation, from fake portraits in social media \cite{Hill2020} to propaganda videos of world leaders \cite{Seymour2019}. Our contribution eliminates certain characteristic artifacts from videos, potentially making them more convincing or deceiving, depending on the application.
Viable solutions include model watermarking \cite{Skripniuk2020} along with large-scale authenticity assessment in major social media sites. 
\FINAL{This entire project} consumed 92 GPU years and 225 MWh of electricity on an in-house cluster of NVIDIA V100s. The new \FINAL{StyleGAN3} generator is only marginally costlier to train or use than that of StyleGAN2.

\vspace{-1.25mm}%
\section{Acknowledgments} %
\vspace{-1.25mm}%
We thank David Luebke, Ming-Yu Liu, Koki Nagano, Tuomas Kynk\"a\"anniemi, and Timo Viitanen for reviewing early drafts and helpful suggestions.
Fr\'{e}do Durand for early discussions.
Tero Kuosmanen for maintaining our compute infrastructure. 
AFHQ authors for an updated version of their dataset. 
Getty Images for the training images in the \textsc{Beaches} dataset.
We did not receive external funding or additional revenues for this project.

{\small
	\bibliographystyle{ieee}
	\bibliography{paper}
}

\ifappendix
  \newpage
  \appendix
  {\LARGE\bf Appendices}
	\newcommand{\refpaper}[1]{\ref{#1}}
  \newcommand{\figUncuratedFFHQU}[1]{
\begin{figure}[p]
\footnotesize%
\renewcommand{\h}{0.485\linewidth}%
\makebox[\h][c]{Real images from the training set}\hfill%
\makebox[\h][c]{\FINAL{StyleGAN2}, FID 3.79}%
\vspace{0mm}\\
\includegraphics[width=\h]{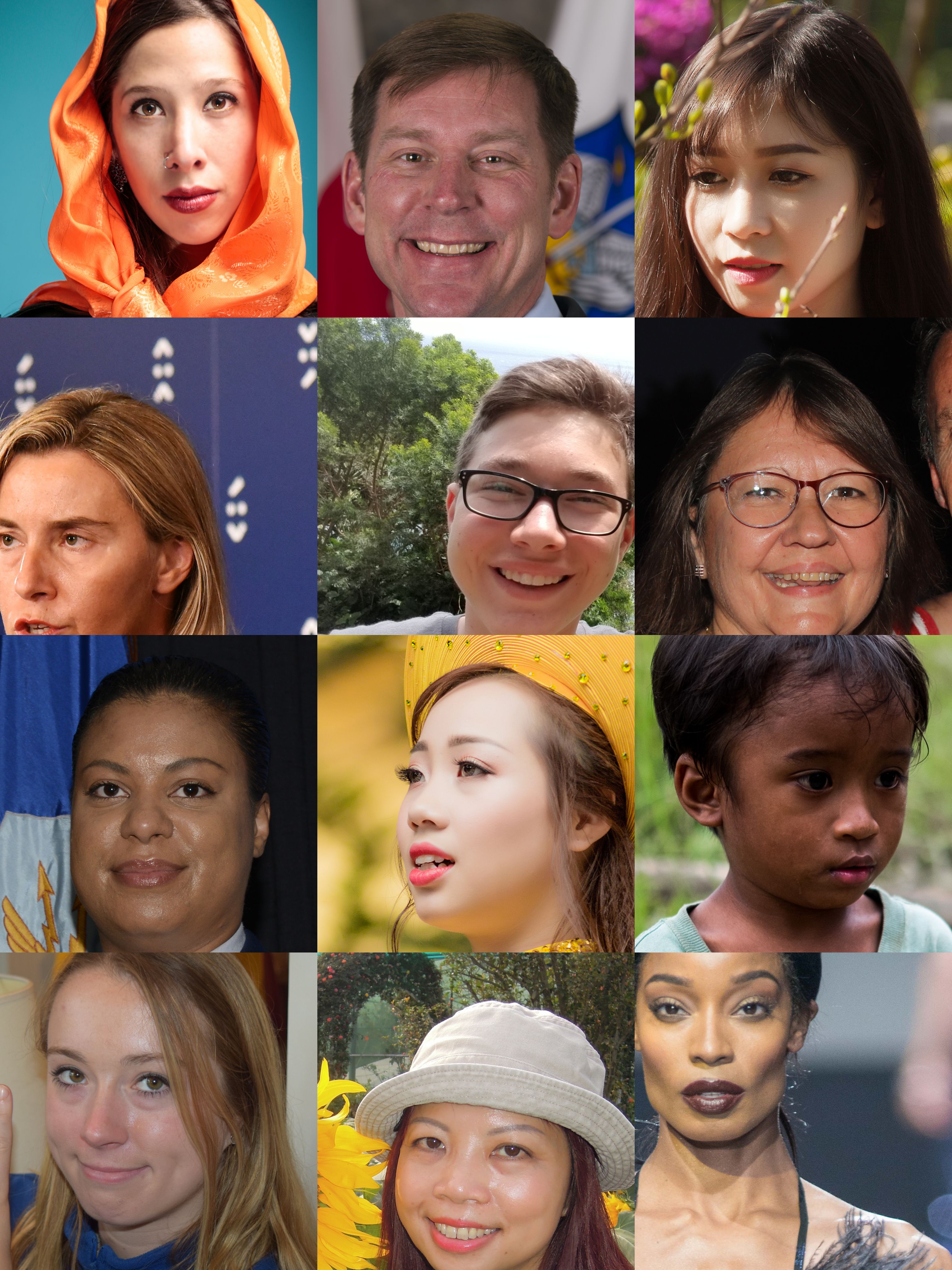}\hfill%
\includegraphics[width=\h]{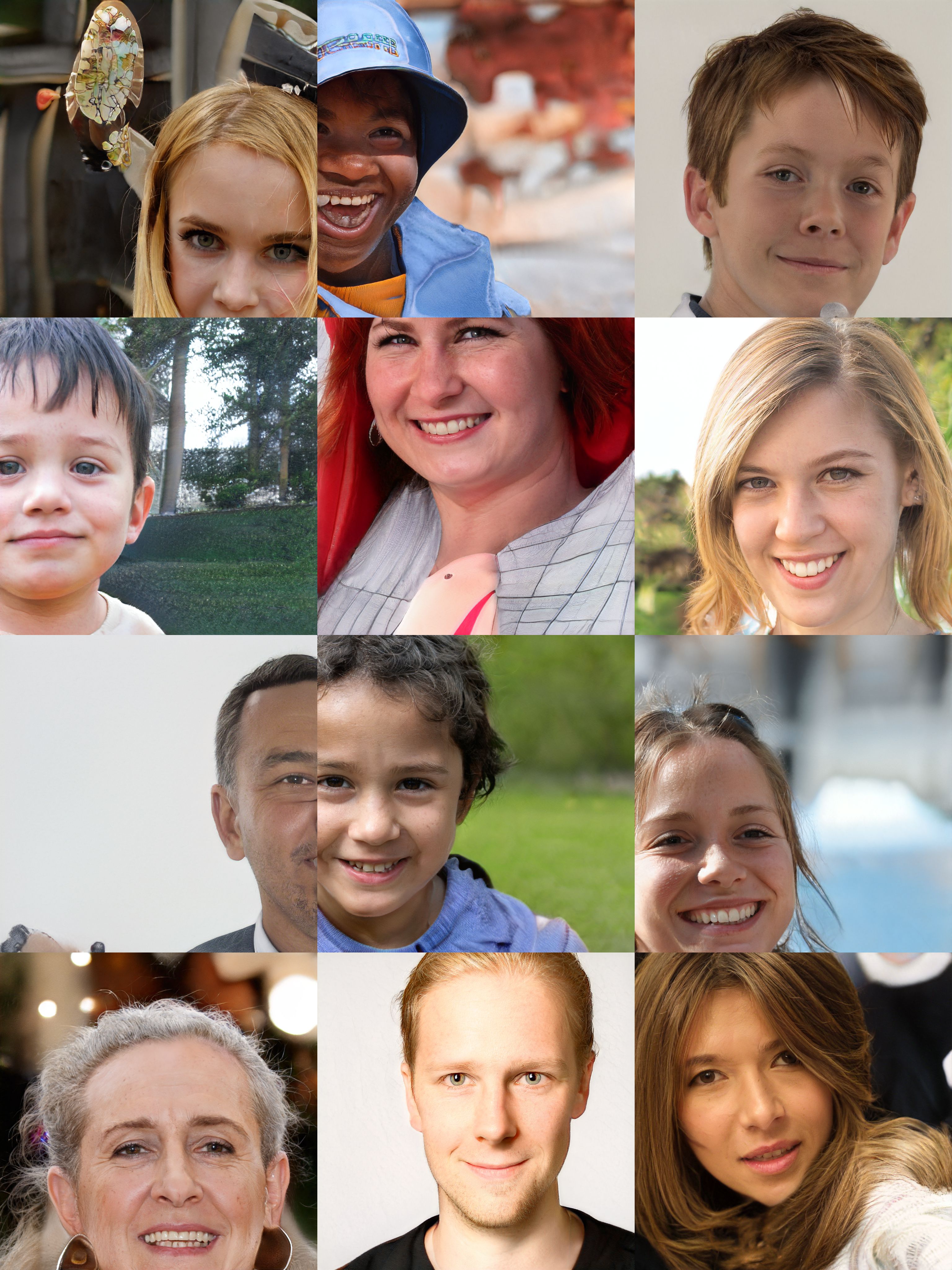}%
\vspace{1.1mm}\\
\makebox[\h][c]{\FINAL{StyleGAN3-T (ours)}, FID 3.67}\hfill%
\makebox[\h][c]{\FINAL{StyleGAN3-R (ours)}, FID 3.66}%
\vspace{0mm}\\
\includegraphics[width=\h]{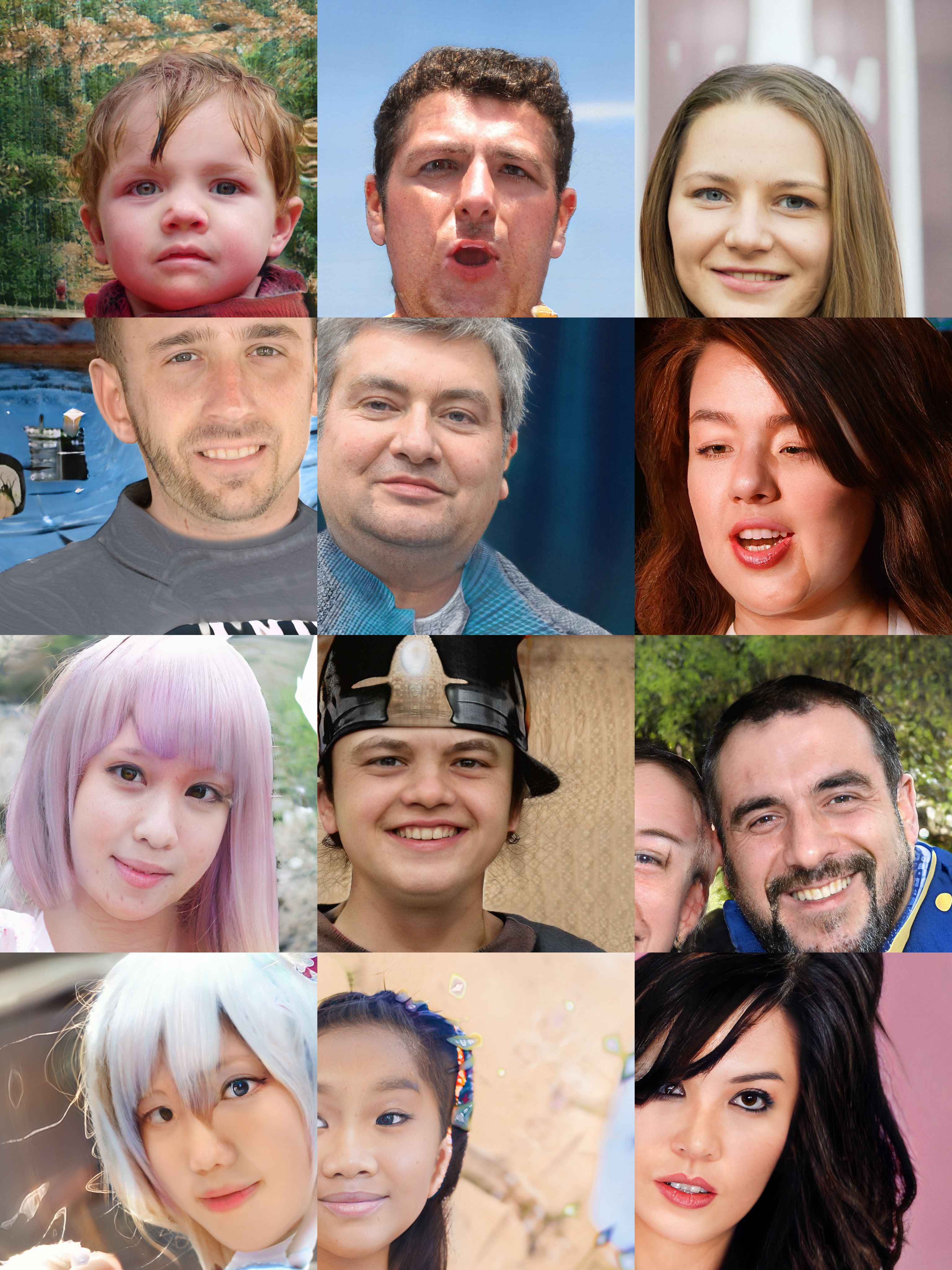}\hfill%
\includegraphics[width=\h]{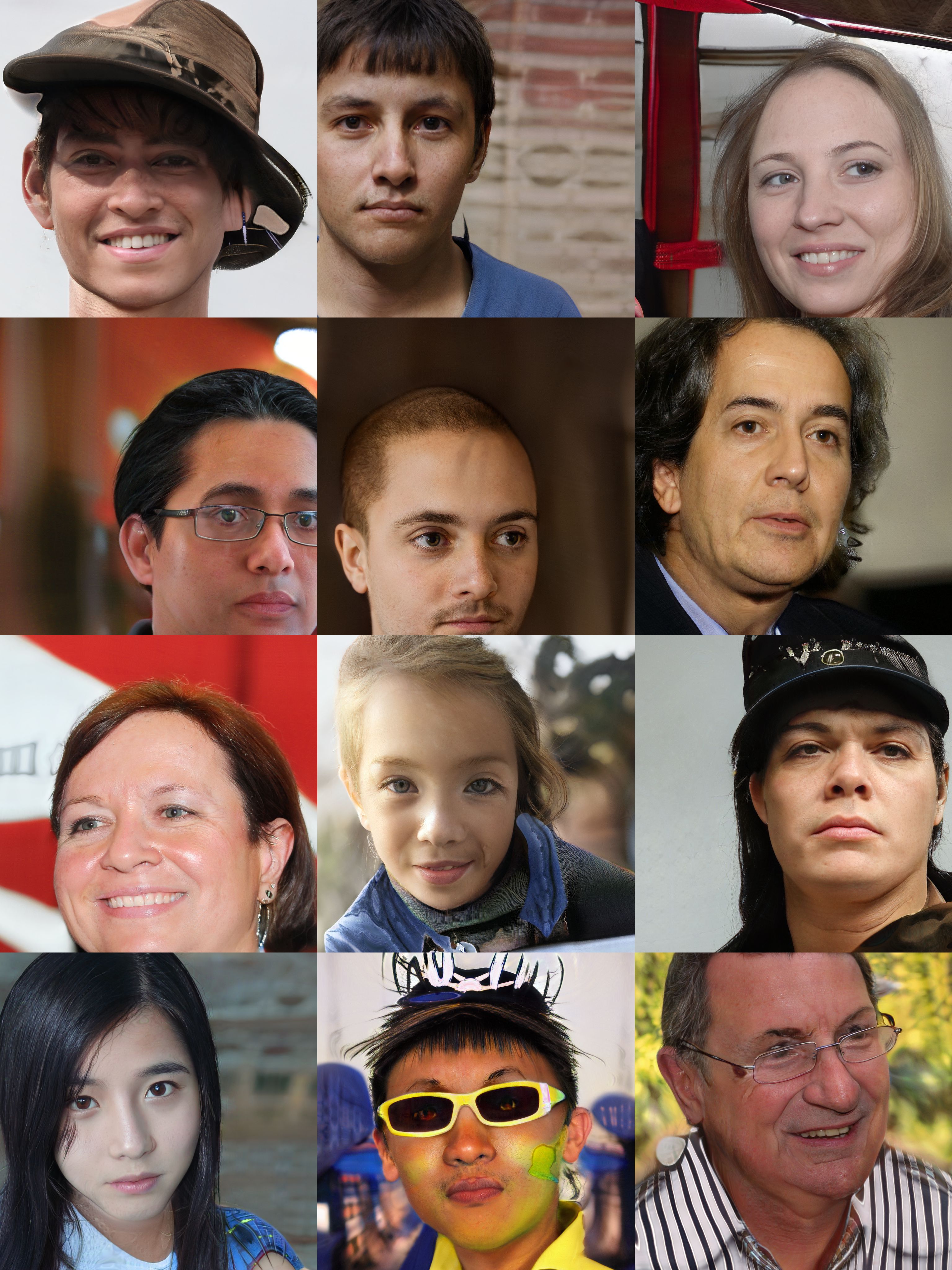}%
\caption{
Uncurated samples for unaligned FFHQ (FFHQ-U). Truncation was not used.
}
\label{#1}
\end{figure}
}

\newcommand{\figUncuratedMetFacesU}[1]{
\begin{figure}[p]
\footnotesize%
\renewcommand{\h}{0.485\linewidth}%
\makebox[\h][c]{Real images from the training set}\hfill%
\makebox[\h][c]{\FINAL{StyleGAN2}, FID 18.98}%
\vspace{0mm}\\
\includegraphics[width=\h]{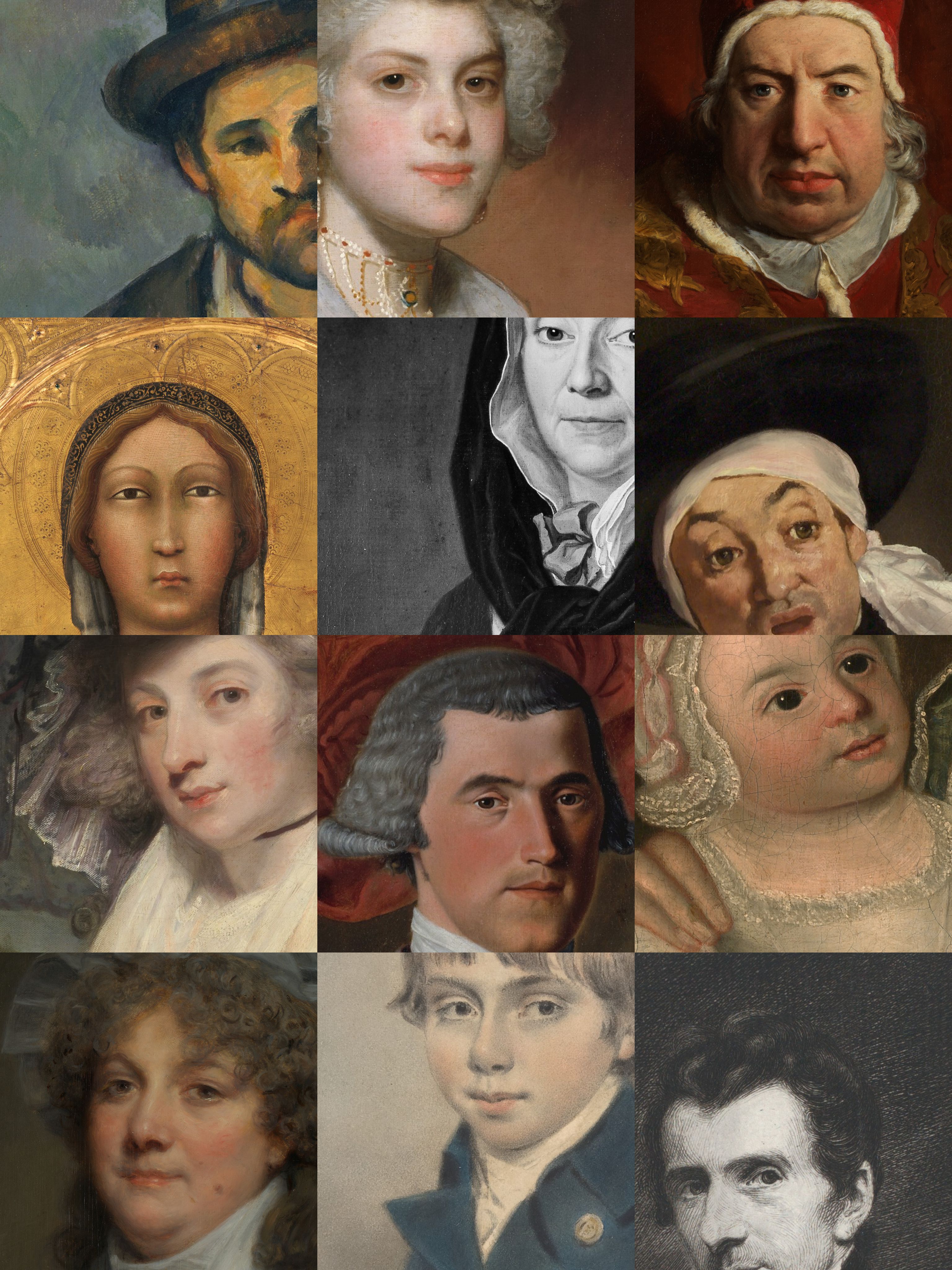}\hfill%
\includegraphics[width=\h]{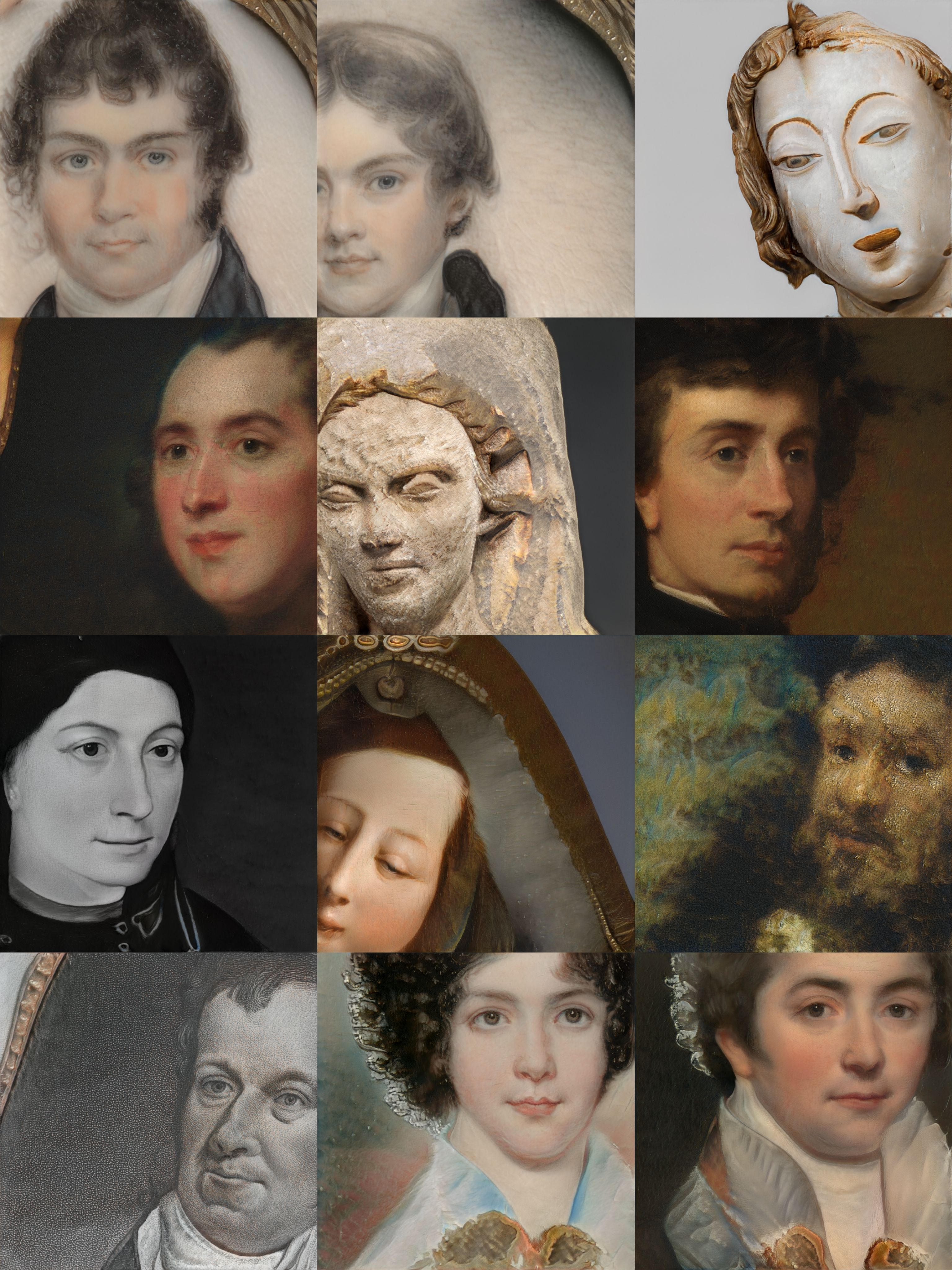}%
\vspace{1.1mm}\\
\makebox[\h][c]{\FINAL{StyleGAN3-T (ours)}, FID 18.75}\hfill%
\makebox[\h][c]{\FINAL{StyleGAN3-R (ours)}, FID 18.75}%
\vspace{0mm}\\
\includegraphics[width=\h]{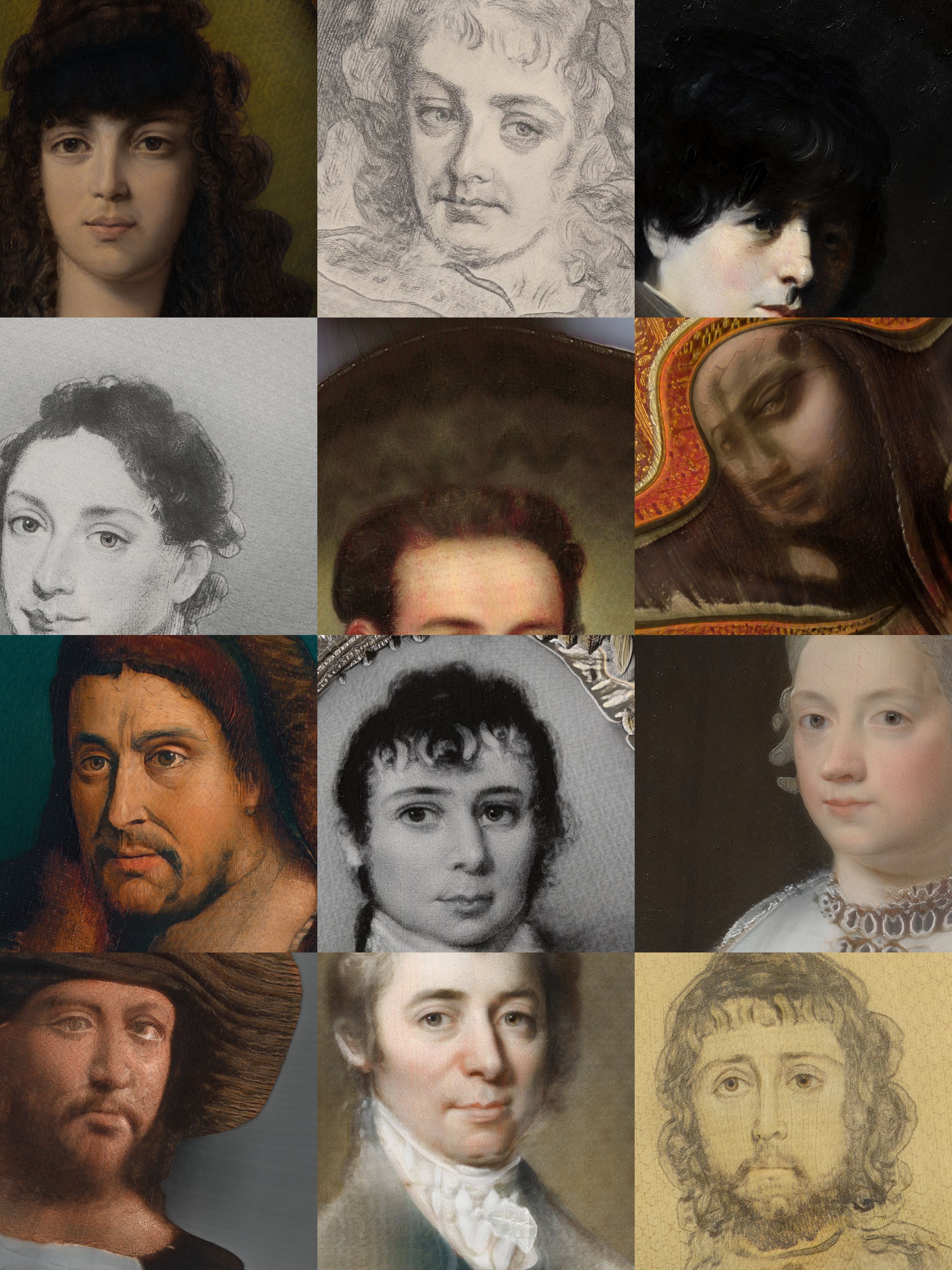}\hfill%
\includegraphics[width=\h]{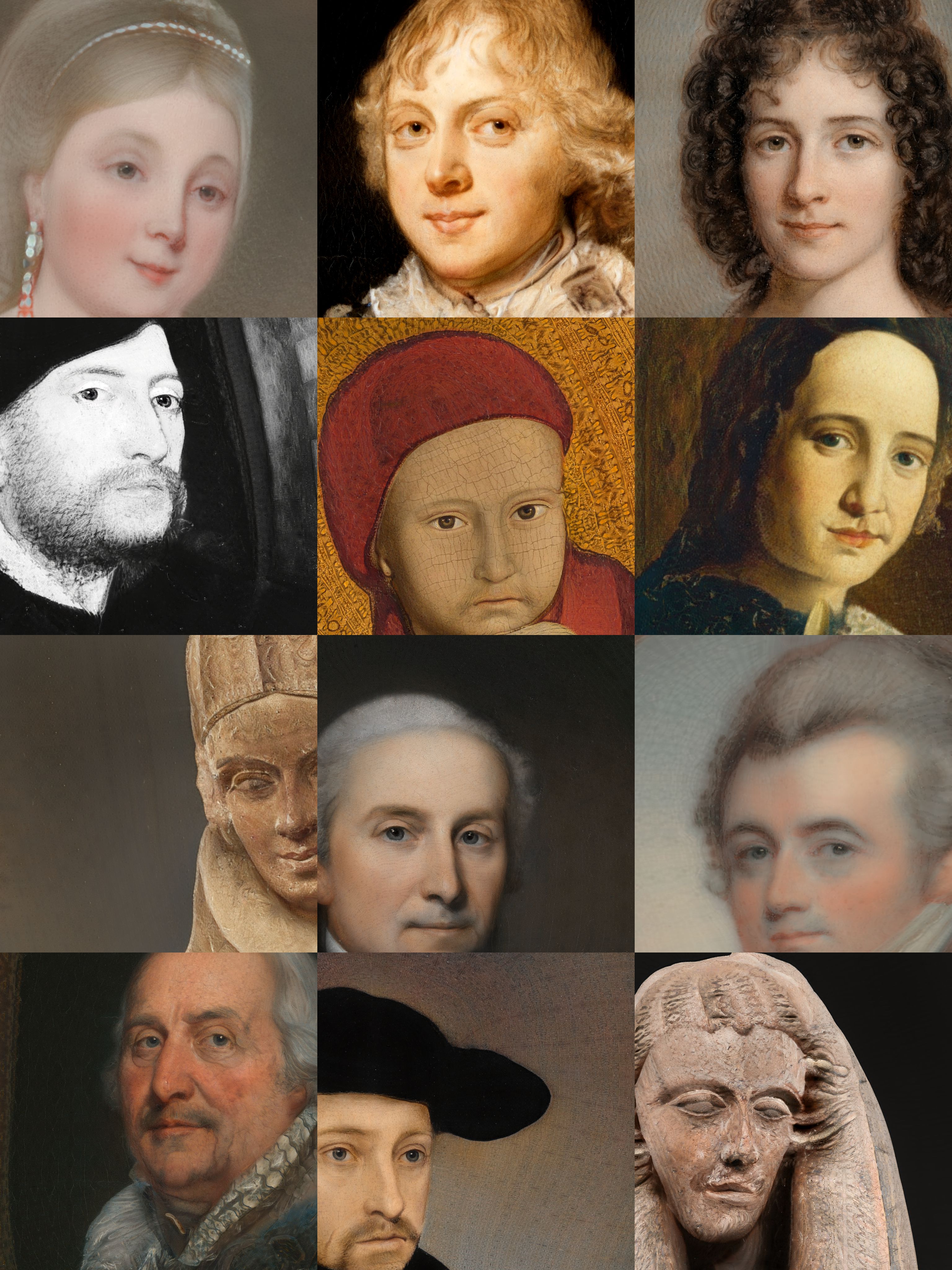}%
\caption{
Uncurated samples for unaligned \textsc{MetFaces} (\textsc{Metfaces-U}). Truncation was not used.
}
\label{#1}
\end{figure}
}

\newcommand{\figUncuratedAFHQtwo}[1]{
\begin{figure}[p]
\footnotesize%
\renewcommand{\h}{0.485\linewidth}%
\makebox[\h][c]{Real images from the training set}\hfill%
\makebox[\h][c]{\FINAL{StyleGAN2}, FID 4.62}%
\vspace{0mm}\\
\includegraphics[width=\h]{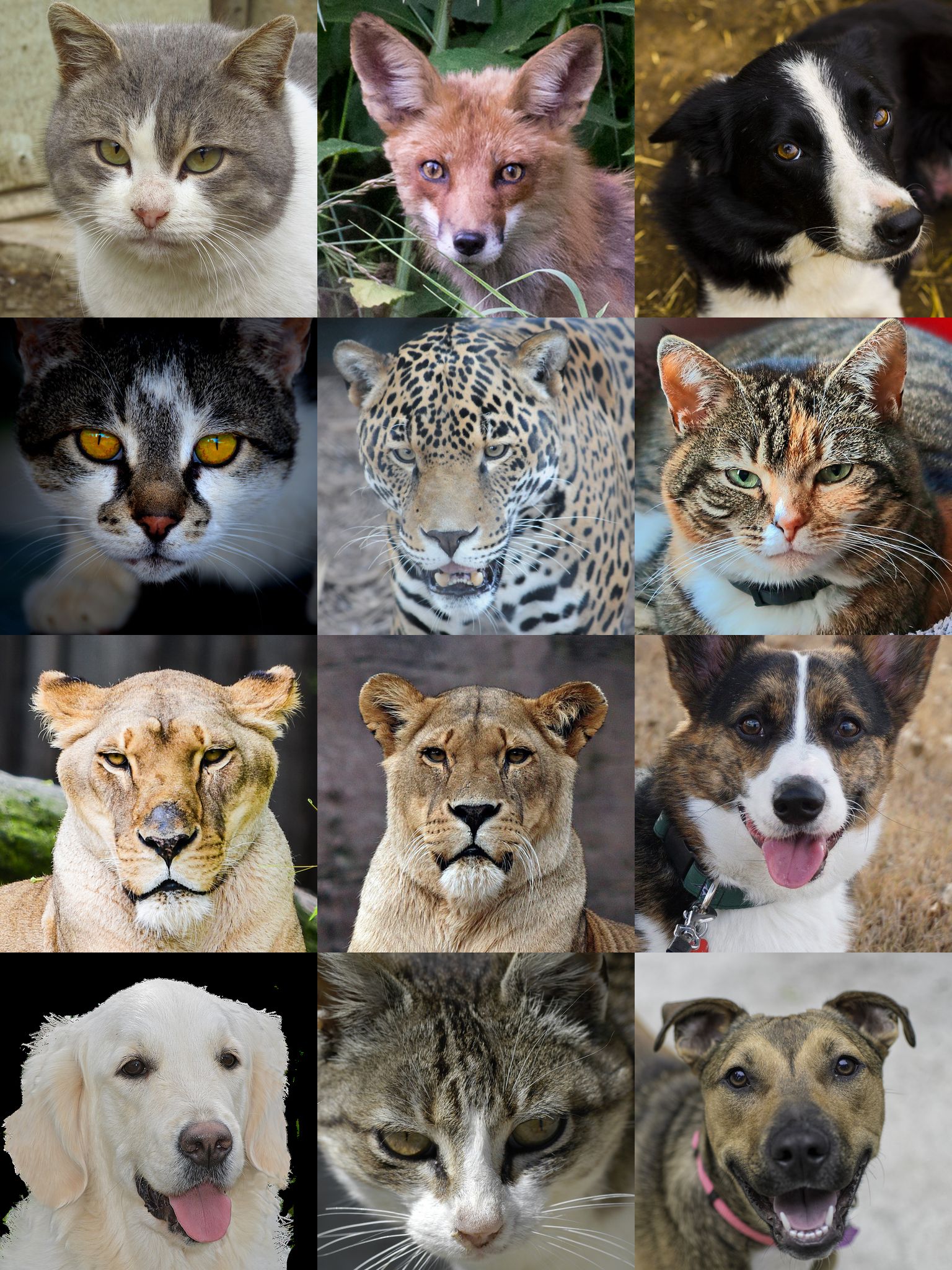}\hfill%
\includegraphics[width=\h]{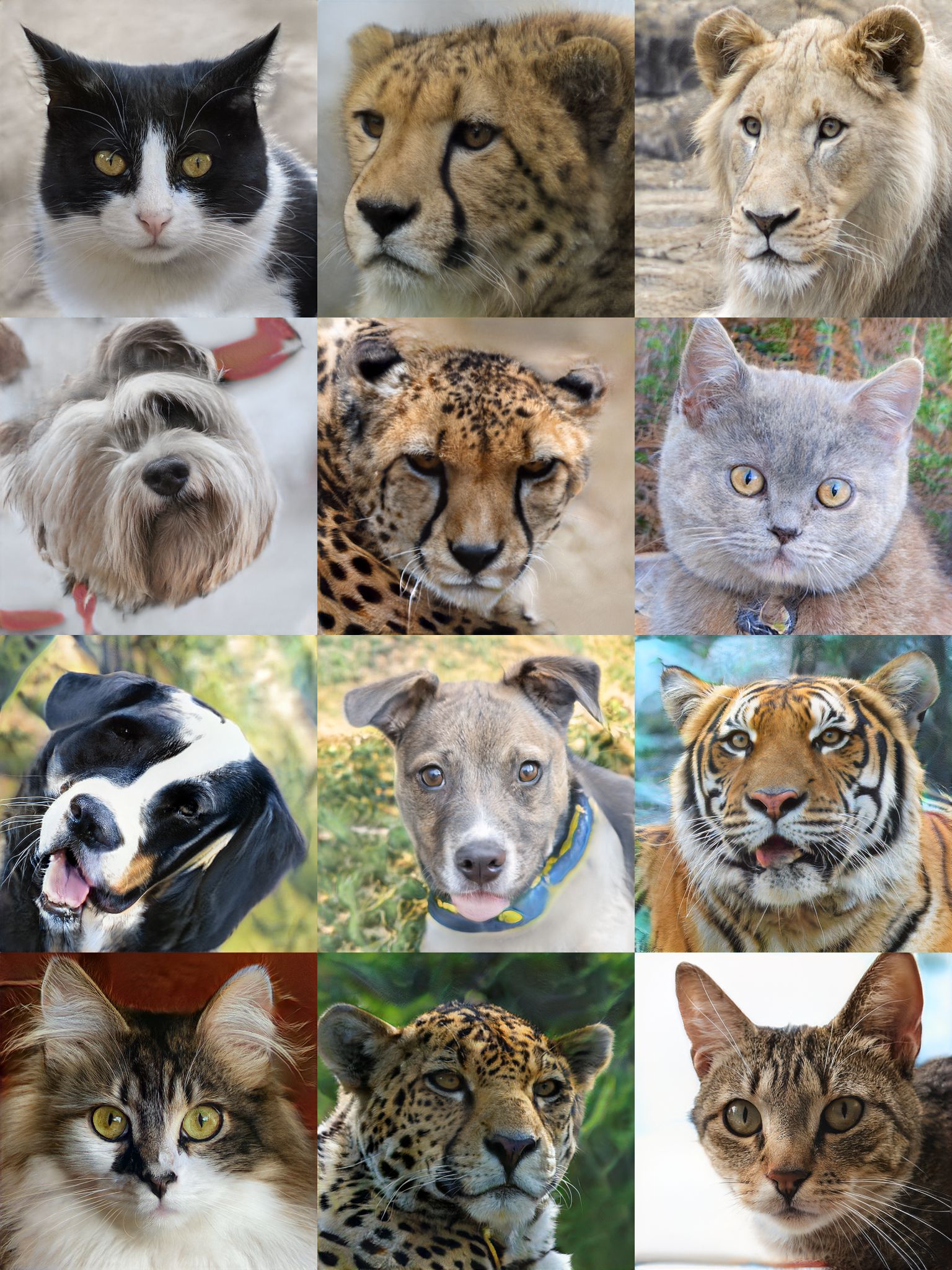}%
\vspace{1.1mm}\\
\makebox[\h][c]{\FINAL{StyleGAN3-T (ours)}, FID 4.04}\hfill%
\makebox[\h][c]{\FINAL{StyleGAN3-R (ours)}, FID 4.40}%
\vspace{0mm}\\
\includegraphics[width=\h]{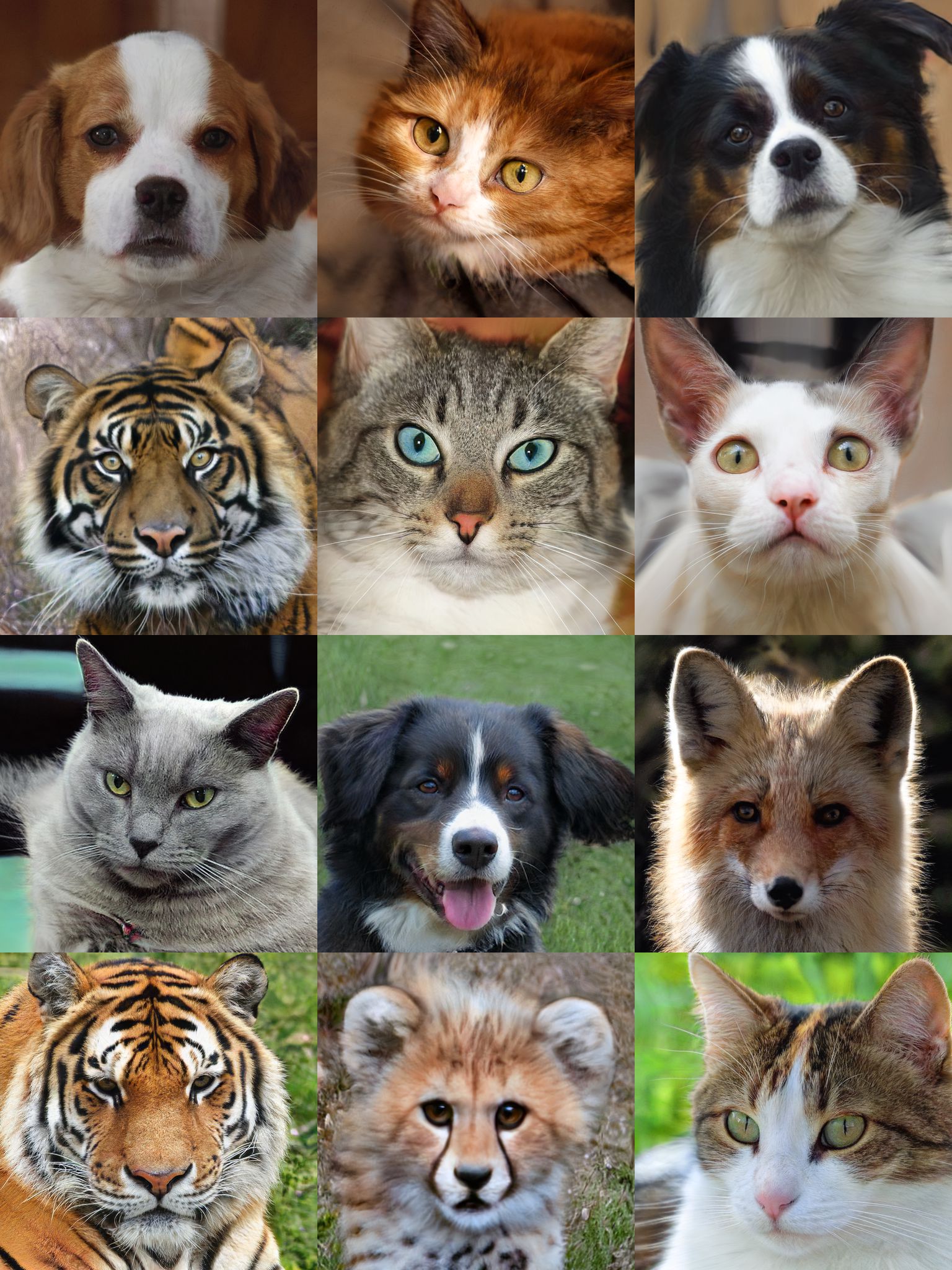}\hfill%
\includegraphics[width=\h]{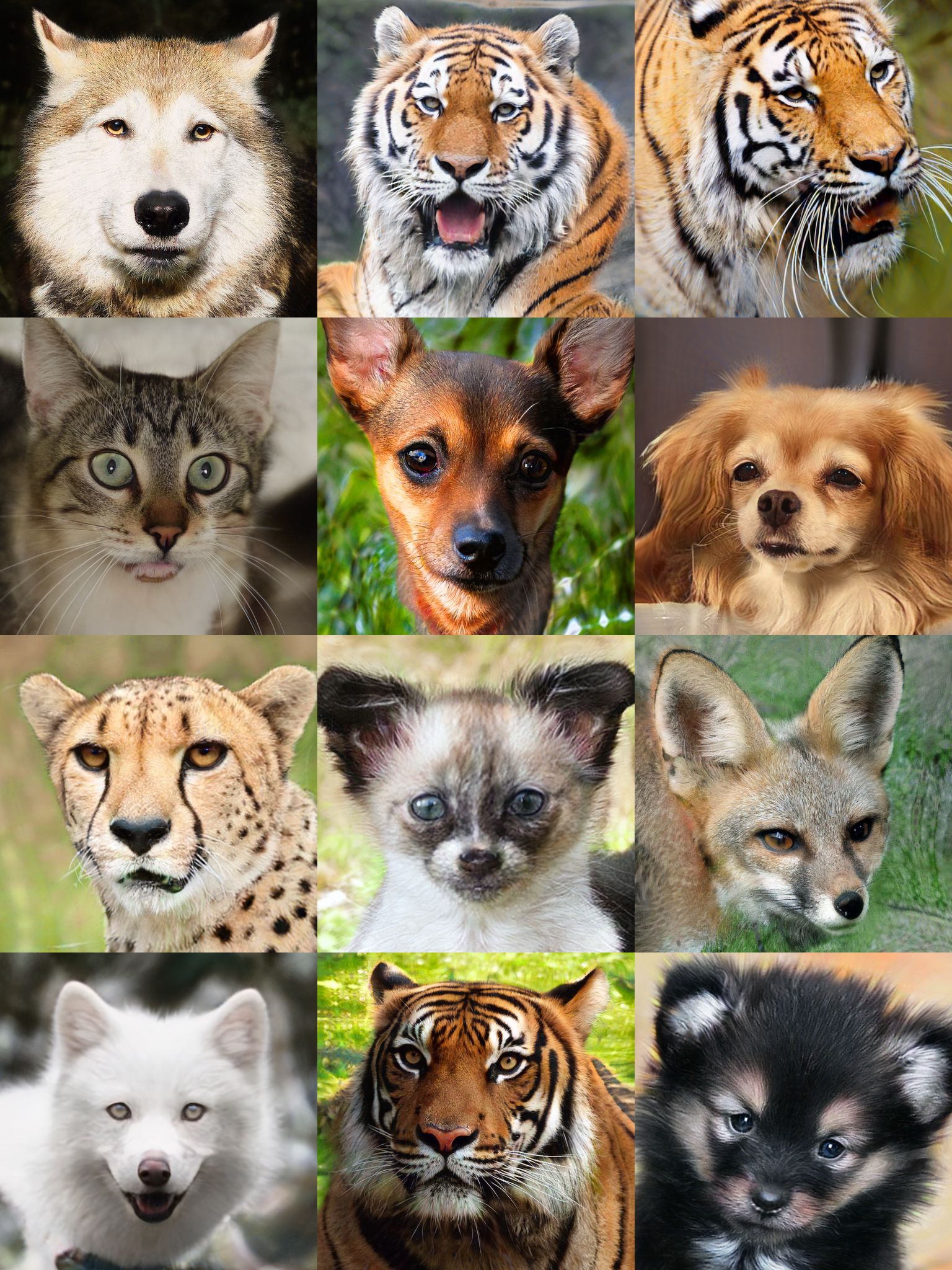}%
\caption{
Uncurated samples for \textsc{AFHQv2}. Truncation was not used.
}
\label{#1}
\end{figure}
}

\newcommand{\figUncuratedBeaches}[1]{
\begin{figure}[p]
\footnotesize%
\renewcommand{\h}{0.485\linewidth}%
\makebox[\h][c]{Real images from the training set}\hfill%
\makebox[\h][c]{\FINAL{StyleGAN2}, FID 5.03}%
\vspace{0mm}\\
\includegraphics[width=\h]{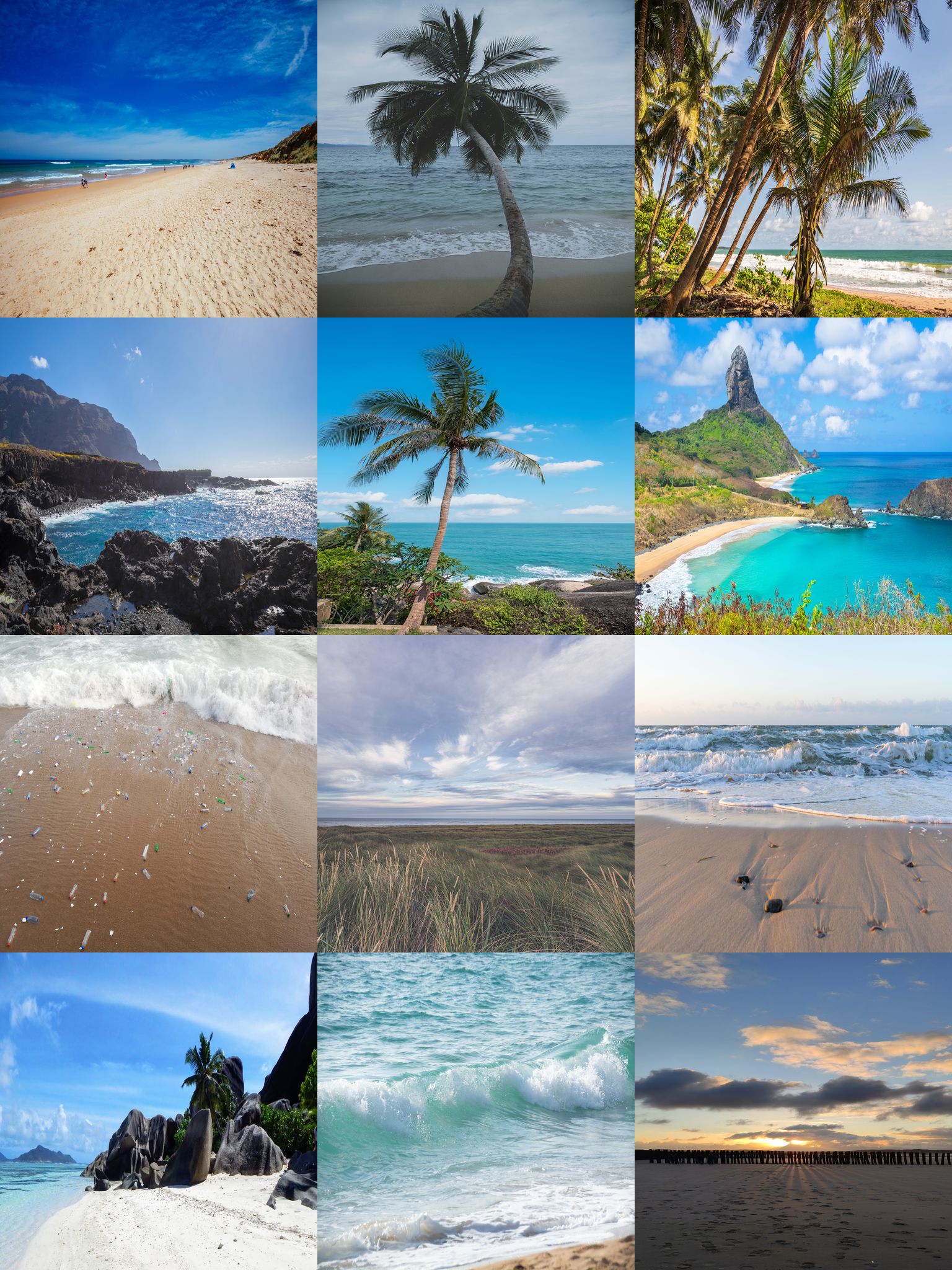}\hfill%
\includegraphics[width=\h]{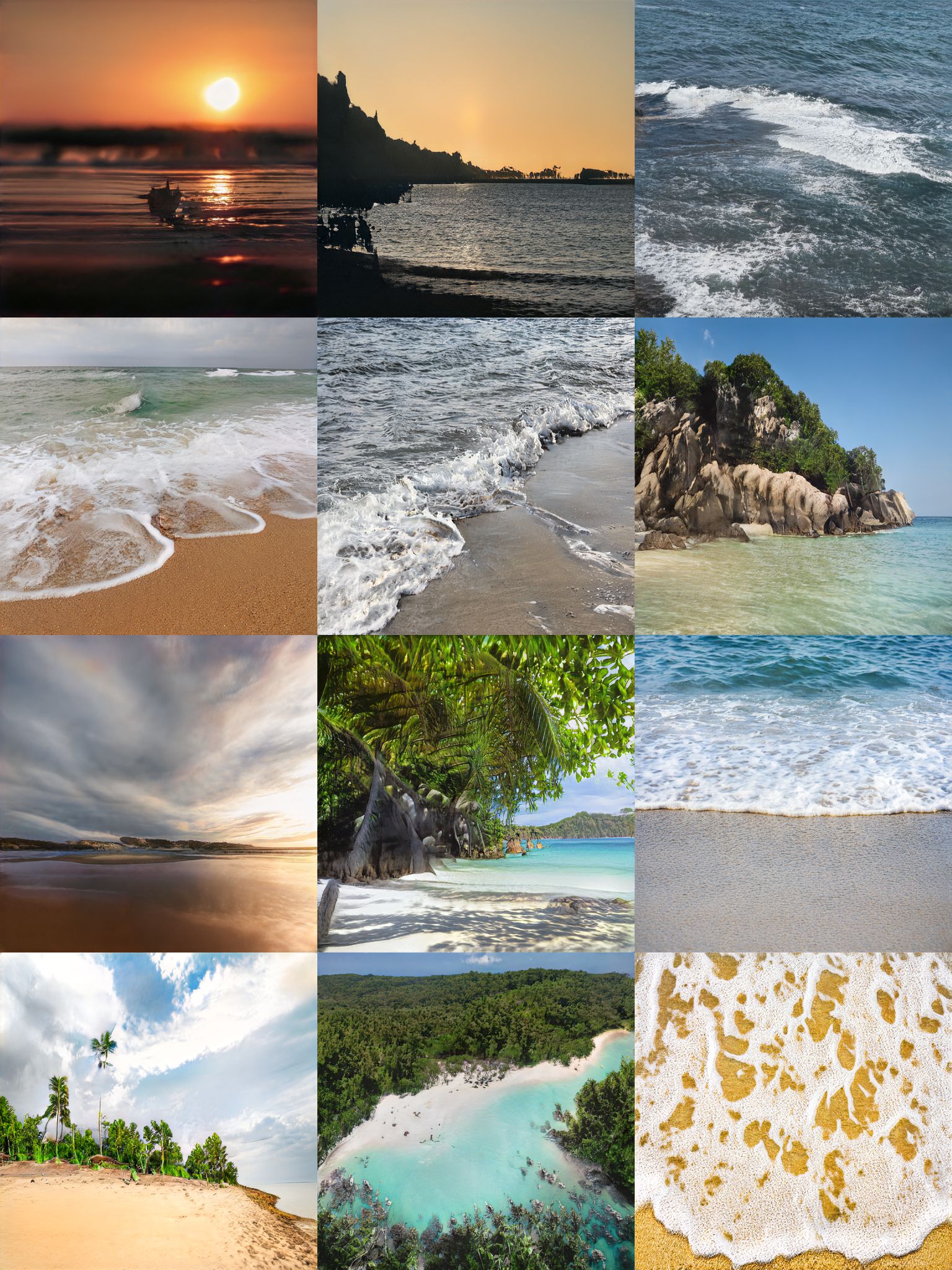}%
\vspace{1.1mm}\\
\makebox[\h][c]{\FINAL{StyleGAN3-T (ours)}, FID 4.32}\hfill%
\makebox[\h][c]{\FINAL{StyleGAN3-R (ours)}, FID 4.57}%
\vspace{0mm}\\
\includegraphics[width=\h]{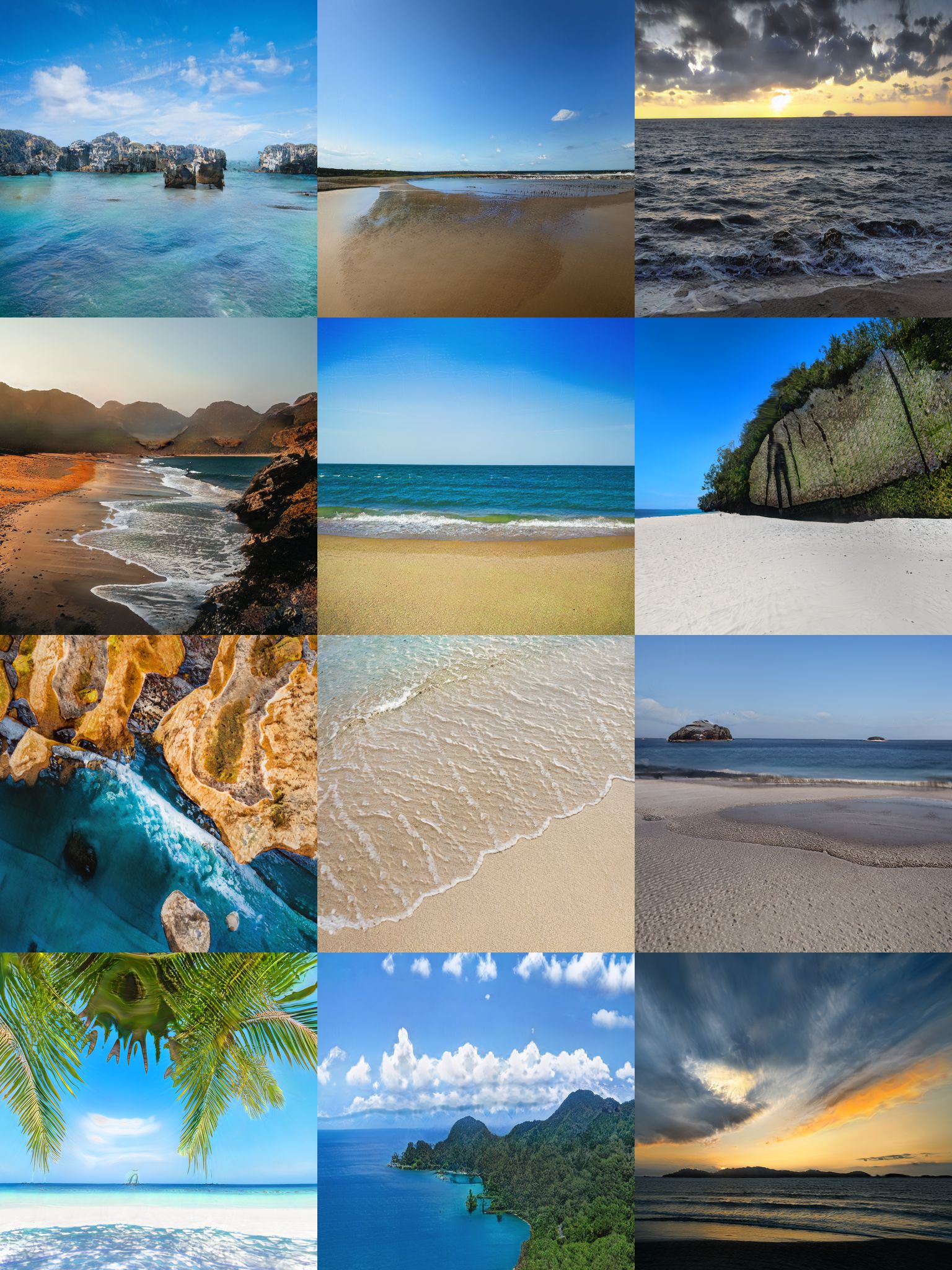}\hfill%
\includegraphics[width=\h]{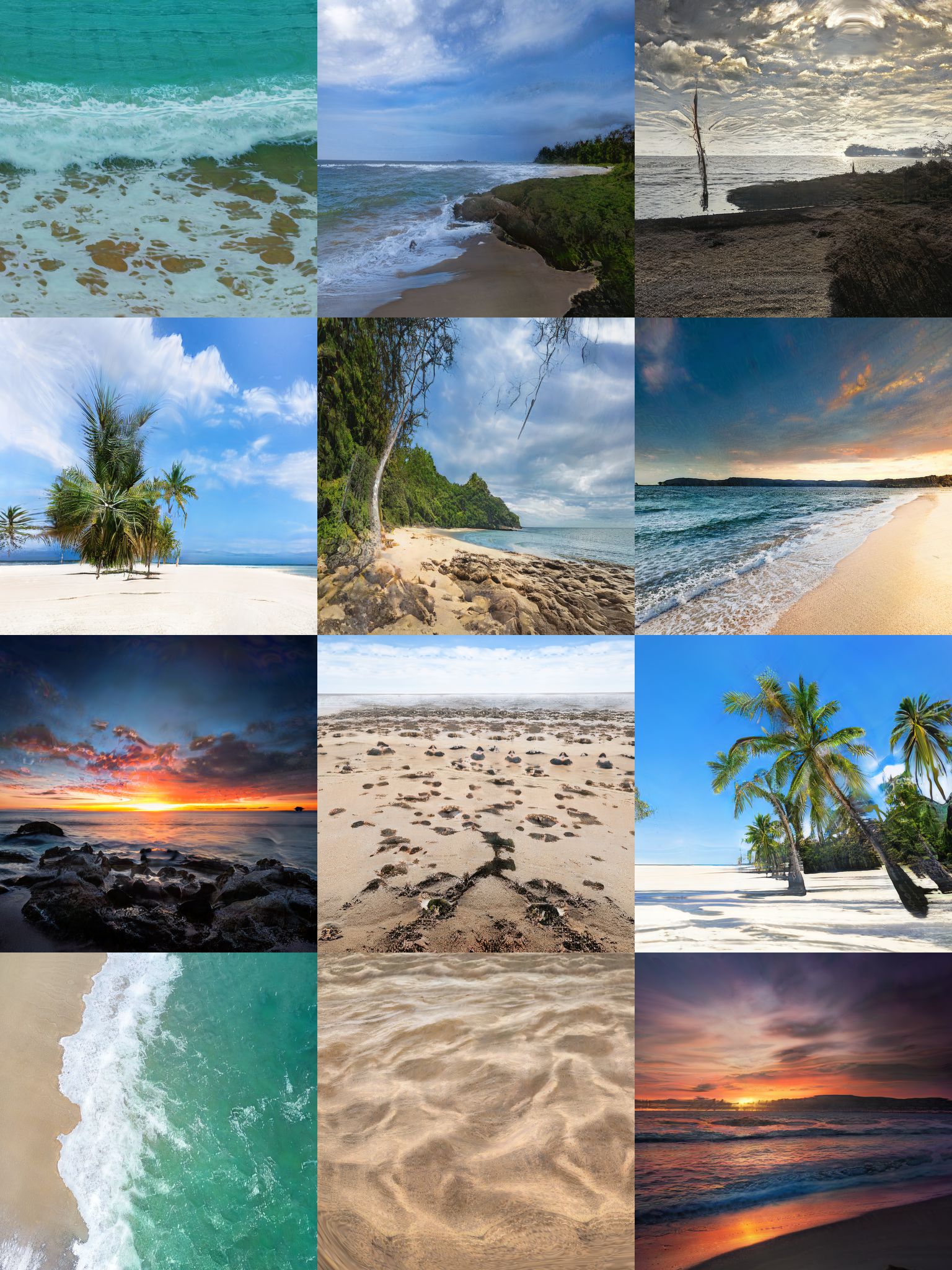}%
\caption{
Uncurated samples for \textsc{Beaches}. Truncation was not used.
}
\label{#1}
\end{figure}
}

\newcommand{\figStyleMixing}[1]{
\begin{figure}[t]
\includegraphics[width=\linewidth]{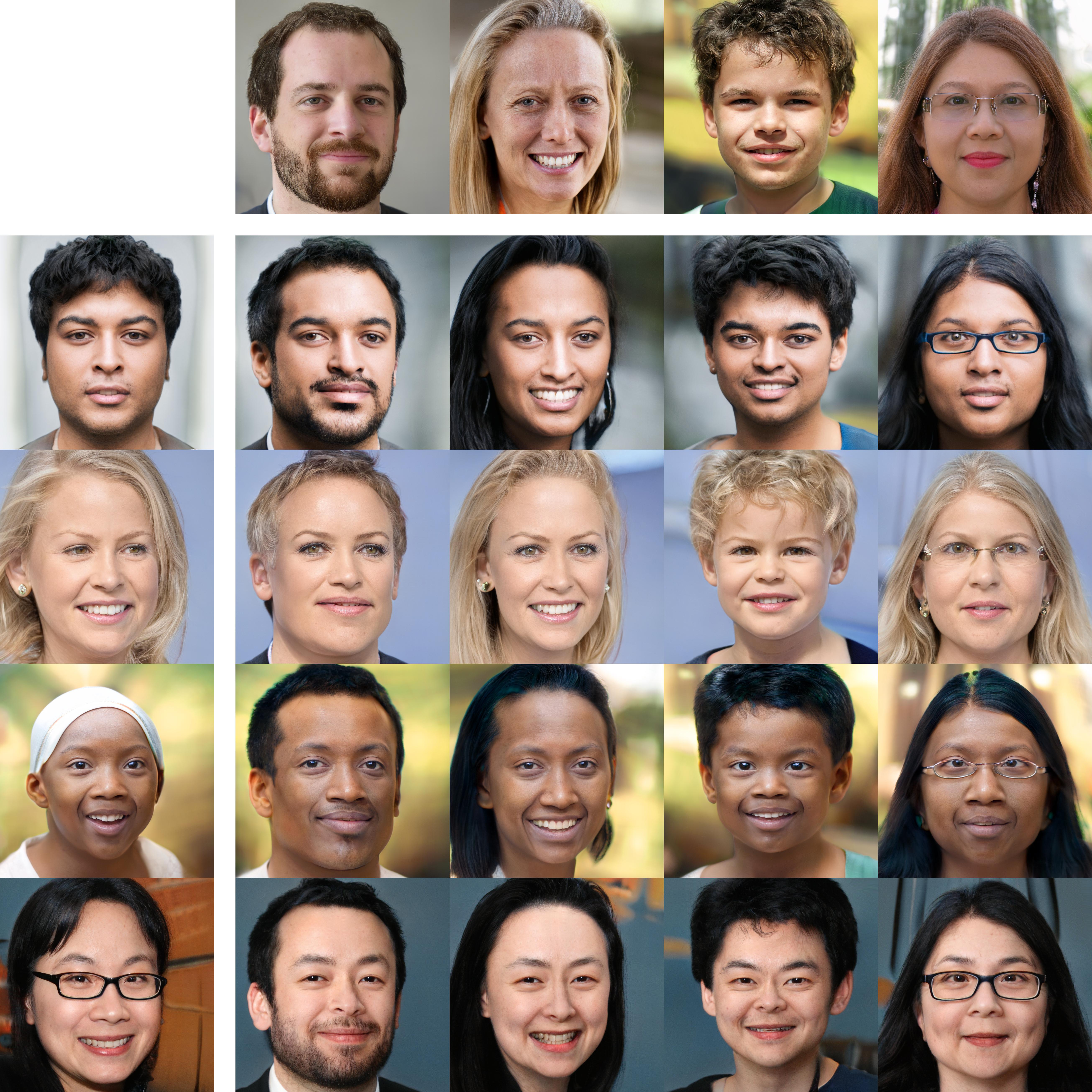}%
\caption{
Hand-picked style mixing examples where the coarse (0--6) and fine (7--14) layers use a different latent code. 
Mixing regularization was not used during training. 
Head pose, coarse facial shape, hair length and glasses seem to get inherited from the coarse layers (top row), while coloring and finer facial features are mostly inherited from the fine layers (leftmost column). 
The control is not quite perfect: e.g.,~feminine/masculine features are not reliably copied from exactly one of the sources. 
Moving the fine/coarse boundary fixes this particular issue, but other similar problems persist. 
}
\label{#1}
\end{figure}
}

\newcommand{\figConvergence}[1]{
\begin{figure}[t]
\footnotesize%
\renewcommand{\h}{0.33\linewidth}%
\includegraphics[width=\h]{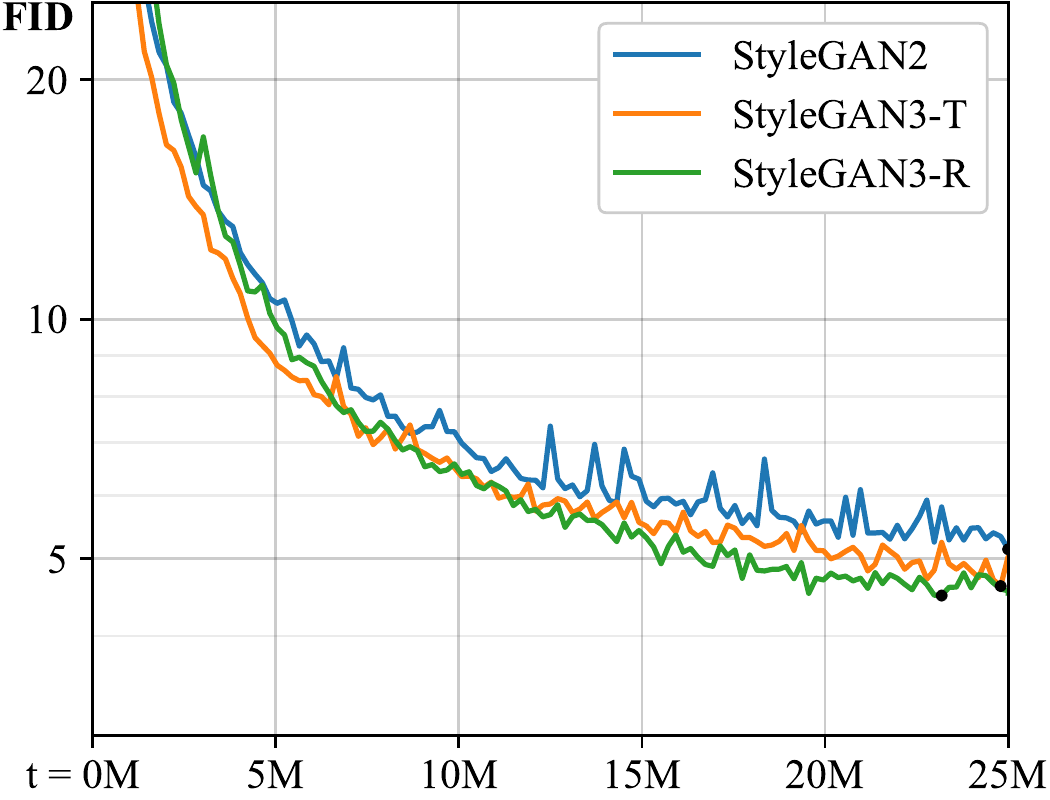}\hfill%
\includegraphics[width=\h]{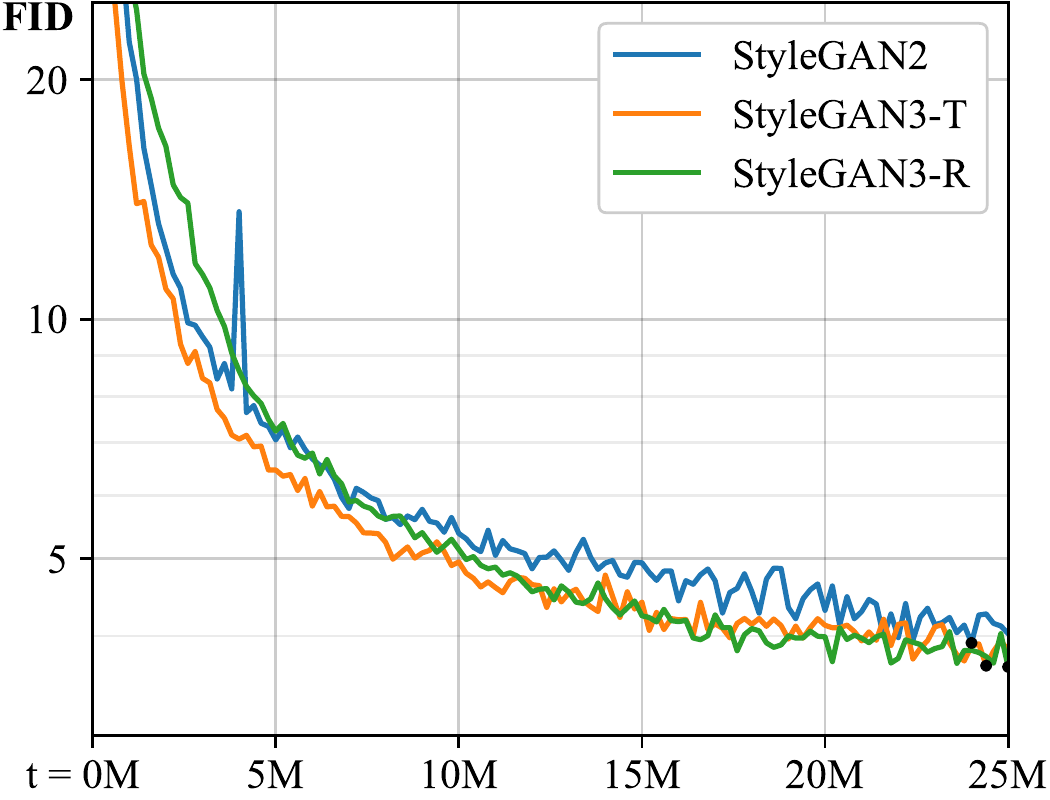}\hfill%
\includegraphics[width=\h]{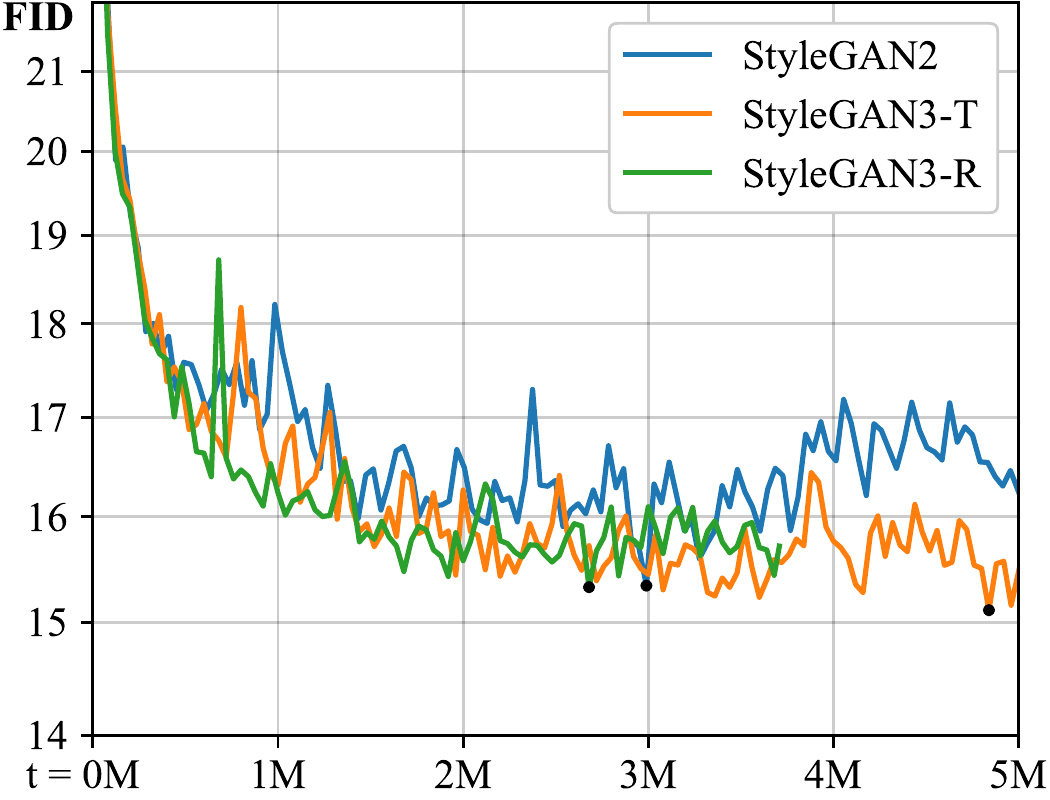}\\
\makebox[\h][c]{(a) \textsc{FFHQ-U} at 256$\times$256}\hfill%
\makebox[\h][c]{(b) \textsc{FFHQ} at 1024$\times$1024}\hfill%
\makebox[\h][c]{(c) \textsc{MetFaces} at 1024$\times$1024}%
\caption{
Training convergence with three datasets using StyleGAN2 and our main configurations (config~\textsc{t} and~\textsc{r}).
$x$-axis corresponds to the total number of real images shown to the discriminator and $y$-axis is the Fr\'echet inception distance (FID), computed between 50k generated images and all training images~\cite{Heusel2017,Karras2020}; lower is better.
The black dots indicate the best FID for each training run, matching the corresponding cases in Figures~\refpaper{fig:BridgeTables} and~\refpaper{fig:ResultTables}.
\textsc{MetFaces} was trained using adaptive discriminator augmentation (ADA)~\cite{Karras2020}, starting from the corresponding \textsc{FFHQ} snapshot with the lowest FID.
}
\label{#1}
\end{figure}
}

\newcommand{\figAverageSpectra}[1]{
\begin{figure}[t]
\footnotesize%
\renewcommand{\h}{0.25\linewidth}%
\renewcommand{\hh}{0.33\linewidth}%
\includegraphics[width=\h]{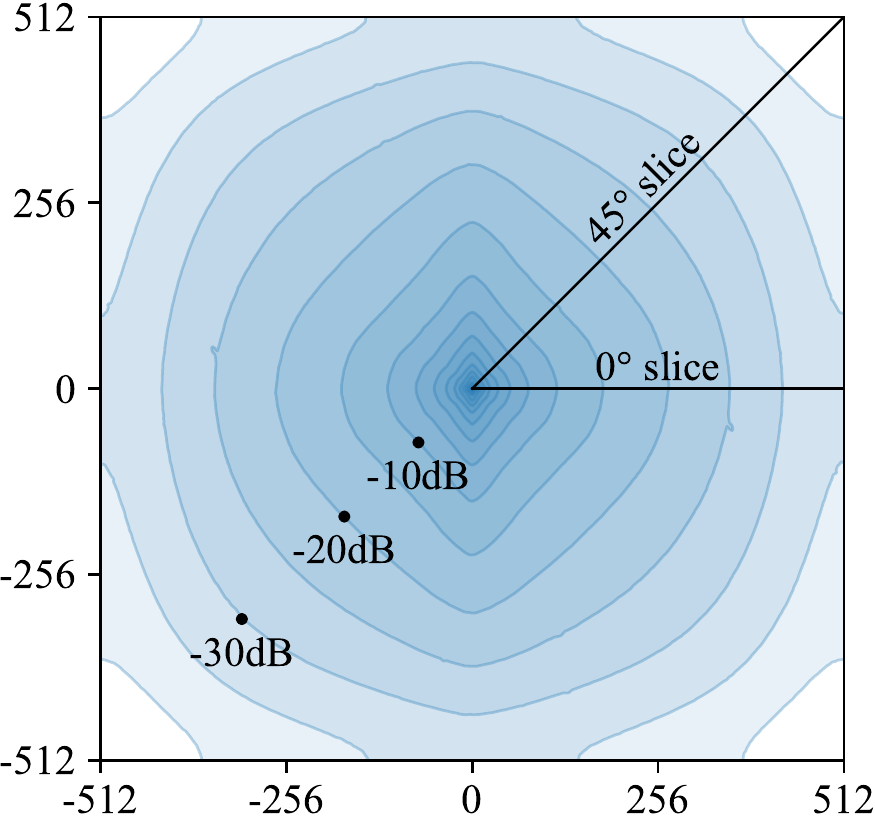}\hfill%
\includegraphics[width=\h]{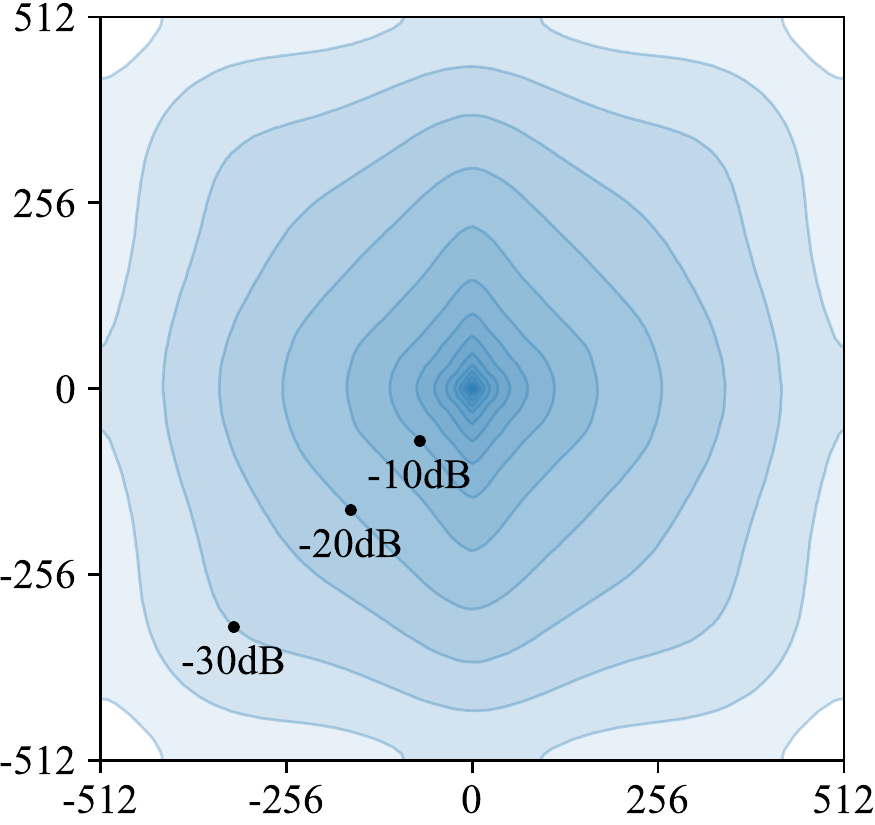}\hfill%
\includegraphics[width=\h]{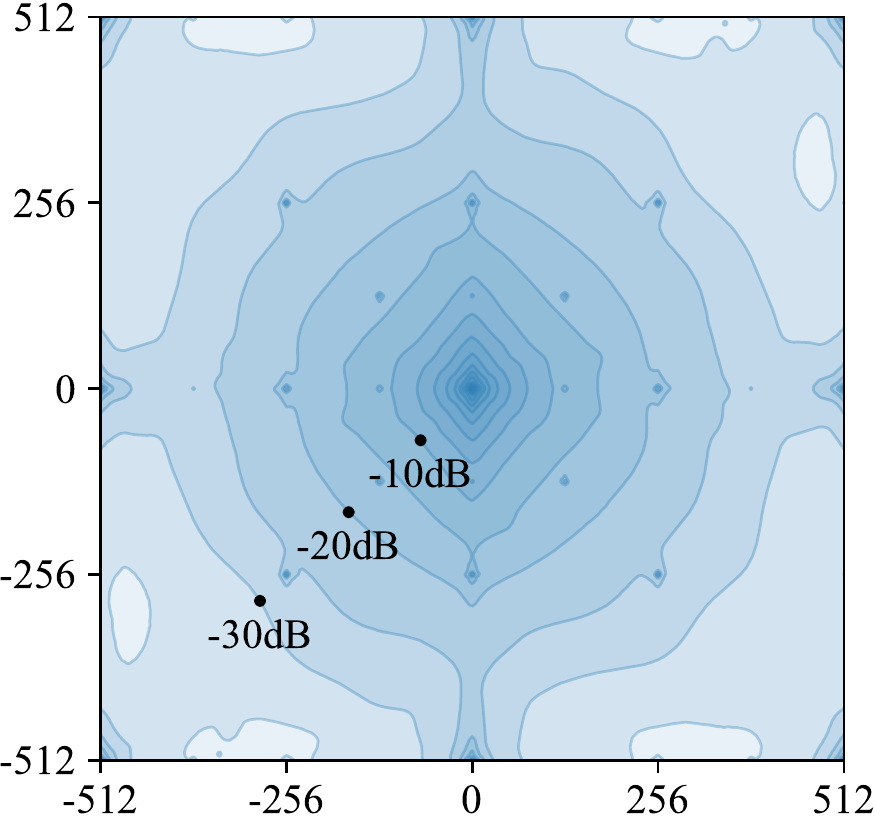}\hfill%
\includegraphics[width=\h]{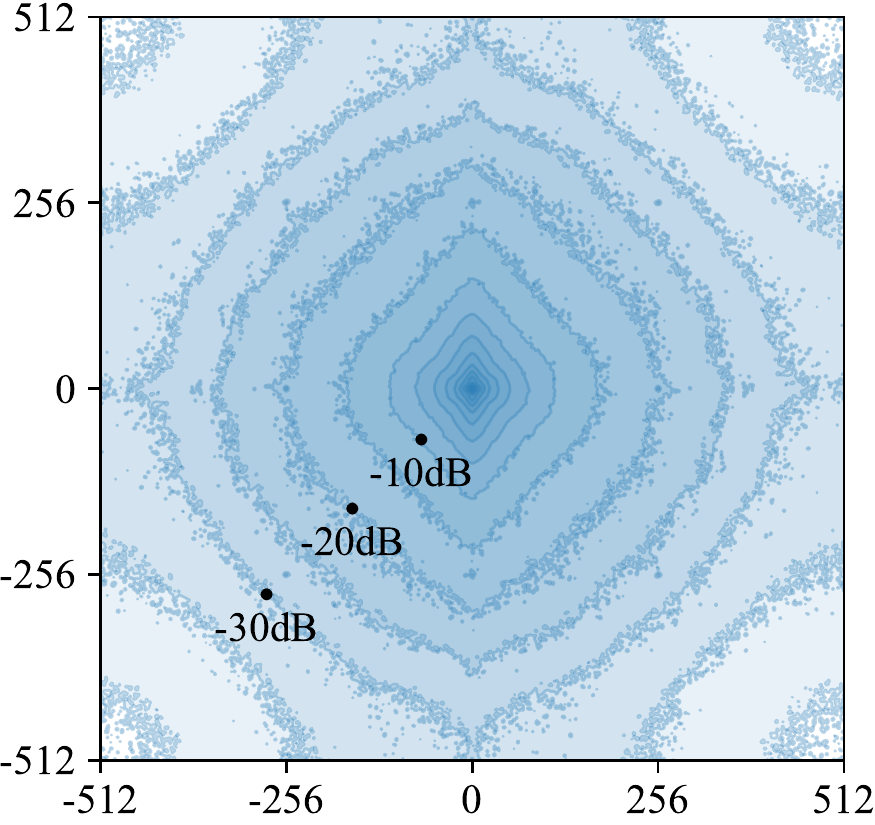}\\[-0.5mm]
\makebox[\h][c]{(a) Training data}\hfill%
\makebox[\h][c]{(b) \FINAL{StyleGAN3-T}}\hfill%
\makebox[\h][c]{(c) SWAGAN~\cite{Gal2021}}\hfill%
\makebox[\h][c]{(d) CIPS~\cite{Anokhin2020}}\\[2mm]
\includegraphics[width=\hh]{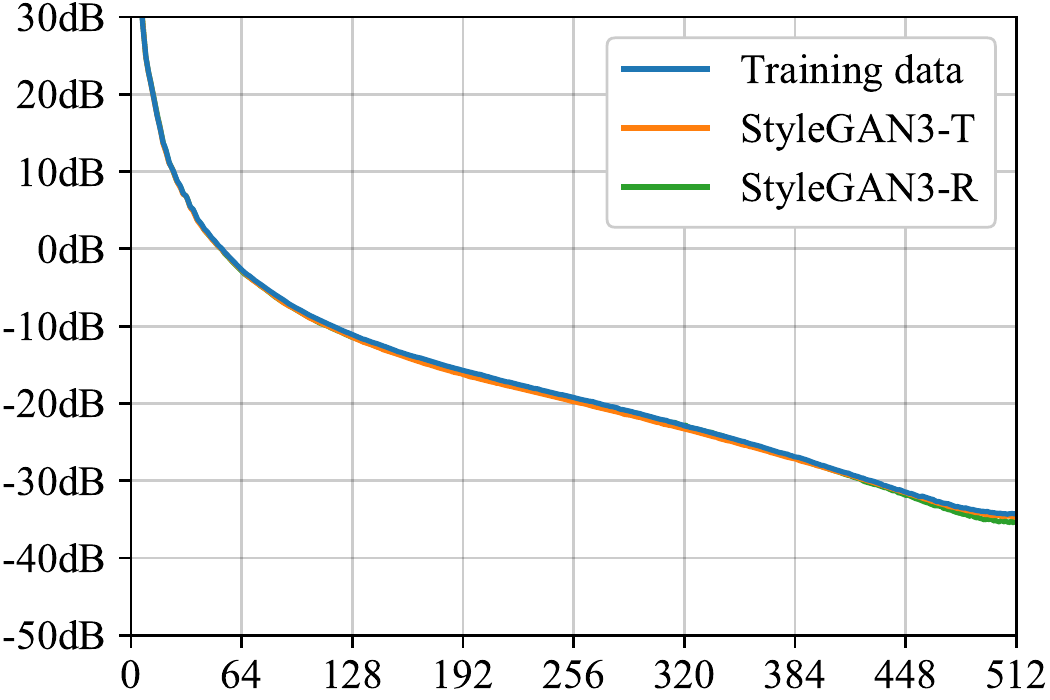}\hfill%
\includegraphics[width=\hh]{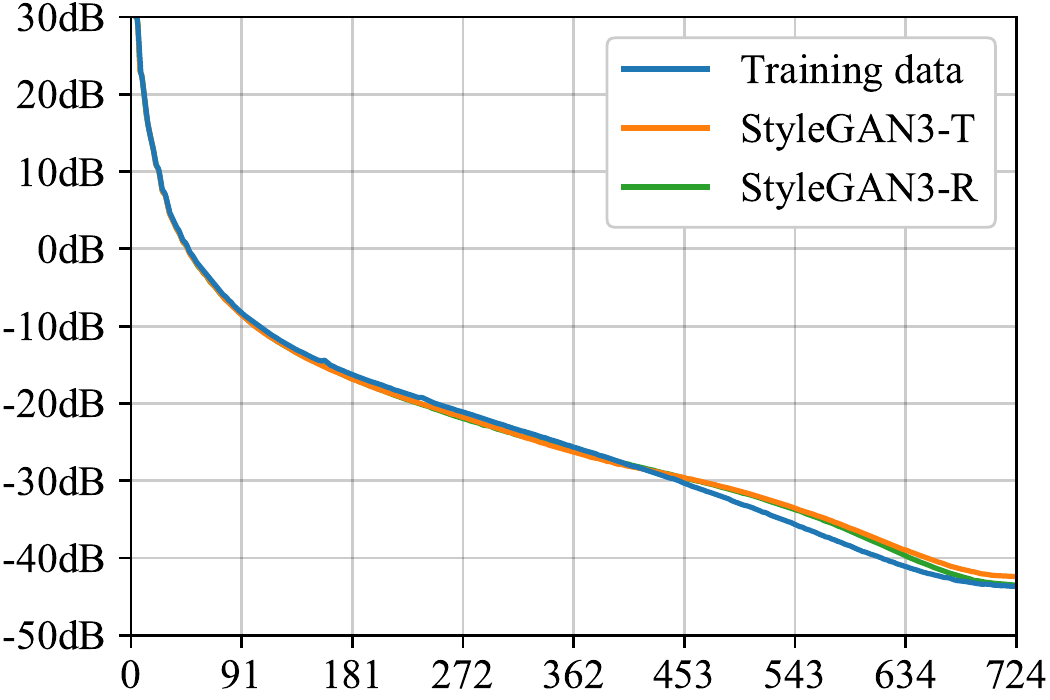}\hfill%
\includegraphics[width=\hh]{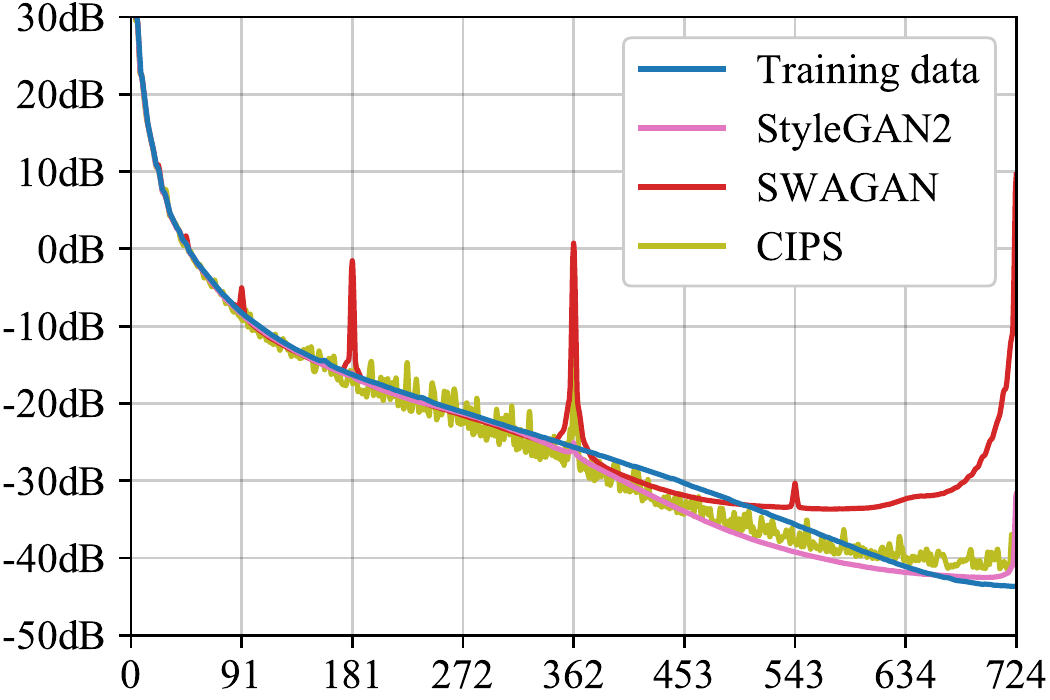}\\
\makebox[\hh][c]{(e) \FINAL{StyleGAN3}, 0$^\circ$ slice}\hfill%
\makebox[\hh][c]{(f) \FINAL{StyleGAN3}, 45$^\circ$ slice}\hfill%
\makebox[\hh][c]{(g) Comparison methods, 45$^\circ$ slice}%
\caption{
\textbf{Top:} Average 2D power spectrum of the training images in \textsc{FFHQ} at 1024$\times$1024 resolution, along with the corresponding spectra of random images generated using \FINAL{StyleGAN3-T}, SWAGAN~\cite{Gal2021}, and CIPS~\cite{Anokhin2020}.
Each plot represents the average power over 70k images, computed as follows. %
From each image, we subtract the training dataset mean, after which we divide it by the training dataset standard deviation. Note that these normalizing quantities represent the entire dataset reduced to two scalars, and do not vary by color channel or pixel coordinate.
The image is then multiplied with a separable Kaiser window with $\beta=8$, and its power spectrum is computed as the absolute values of the FFT raised to the second power.
This processing is applied to each color channel separately and the result is averaged over them.
These spectra are then averaged over all the images in the dataset. The result is plotted on the decibel scale.
\textbf{Bottom:} One-dimensional slices of the power spectra at 0$^\circ$ and 45$^\circ$ angles.
}
\label{#1}
\end{figure}
}

\newcommand{\z}{\leavevmode\hphantom{0}}
\newcommand{\cbs}[1]{{#1}}
\newcommand{\cbhdr}[2]{\makebox[5em][l]{{#1}sample} {#2}$\times$}
\newcommand{\cbheadroom}{\raisebox{0mm}[2.1ex][0mm]{}}
\newcommand{\figCudaBenchmark}[1]{
\begin{figure}[t]
\centering
\scriptsize
\newcolumntype{C}{>{\centering\arraybackslash}p{4.1em}}
\tabulinestyle{0.13mm}%
\begin{tabu}{|l|@{}C@{}C@{}C@{}C@{}|@{}C@{}C@{}C@{}C@{}|@{}C@{}C@{}C@{}C@{}|}
\tabucline{-}\cbheadroom%
& \multicolumn{4}{c|}{\cbhdr{up}{2}  } & \multicolumn{4}{c|}{\cbhdr{up}{4}}   & \multicolumn{4}{c|}{\cbhdr{up}{2}}   \\
& \multicolumn{4}{c|}{\cbhdr{down}{2}} & \multicolumn{4}{c|}{\cbhdr{down}{2}} & \multicolumn{4}{c|}{\cbhdr{down}{4}} \\
Sep. up   & yes & yes & no  & no & yes & yes & no  & no & yes & yes & no  & no \\
Sep. down & yes & no  & yes & no & yes & no  & yes & no & yes & no  & yes & no \\
\tabucline{-}\cbheadroom%
PyTorch \hfill (ms) & 7.88 &  12.40 &  12.68 &  17.12 &  10.07 &  31.51 &  14.96 &  36.33 &  39.35 &  56.73 &   125.83 &   143.15 \\
Ours \hfill (ms)    & 0.42 & \z0.59 & \z0.66 & \z0.92 & \z0.49 & \z0.84 & \z0.80 & \z1.01 & \z1.20 & \z1.89 & \z\z3.04 & \z\z3.66 \\
\tabucline{-}\cbheadroom%
Speedup $\times$& \cbs{19} & \cbs{21} & \cbs{19} & \cbs{19} & \cbs{21} & \cbs{38} & \cbs{19} & \cbs{36} & \cbs{33} & \cbs{30} & \cbs{\z41} & \cbs{\z39} \\
\tabucline{-}
\end{tabu}
\caption{
\footnotesize 
Upsample-nonlinearity-downsample timings in milliseconds using native PyTorch operations vs our optimized CUDA kernel.
The benchmarks were run on NVIDIA Titan V GPU, using input size 512$\times$512$\times$32 and filter size $\ftaps=6$, i.e., $\ftapsup=12$ and $\ftapsup=24$ for \mbox{up-/downsampling} rates of 2 and 4, respectively.
Sep.~up and Sep.~down indicate the use of separable up-/downsampling filters.
}
\label{#1}
\end{figure}
}

\newcommand{\genBridgeUnalignedFfhqSmallFrac}{%
         & {\bf Configuration}                            & FID         & EQ-T         & EQ-T\textsubscript{frac}  \\
\tabucline{-}
{\sc a}  & StyleGAN2                                      & 5.14        & --           & --                        \\
{\sc b}  & + Fourier features                             & 4.79        & 16.23        & 16.28                     \\
{\sc c}  & + No noise inputs                              & 4.54        & 15.81        & 15.84                     \\
{\sc d}  & + Simplified generator                         & 5.21        & 19.47        & 19.57                     \\
\tabucline{-}
{\sc e}  & + Boundaries \& upsampling                     & 6.02        & 24.62        & 24.70                     \\
{\sc f}  & + Filtered nonlinearities                      & 6.35        & 30.60        & 30.68                     \\
{\sc g}  & + Non-critical sampling                        & 4.78        & 43.90        & 42.24                     \\
{\sc h}  & + Transformed Fourier features                 & 4.64        & 45.20        & 42.78                     \\
\tabucline{-}
{\sc t}  & + Flexible layers \hfill(\FINAL{StyleGAN3-T})  & 4.62        & 63.01        & {\bf 46.40}               \\
{\sc r}  & + Rotation equiv. \hfill(\FINAL{StyleGAN3-R})  & {\bf 4.50}  & {\bf 66.65}  & 45.92                     \\
}

\newcommand{\genSweepsCfgRFrac}{%
         & {\bf Parameter}                  & FID         & EQ-T         & EQ-T\textsubscript{frac}  \\
\tabucline{-}
{\sc }   & Filter size $n = 4$              & 4.72        & 57.49        & 44.65                     \\
{\sc *}  & Filter size $n = 6$              & {\bf 4.50}  & {\bf 66.65}  & 45.92                     \\
{\sc }   & Filter size $n = 8$              & 4.66        & 65.57        & {\bf 46.57}               \\
\tabucline{-}
{\sc }   & Upsampling $m = 1$               & {\bf 4.38}  & 39.96        & 37.55                     \\
{\sc *}  & Upsampling $m = 2$               & 4.50        & 66.65        & 45.92                     \\
{\sc }   & Upsampling $m = 4$               & 4.57        & {\bf 74.21}  & {\bf 46.81}               \\
\tabucline{-}
{\sc }   & Stopband $\freqtzero = 2^{1.5}$  & 4.62        & 51.10        & 44.46                     \\
{\sc *}  & Stopband $\freqtzero = 2^{2.1}$  & {\bf 4.50}  & 66.65        & 45.92                     \\
{\sc }   & Stopband $\freqtzero = 2^{3.1}$  & 4.68        & {\bf 73.13}  & {\bf 46.27}               \\
}

\newcommand{\tabBridgeUnalignedFfhqSmallFrac}{%
\newcolumntype{x}{>{\centering\arraybackslash\hspace{0pt}}p{11.8mm}}%
\tabulinesep=0.50mm%
\tabulinestyle{0.17mm}%
\begin{tabu}{|c@{\hspace{2.4mm}}l|xxx|}
\tabucline{-}
\genBridgeUnalignedFfhqSmallFrac
\tabucline{-}
\end{tabu}%
}

\newcommand{\tabSweepsCfgRFrac}{%
\newcolumntype{x}{>{\centering\arraybackslash\hspace{0pt}}p{11.8mm}}%
\tabulinesep=0.605mm%
\tabulinestyle{0.17mm}%
\begin{tabu}{|c@{\hspace{1.4mm}}l|xxx|}
\tabucline{-}
\genSweepsCfgRFrac
\tabucline{-}
\end{tabu}%
}

\newcommand{\figBridgeTablesFrac}[1]{
\begin{figure}[t]
\centering%
\footnotesize%
\scalebox{0.75}{\tabBridgeUnalignedFfhqSmallFrac}\hfill%
\scalebox{0.75}{\tabSweepsCfgRFrac}%
\caption{
Results with our alternative translation equivariance metric \metrict{}\textsubscript{frac}; higher is better.
}
\label{#1}
\end{figure}
}

\newcommand{\tabHyperparams}{
\newcolumntype{x}{>{\centering\arraybackslash\hspace{0pt}}m{10.9mm}}%
\renewcommand{\cs}{\hspace{1.3mm}}%
\tabulinesep=0.50mm%
\tabulinestyle{0.17mm}%
\begin{tabu}{|l|x@{\cs}x@{\cs}x|x@{\cs}x@{\cs}x|}
\tabucline{-}
{\bf Parameter}             & \multicolumn{3}{c|}{\bf Datasets (Figure~\refpaper{fig:ResultTables}, left)} & \multicolumn{3}{c|}{\bf Ablations at 256$\times$256} \\
\hfill Config               & {\sc b}   & {\sc t}   & {\sc r}   &{\sc a--c} &{\sc d--t} & {\sc r}   \\
\tabucline{-}
Batch size                  & 32        & 32        & 32        & 64        & 64        & 64        \\
Moving average              & 10k       & 10k       & 10k       & 20k       & 20k       & 20k       \\
Mapping net depth           & 8         & 2         & 2         & 8         & 2         & 2         \\
Minibatch stddev            & 4         & 4         & 4         & 8         & 4         & 4         \\
\tabucline{-}
G layers                    & 15/17     & 14        & 14        & 13        & 14        & 14        \\
G capacity: $C_\text{base}$ & $2^{15}$  & $2^{15}$  & $2^{16}$  & $2^{14}$  & $2^{14}$  & $2^{15}$  \\
G capacity: $C_\text{max}$  & 512       & 512       & 1024      & 512       & 512       & 1024      \\
G learning rate             & 0.0020    & \FINAL{0.0025} & \FINAL{0.0025} & 0.0025 & \FINAL{0.0025} & \FINAL{0.0025} \\
D learning rate             & 0.0020    & 0.0020    & 0.0020    & 0.0025    & 0.0025    & 0.0025    \\
\tabucline{-}
\end{tabu}%
}

\newcommand{\tabGamma}{
\newcolumntype{x}{>{\centering\arraybackslash\hspace{0pt}}m{10.9mm}}%
\renewcommand{\cs}{\hspace{1.3mm}}%
\tabulinesep=0.95mm%
\tabulinestyle{0.17mm}%
\begin{tabu}{|l@{\cs}c|x@{\cs}x@{\cs}x|}
\tabucline{-}
& & & & \\[-2.5mm]
\multicolumn{2}{|l|}{\bf R\textsubscript{1} regularization $\gamma$} & {\sc b} & {\sc t} & {\sc r} \\
& & & & \\[-2.5mm]
\tabucline{-}
\textsc{FFHQ-U}     & \s256$^2$ & \s1.0 & \s1.0 & \s1.0 \\
\textsc{FFHQ-U}     & 1024$^2$  & 10.0  & 32.8  & 32.8  \\
\textsc{FFHQ}       & 1024$^2$  & 10.0  & 32.8  & 32.8  \\
\textsc{MetFaces-U} & 1024$^2$  & 10.0  & 16.4  & \s6.6 \\
\textsc{MetFaces}   & 1024$^2$  & \s5.0 & \s6.6 & \s3.3 \\
\textsc{AFHQv2}     & \s512$^2$ & \s5.0 & \s8.2 & 16.4  \\
\textsc{Beaches}    & \s512$^2$ & \s2.0 & \s4.1 & 12.3  \\
\tabucline{-}
\end{tabu}%
}

\newcommand{\figHyperparams}[1]{
\begin{figure}[t]
\footnotesize%
\centering%
\scalebox{0.75}{\tabHyperparams}\hfill%
\scalebox{0.75}{\tabGamma}%
\caption{
\textbf{Left:} Hyperparameters used in each experiment.
\textbf{Right:} $R_1$ regularization weights.
}
\label{#1}
\end{figure}
}

\newcommand{\SUBT}[1]{{\color{grey}\hspace{4mm}#1}}
\newcommand{\SUBV}[1]{{\color{grey}#1}}

\newcommand{\figEnergy}[1]{
\begin{figure}[t]
\centering%
\footnotesize%
\newcolumntype{x}{>{\arraybackslash\hspace{0pt}}p{52.7mm}}%
\newcolumntype{y}{>{\centering\arraybackslash\hspace{0pt}}p{23mm}}%
\tabulinesep=0.50mm%
\tabulinestyle{0.17mm}%
\begin{tabu}{|@{\hspace{2.5mm}}x|y|y|y|}
\hline
\multirow{2}{*}{\bf Item}                       & {\bf Number of}       & {\bf GPU years}   & {\bf Electricity} \\
                                                & {\bf training runs}   & {\bf (Volta)}     & {\bf (MWh)}       \\
\tabucline{-}
Early exploration                               & \s233                 & 18.02             & \s42.45           \\
Project exploration                             & 1207                  & 48.93             & 118.13            \\
Setting up ablations                            & \s297                 & 13.30             & \s32.48           \\
Per-dataset tuning                              & \s\s63                & \s4.54            & \s13.28           \\
Producing results in the paper                  & \s\s53                & \s5.26            & \s14.35           \\
\SUBT{\FINAL{StyleGAN3-R} at 1024$\times$1024}  & \SUBV{\s\s\s1}        & \SUBV{\s0.30}     & \SUBV{\s\s0.87}   \\
\SUBT{Other runs in the dataset table}          & \SUBV{\s\s17}         & \SUBV{\s2.35}     & \SUBV{\s\s6.88}   \\
\SUBT{Ablation tables}                          & \SUBV{\s\s35}         & \SUBV{\s2.61}     & \SUBV{\s\s6.60}   \\
Results intentionally left out                  & \s\s23                & \s1.72            & \s\s3.93          \\
\tabucline{-}
Total                                           & 1876                  & 91.77             & 224.62            \\
\tabucline{-}
\end{tabu}
\caption{
Computational effort expenditure and electricity consumption data for this project. The unit for computation is GPU-years on a single NVIDIA V100 GPU\,---\,it would have taken approximately 92 years to execute this project using a single GPU. See the text for additional details about the computation and energy consumption estimates.
\textbf{Early exploration} includes early training runs that affected our decision to start this project.
\textbf{Project exploration} includes training runs that were done specifically for this project, leading to the final \FINAL{StyleGAN3-T} and \FINAL{StyleGAN3-R} configurations. These runs were not intended to be used in the paper as-is.
\textbf{Setting up ablations} includes hyperparameter tuning for the intermediate configurations and ablation experiments in Figures~\refpaper{fig:BridgeTables} and~\refpaper{fig:ResultTables}.
\textbf{Per-dataset tuning} includes hyperparameter tuning for individual datasets, mainly the grid search for $R_1$ regularization weight.
\textbf{Config \textsc{r} at 1024$\times$1024} corresponds to one training run in Figure~\refpaper{fig:ResultTables}, left, and \textbf{Other runs in the dataset table} includes the remaining runs.
\textbf{Ablation tables} includes the low-resolution ablations in Figures~\refpaper{fig:BridgeTables} and Figure~\refpaper{fig:ResultTables}.
\textbf{Results intentionally left out} includes additional results that were initially planned, but then left out to improve focus and clarity.
}
\label{#1}
\end{figure}
}

  \setcounter{figure}{6} %
\setcounter{equation}{3} %

\section{Additional results}
\label{app:results}

\figUncuratedFFHQU{fig:UncuratedFFHQU} %
\figUncuratedMetFacesU{fig:UncuratedMetFacesU} %
\figUncuratedAFHQtwo{fig:UncuratedAFHQtwo} %
\figUncuratedBeaches{fig:UncuratedBeaches} %

Uncurated sets of samples for \FINAL{StyleGAN2 (baseline config~\textsc{b} with Fourier features) and our alias-free generators StyleGAN3-T and StyleGAN3-R} are shown in Figures~\ref{fig:UncuratedFFHQU} (\mbox{\textsc{FFHQ-U}}), \ref{fig:UncuratedMetFacesU}~(\textsc{MetFaces-U}), \ref{fig:UncuratedAFHQtwo}~ (\textsc{AFHQv2}), and \ref{fig:UncuratedBeaches}~(\textsc{Beaches}).
Truncation trick was not used when generating the images.

StyleGAN2 and our generators yield comparable FIDs in all of these datasets. 
Visual inspection did not reveal anything surprising in the first three datasets, but in \textsc{Beaches} our new generators seem to generate a somewhat reduced set of possible scene layouts properly. 
We suspect that this is related to the lack of noise inputs, which forces the generators to waste capacity for what is essentially random number generation \cite{Karras2019}. Finding a way to reintroduce noise inputs without breaking equivariances is therefore an important avenue of future work.

The accompanying interpolation videos reveal major differences between StyleGAN2 and \FINAL{StyleGAN3-R}. 
For example, in \textsc{MetFaces} much of details such as brushstrokes or cracked paint seems to be glued to the pixel coordinates in StyleGAN2, whereas with \FINAL{StyleGAN3} all details move together with the depicted model. 
The same is evident in \textsc{AFHQv2} with the fur moving credibly in \FINAL{StyleGAN3} interpolations, while mostly sticking to the image coordinates in StyleGAN2. 
In \textsc{Beaches} we furthermore observe that StyleGAN2 tends to ``fade in'' details while retaining a mostly fixed viewing position, while \FINAL{StyleGAN3} creates plenty of apparent rotations and movement.
The videos use hand-picked seeds to better showcase the relevant effects.

In a further test we created two example cinemagraphs that mimic small-scale head movement and facial animation in FFHQ.
The geometric head motion was generated as a random latent space walk along hand-picked directions from GANSpace~\cite{Harkonen2020} and SeFa~\cite{Shen2021}.
The changes in expression were realized by applying the ``global directions'' method of StyleCLIP~\cite{Patashnik2021}, using the prompts ``angry face'', ``laughing face'', ``kissing face'', ``sad face'', ``singing face'', and ``surprised face''.
The differences between StyleGAN2 and \FINAL{StyleGAN3} are again very prominent, with the former displaying jarring sticking of facial hair and skin texture, even under subtle movements.

The equivariance quality videos illustrate the practical relevance of the PSNR numbers in Figures~3 and~5 of the main paper.
We observe that for EQ-T numbers over $\sim$50\,dB indicate high-quality results, and for EQ-R $\sim$40\,dB look good.

We also provide an animated version of the nonlinearity visualization in Figure~2.

\figStyleMixing{fig:StyleMixing} %

In style mixing~\cite{Karras2019} two or more independently chosen latent codes are fed into different layers of the generator. 
Ideally all combinations would produce images that are not obviously broken, and furthermore, it would be desirable that specific layers end up controlling well-defined semantic aspects in the images. 
StyleGAN uses mixing regularization~\cite{Karras2019} during training to achieve these goals.
We observe that mixing regularization continues to work similarly in \FINAL{StyleGAN3}, but we also wanted to know whether it is truly necessary because the regularization is known to be detrimental for many complex and multi-modal datasets~\cite{Gwern2020}.
When we disable the regularization, obviously broken images remain rare, based on a visual inspection of a large number of images.
The semantically meaningful controls are somewhat compromised, however, as Figure~\ref{fig:StyleMixing} shows.

\figConvergence{fig:Convergence} %
Figure~\ref{fig:Convergence} compares the convergence of our main configurations (config~\textsc{t} and~\textsc{r}) against the results of Karras~et~al.~\cite{Karras2019,Karras2020}.
The overall shape of the curves is similar; introducing translation and rotation equivariance in the generator does not appear to significantly alter the training dynamics.

\figAverageSpectra{fig:AverageSpectra} %
Following recent works that address signal processing issues in GANs~\cite{Anokhin2020,Gal2021}, we show average power spectra of the generated 
and real images in Figure~\ref{fig:AverageSpectra}. The plots are computed from images that are whitened with the overall training dataset 
mean and standard deviation. Because FFT interprets the signal as periodic, we eliminate the sharp step edge across the image borders by 
windowing the pixel values prior to the transform. This eliminates the axis-aligned cross artifact which may obscure meaningful detail in 
the spectrum. We display the average 2D spectrum as a contour plot, which makes the orientation-dependent falloff apparent, and highlights
detail like regularly spaced residuals of upsampling grids, and fixed noise patterns. We also plot 1D slices of the spectrum along the 
horizontal and diagonal angle without azimuthal integration, so as to not average out the detail. 
\FINAL{The code for reproducing these steps is included in the public release}.

\section{Datasets}
\label{app:datasets}

In this section, we describe the new datasets and list the licenses of all datasets. 

\subsection{FFHQ-U and MetFaces-U}
We built unaligned variants of the existing \mbox{\textsc{FFHQ}}~\cite{Karras2018} and \mbox{\textsc{MetFaces}}~\cite{Karras2020} datasets. 
The originals are available at {\small\url{https://github.com/NVlabs/ffhq-dataset}} and
{\small\url{https://github.com/NVlabs/metfaces-dataset}}, respectively. 
The datasets were rebuilt with a modification of the original procedure based on the original code, raw uncropped images, and facial landmark metadata.
\FINAL{The code required to reproduce the modified datasets is included in the public release}.

We use axis-aligned crop rectangles, and do not rotate them to match the orientation of the face. This retains the natural variation of camera and head tilt angles. Note that the images are still generally upright, i.e., never upside down or at $90^\circ$ angle. The scale of the rectangle is determined as before.
For each image, the crop rectangle is randomly shifted from its original face-centered position, with the horizontal and vertical offset independently drawn from a normal distribution. The standard deviation is chosen as $20 \%$ of the crop rectangle dimension. If the crop rectangle falls partially outside the original image boundaries, we keep drawing new random offsets until we find one that does not. This removes the need to pad the images with fictional mirrored content, and we explicitly disabled this feature of the original build script.

Aside from the exact image content, the number of images and other specifications match the original dataset exactly.
While FFHQ-U contains identifiable images of persons, it does not introduce new images beyond those already in the original FFHQ. 

\subsection{AFHQv2}
We used an updated version of the \mbox{\textsc{AFHQ}} dataset~\cite{AFHQ} where the resampling filtering has been improved.
The original dataset suffers from pixel-level artifacts caused by inadequate downsampling filters~\cite{Parmar2021}.
This caused convergence problems with our models, as the sharp ``stair-step'' aliasing artifacts are difficult to reproduce without direct access to the pixel grid.

The dataset was rebuilt using the original uncropped images and crop rectangle metadata, 
using the PIL library implementation of Lanczos resampling as recommended by Parmar et al.~\cite{Parmar2021}. 
In a minority of cases, the crop rectangles were modified to remove non-isotropic scaling and other unnecessary transformations. 
A small amount ($\sim 2\%$) of images were dropped for technical reasons, leaving a total of $15803$ images. 
Aside from this, the specifications of the dataset match the original. We use all images of all the three classes (cats, dogs, and wild animals) as one training dataset.

\subsection{Beaches}

\textsc{Beaches} is a new dataset of $20155$ photographs of beaches at resolution 512$\times$512.
The training images were provided by Getty Images. 
\mbox{\textsc{Beaches}} is a proprietary dataset that we are licensed to use, but not to redistribute.
We are therefore unable to release the full training data or pre-trained models for this dataset.

\subsection{Licenses}
The \textsc{FFHQ} dataset is available under Creative Commons BY-NC-SA 4.0 license by NVIDIA Corporation, and consist of images published by respective authors under Creative Commons BY 2.0, Creative Commons BY-NC 2.0, Public Domain Mark 1.0, Public Domain CC0 1.0, and U.S. Government Works license. 

The \textsc{MetFaces} dataset is available under Creative Commons BY-NC 2.0 license by NVIDIA Corporation, and consists of images available under the Creative Commons Zero (CC0) license by the Metropolitan Museum of Art.

The original AFHQ dataset is available at {\small\url{https://github.com/clovaai/stargan-v2}} under Creative Commons BY-NC 4.0 license by NAVER Corporation. 

\newcommand{\fkaiser}{h_K}
\newcommand{\fkaiserup}{h'_K}
\newcommand{\fkaisersep}{h_K^+}
\newcommand{\fkaiserrad}{h_K^\circ}
\newcommand{\flanczos}{h_L}
\newcommand{\flanczossep}{h_L^+}
\newcommand{\fgaussian}{h_G}
\newcommand{\fgaussiansep}{h_G^+}

\newcommand{\wkaiser}{w_K}
\newcommand{\kext}{L}
\newcommand{\kextup}{L'}
\newcommand{\katten}{A}
\newcommand{\ftaps}{n}
\newcommand{\ftapsup}{n'}
\newcommand{\fwidth}{\Delta\freq}
\newcommand{\fwidthup}{\Delta\freq'}
\newcommand{\sfrequp}{\sfreq'}

\newcommand{\ff}{\boldsymbol{f}}
\newcommand{\klanczos}{k_L}
\newcommand{\kgaussian}{k_G}

\section{Filter details}
\label{app:filter}

In this section, we review basic FIR filter design methodology and detail the recipe used to construct the upsampling and downsampling filters in our generator.
We start with simple Kaiser filters in one dimension, discussing parameter selection and the necessary modifications needed for upsampling and downsampling.
We then proceed to extend the filters to two dimensions and conclude by detailing the alternative filters evaluated in Figure~\refpaper{fig:ResultTables}, right.
Our definitions are consistent with standard signal processing literature (e.g., Oppenheim~\cite{Oppenheim2009}) as well as widely used software packages (e.g., \texttt{scipy.signal.firwin}).

\subsection{Kaiser low-pass filters}

In one dimension, the ideal continuous-time low-pass filter with cutoff $\freqc$ is given by $\ideallow(x) = 2 \freqc \cdot \sinc(2 \freqc x)$, where $\sinc(x) = \sin(\pi x)/(\pi x)$.
The ideal filter has infinite attenuation in the stopband, i.e., it completely eliminates all frequencies above $\freqc$.
However, its impulse response is also infinite, which makes it impractical for three reasons: implementation efficiency, border artifacts, and \emph{ringing} caused by long-distance interactions.
The most common way to overcome these issues is to limit the spatial extent of the filter using the window method~\cite{Oppenheim2009}:
\begin{equation}
  \fkaiser(x) = 2 \freqc \cdot \sinc(2 \freqc x) \cdot \wkaiser(x),
\end{equation}
where $\wkaiser(x)$ is a \emph{window function} and $\fkaiser(x)$ is the resulting practical approximation of $\ideallow(x)$.
Different window functions represent different tradeoffs between the frequency response and spatial extent; the smaller the spatial extent, the weaker the attenuation.
In this paper we use the Kaiser window~\cite{Kaiser1974}, also known as the Kaiser--Bessel window, that provides explicit control over this tradeoff.
The Kaiser window is defined as
\begin{equation}
  \wkaiser(x) =
    \begin{cases}
      I_0 \Big( \beta \sqrt{1 - (2 x / \kext)^2} \Big) \big/ I_0 \big( \beta \big), & \textrm{if } |x| \le \kext / 2, \\
      0,                                                                            & \textrm{if } |x| > \kext / 2,
    \end{cases}
\end{equation}
where $\kext$ is the desired spatial extent, $\beta$ is a free parameter that controls the shape of the window, and $I_0$ is the zeroth-order modified Bessel function of the first kind.
Note that the window has discontinuities at $\pm \kext / 2$; the value is strictly positive at $x = \kext / 2$ but zero at $x = \kext / 2 + \epsilon$.

When operating on discretely sampled signals, it is necessary to discretize the filter as well:
\begin{equation}
  \label{eq:app_discretization}
  \fkaiser[i] = \fkaiser \Big( \big( i - (\ftaps - 1) / 2 \big) / \sfreq \Big) \big/ s, \hspace{2mm} \textrm{for } i \in \{0, 1, \ldots, \ftaps - 1\},
\end{equation}
where $\fkaiser[i]$ is the discretized version of $\fkaiser(x)$ and $\sfreq$ is the sampling rate.
The filter is defined at $\ftaps$ discrete spatial locations, i.e., \emph{taps}, located $1/s$ units apart and placed symmetrically around zero.
Given the values of $\ftaps$ and $\sfreq$, the spatial extent can be expressed as $\kext = (\ftaps - 1) / \sfreq$.
An odd value of $\ftaps$ results in a \emph{zero-phase} filter that preserves the original sample locations, whereas an even value shifts the sample locations by $1/(2s)$ units.

The filters considered in this paper are approximately normalized by construction, i.e., $\int_x \fkaiser(x) \approx \sum_i \fkaiser[i] \approx 1$.
Nevertheless, we have found it beneficial to explicitly normalize them after discretization.
In other words, we strictly enforce $\sum_i \fkaiser[i] = 1$ by scaling the filter taps to reduce the risk of introducing cumulative scaling errors when the signal is passed through several consecutive layers.

\subsection{Selecting window parameters}

Kaiser~\cite{Kaiser1974} provides convenient empirical formulas to connect the parameters of $\wkaiser$ to the properties of $\fkaiser$.
Given the number of taps and the desired transition band width, the maximum attenuation achievable with $\fkaiser[i]$ is approximated by
\begin{equation}
  \katten = 2.285 \cdot (\ftaps - 1) \cdot \pi \cdot \fwidth + 7.95,
\end{equation}
where $\katten$ is the attenuation measured in decibels and $\fwidth$ is the width of the transition band expressed as a fraction of $\sfreq / 2$.
We choose to define the transition band using half-width $\freqh$, which gives $\fwidth = (2 \freqh)/(\sfreq / 2)$.
Given the value of $A$, the optimal choice for the shape parameter $\beta$ is then approximated~\cite{Kaiser1974} by
\begin{equation}
  \beta =
    \begin{cases}
      0.1102 \cdot (\katten - 8.7),                                     & \textrm{if } \katten > 50, \\
      0.5842 \cdot (\katten - 21)^{0.4} + 0.07886 \cdot (\katten - 21), & \textrm{if } 21 \le \katten \le 50, \\
      0,                                                                & \textrm{if } \katten < 21,
    \end{cases}
\end{equation}
This leaves us with two free parameters: $\ftaps$ controls the spatial extent while $\freqh$ controls the transition band.
The choice of these parameters directly influences the resulting attenuation; increasing either parameter yields a higher value for $A$.

\subsection{Upsampling and downsampling}

When upsampling a signal, i.e., $\Layern{up}(\Feat) = \combsp\mult(\idealsqs \conv \Feat) = 1/\sfreq^2 \cdot \combsp\mult(\ideallow_\sfreq \conv \Feat)$, we are concerned not only the with input sampling rate $\sfreq$, but also with the output sampling rate $\sfrequp$.
With an integer upsampling factor $m$, we can think of the upsampling operation as consisting of two steps: we first increase the sampling rate to $\sfrequp = \sfreq \cdot m$ by interleaving $m - 1$ zeros between each input sample by and then low-pass filter the resulting signal to eliminate the alias frequencies above $\sfreq / 2$.
In order to keep the signal magnitude unchanged, we must also scale the result by $m$ with one-dimensional signals, or by $m^2$ with two-dimensional signals.
Since the filter now operates under $\sfrequp$ instead of $\sfreq$, we must adjust its parameters accordingly:
\begin{align}
  \label{eqn:nprime}
  \ftapsup &= \ftaps \cdot m, &
  \kextup &= (\ftapsup - 1) / \sfrequp, &
  \fwidthup &= (2 \freqh) / (\sfrequp / 2),
\end{align}
which gives us the final upsampling filter
\begin{equation}
  \fkaiserup[i] = \fkaiserup \Big( \big( i - (\ftapsup - 1) / 2 \big) / \sfrequp \Big) \big/ \sfrequp, \hspace{2mm} \textrm{for } i \in \{0, 1, \ldots, \ftapsup - 1\}.
\end{equation}
Multiplying the number of taps by $m$ keeps the spatial extent of the filter unchanged with respect to the input samples, and it also compensates for the reduced attenuation from $\fwidthup < \fwidth$.
Note that if the upsampling factor is even, $\ftapsup$ will be even as well, meaning that $\fkaiserup$ shifts the sample locations by $1/(2\sfrequp)$.
This is the desired behavior\,---\,if we consider sample $i$ to represent the continuous interval $[i \cdot \sfreq, (i+1) \cdot \sfreq]$ in the input signal, the same interval will be represented by $m$ consecutive samples $m \cdot i, \ldots, m \cdot i + m - 1$ in the output signal.
Using a zero-phase upsampling filter, i.e., an odd value for $\ftapsup$, would break this symmetry, leading to inconsistent behavior with respect to the boundaries.
Note that our symmetric interpretation is common in many computer graphics APIs, such as OpenGL, and it is also reflected in our definition of the Dirac comb $\comb$ in Section~\refpaper{sec:theory}.

Upsampling and downsampling are adjoint operations with respect to each other, disregarding the scaling of the signal magnitude.
This means that the above definitions are readily applicable to downsampling as well; to downsample a signal by factor $m$, we first filter it by $\fkaiserup$ and then discard the last $m - 1$ samples within each group of $m$ consecutive samples.
The interpretation of all filter parameters, as well as the sample locations, is analogous to the upsampling case.

\subsection{Two-dimensional filters}

Any one-dimensional filter, including $\fkaiser$, can be trivially extended to two dimensions by defining the corresponding separable filter
\begin{equation}
  \fkaisersep(\xx) = \fkaiser(x_0) \cdot \fkaiser(x_1) = (2 \freqc)^2 \cdot \sinc(2 \freqc x_0) \cdot \sinc(2 \freqc x_1) \cdot \wkaiser(x_0) \cdot \wkaiser(x_1),
\end{equation}
where $\xx = (x_0, x_1)$.
$\fkaisersep$ has the same cutoff as $\fkaiser$ along the coordinate axes, i.e., $\ff_{c,x} = (\freqc, 0)$ and $\ff_{c,y} = (0, \freqc)$, and its frequency response forms a square shape over the 2D plane, implying that the cutoff frequency along the diagonal is $\ff_{c,d} = (\freqc, \freqc)$.
In practice, a separable filter can be implemented efficiently by first filtering each row of the two-dimensional signal independently with $\fkaiser$ and then doing the same for each column.
This makes $\fkaisersep$ an ideal choice for all upsampling filters in our generator, as well as the downsampling filters in configs~\textsc{a--t} (Figure~\refpaper{fig:BridgeTables}, left).

The fact that the spectrum of $\fkaisersep$ is not radially symmetric, i.e., $\Vert \ff_{c,d} \Vert \ne \Vert \ff_{c,x} \Vert$, is problematic considering config~\textsc{r}.
If we rotate the input feature maps of a given layer, their frequency content will rotate as well.
To enforce rotation equivariant behavior, we must ensure that the effective cutoff frequencies remain unchanged by this.
The ideal radially symmetric low-pass filter~\cite{Blahut2004} is given by $\ideallow_\sfreq^\circ(\xx) = (2 \freqc)^2 \cdot \jinc(2 \freqc \lVert\xx\rVert)$.
The $\jinc$ function, also known as besinc, sombrero function, or Airy disk, is defined as $\jinc(x) = 2 J_1(\pi x) / (\pi x)$, where $J_1$ is the first order Bessel function of the first kind.
Using the same windowing scheme as before, we define the corresponding practical filter as
\begin{align}
  \fkaiserrad(\xx) = (2 \freqc)^2 \cdot \jinc(2 \freqc \Vert\xx\Vert) \cdot \wkaiser(x_0) \cdot \wkaiser(x_1).
\end{align}
Note that even though $\jinc$ is radially symmetric, we still treat the window function as separable in order to retain its spectral properties.
In config~\textsc{r}, we perform all downsampling operations using $\fkaiserrad$, except for the last two critically sampled layers where we revert to $\fkaisersep$.

\subsection{Alternative filters}

In Figure~\refpaper{fig:ResultTables}, right, we compare the effectiveness of Kaiser filters against two alternatives: Lanczos and Gaussian.
These filters are typically defined using prototypical filter kernels $\klanczos$ and $\kgaussian$, respectively:
\begin{align}
  \klanczos(x) &=
  \begin{cases}
    \sinc(x) \cdot \sinc(x / a), & \textrm{if } |x| < a, \\
    0,                           & \textrm{if } |x| \ge a,
  \end{cases} \\
  \kgaussian(x) &= \exp \left( -\frac{1}{2} (x / \sigma)^2 \right) \Big/ \left( \sigma \sqrt{2 \pi} \right),
\end{align}
where $a$ is the spatial extent of the Lanczos kernel, typically set to 2 or 3, and $\sigma$ is the standard deviation of the Gaussian kernel.
In Figure~\refpaper{fig:ResultTables} of the main paper we set $a = 2$ and $\sigma = 0.4$; we tested several different values and found these choices to work reasonably well.

The main shortcoming of the prototypical kernels is that they do not provide an explicit way to control the cutoff frequency.
In order to enable apples-to-apples comparison, we assume that the kernels have an implicit cutoff frequency at 0.5 and scale their impulse responses to account for the varying $\freqc$:
\begin{align}
  \label{eq:app_lanczos}
  \flanczos(x) &= 2 \freqc \cdot \klanczos(2 \freqc x), &
  \fgaussian(x) &= 2 \freqc \cdot \kgaussian(2 \freqc x).
\end{align}
We limit the computational complexity of the Gaussian filter by enforcing $\fgaussian(x) = 0$ when $|x| > 8/s$, with respect to the input sampling rate in the upsampling case.
In practice, $\fgaussian(x)$ is already very close to zero in this range, so the effect of this approximation is negligible.
Finally, we extend the filters to two dimensions by defining the corresponding separable filters:
\begin{align}
  \flanczossep(\xx) &= (2 \freqc)^2 \cdot \klanczos(2 \freqc x_0) \cdot \klanczos(2 \freqc x_1), &
  \fgaussiansep(\xx) &= (2 \freqc)^2 \cdot \kgaussian(2 \freqc x_0) \cdot \kgaussian(2 \freqc x_1).
\end{align}
Note that $\fgaussiansep$ is radially symmetric by construction, which makes it ideal for rotation equivariance.
$\flanczossep$, however, has no widely accepted radially symmetric counterpart, so we simply use the same separable filter in config~\textsc{r} as well.

\section{Custom CUDA kernel for filtered nonlinearity}
\label{app:kernel}

Implementing the upsample-nonlinearity-downsample sequence is inefficient using the standard primitives available in modern deep learning frameworks.
The intermediate feature maps have to be transferred between on-chip and off-chip GPU memory multiple times and retained for the backward pass.
This is especially costly because the intermediate steps operate on upsampled, high-resolution data.
To overcome this, we implement the entire sequence as a single operation using a custom CUDA kernel.
This improves training performance by approximately an order of magnitude thanks to reduced memory traffic, and also decreases GPU memory usage significantly.

The combined kernel consists of four phases: input, upsampling, nonlinearity, and downsampling.
The computation is parallelized by subdividing the output feature maps into non-overlapping tiles, and computing one output tile per CUDA thread block.
First, in input phase, the corresponding input region is read into on-chip shared memory of the thread block.
Note that the input regions for neighboring output tiles will overlap spatially due to the spatial extent of filters.

The execution of up-/downsampling phases depends on whether the corresponding filters are separable or not.
For a separable filter, we perform vertical and horizontal 1D convolutions sequentially, whereas a non-separable filter requires a single 2D convolution.
All these convolutions and the nonlinearity operate in on-chip shared memory, and only the final output of the downsampling phase is written to off-chip GPU memory.

\subsection{Gradient computation}

To compute gradients of the combined operation, they need to propagate through each of the phases in reverse order.
Fortunately, the combined upsample-nonlinearity-downsample operation is mostly self-adjoint with proper changes in parameters, e.g., swapping the up-/downsampling factors and the associated filters.
The only problematic part is the nonlinearity that is performed in the upsampled resolution.
A na\"ive but general solution would be to store the intermediate high-resolution input to the nonlinearity, but the memory consumption would be infeasible for training large models.

Our kernel is specialized to use leaky ReLU as the nonlinearity, which offers a straightforward way to conserve memory: to propagate gradients, it is sufficient to know whether the corresponding input value to nonlinearity was positive or negative.
When using 16-bit floating-point datatypes, there is an additional complication because the outputs of the nonlinearity need to be clamped~\cite{Karras2020}, and when this occurs, the corresponding gradients must be zero.
Therefore, in the forward pass we store two bits of auxiliary information per value to cover the three possible cases: positive, negative, or clamped.
In the backward pass, reading these bits is sufficient for correct gradient computation\,---\,no other information from the forward pass is needed.

\subsection{Optimizations for common upsampling factors}

Let us consider one-dimensional 2$\times$ upsampling where the input is (virtually) interleaved with zeros and convolved with an $\ftapsup$-tap filter where $\ftapsup=2\ftaps$ (cf.~Equation~\ref{eqn:nprime}).
There are $\ftaps$ nonzero input values under the $\ftapsup$-tap kernel, so if each output pixel is computed separately, the convolution requires $\ftaps$ multiply-add operations per pixel and equally many shared memory load instructions, for a total of $2\ftaps$ instructions per output pixel.%
\footnote{Input of the upsampling is stored in shared memory, but the filter weights can be stored in CUDA constant memory where they can be accessed without a separate load instruction.}
However, note that the computation of two neighboring output pixels accesses only $\ftaps+1$ input pixels in total.
By computing two output pixels at a time and avoiding redundant shared memory load instructions, we obtain an average cost of \low{$\frac{3}{2}\ftaps+\frac{1}{2}$} instructions per pixel\,---\,close to 25\% savings.
For 4$\times$ upsampling, we can similarly reduce the instruction count by up to 37.5\% by computing four output pixels at a time.
We apply these optimizations in 2$\times$ and 4$\times$ upsampling for both separable and non-separable filters.

\figCudaBenchmark{fig:cudabenchmark} %
Figure~\ref{fig:cudabenchmark} benchmarks the performance of our kernel with various up-/downsampling factors and with separable and non-separable filters.
In network layers that keep the sampling rate fixed, both factors are 2$\times$, whereas layers that increase the sampling rate by a factor of two, 4$\times$ upsampling is combined with 2$\times$ downsampling.
The remaining combination of 2$\times$ upsampling and 4$\times$ downsampling is needed when computing gradients of the latter case.
The speedup over native PyTorch operations varies between $\sim$20--40$\times$, which yields an overall training speedup of approximately 10$\times$.

\newcommand{\pp}{\boldsymbol{p}}
\newcommand{\wlanczossep}{w^+_L}
\newcommand{\pseudoRot}{\Rot^*}

\section{Equivariance metrics}
\label{app:metrics}

In this section, we describe our equivariance metrics, \metrict{} and \metricr{}, in detail.
We also present additional results using an alternative translation metric, \metrict{}\textsubscript{frac}, based on fractional sub-pixel translation.

We express each of our metrics as the \emph{peak signal-to-noise ratio} (PSNR) between two sets of images, measured in decibels (dB).
PSNR is a commonly used metric in image restoration literature.
In the typical setting we have two signals, reference $I$ and its noisy approximation $K$, defined over discrete domain $\mathcal{D}$\,---\,usually a two-dimensional pixel grid.
The PSNR between $I$ and $K$ is then defined via the mean squared error (MSE):
\begin{align}
  \text{MSE}_\mathcal{D}(I,K) &= \frac{1}{\Vert\mathcal{D}\Vert} \sum_{i\in\mathcal{D}} \big( I[i] - K[i] \big)^2, \\
	\text{PSNR}_\mathcal{D}(I,K) &= 10 \cdot \log_{10} \left( \frac{I^2_\mathit{max}}{\text{MSE}_\mathcal{D}(I,K)} \right),
\end{align}
where $\text{MSE}_\mathcal{D}(I,K)$ is the average squared difference between matching elements of $I$ and $K$.
$I_\mathit{max}$ is the expected dynamic range of the reference signal, i.e., $I_\mathit{max} \approx \max_{i\in\mathcal{D}}(I[i]) - \min_{i\in\mathcal{D}}(I[i])$.
The dynamic range is usually considered to be a global constant, e.g., the range of valid RGB values, as opposed to being dependent on the content of $I$.
In our case, $I$ and $K$ represent desired and actual outputs of the synthesis network, respectively, with a dynamic range of $[-1,1]$.
This implies that $I_\mathit{max} = 2$.
High PSNR values indicate that $K$ is close to $I$; in the extreme case, where $K = I$, we have $\text{PSNR}_\mathcal{D}(I,K) = \infty$ dB.

Since we are interested in \emph{sets} of images, we use a slightly extended definition for MSE that allows $I$ and $K$ to be defined over an arbitrary, potentially uncountable domain:
\begin{equation}
  \text{MSE}_\mathcal{D}(I,K) = \expectation_{i \sim \mathcal{D}} \left[ \big( I(i) - K(i) \big)^2 \right].
\end{equation}

\subsection{Integer translation}

The goal of our integer translation metric, \metrict{}, is to measure how closely, on average, the output the synthesis network $\Synthesis$ matches a translated reference image when we translate the input of $\Synthesis$.
In other words,
\begin{align}
  \label{eq:app_metrict}
  \begin{array}{l}
    \metrict = \text{PSNR}_{\WW \times \mathcal{X}^2 \times \mathcal{V} \times \mathcal{C}}(I_\trans, K_\trans), \\[2.5mm]
    I_\trans(\ww, \xx, \pp, c) = \Trans_{\xx} \big[ \Synthesis(\feat_0; \ww) \big] [\pp, c], \\[2.5mm]
    K_\trans(\ww, \xx, \pp, c) = \Synthesis(\trans_{\xx}[\feat_0]; \ww)[\pp, c],
  \end{array}
\end{align}
where $\ww \sim \WW$ is a random intermediate latent code produced by the mapping network, $\xx = (x_0, x_1) \sim \mathcal{X}^2$ is a random translation offset, $\pp$ enumerates pixel locations in the mutually valid region $\mathcal{V}$, $c \sim \mathcal{C}$ is the color channel, and $\feat_0$ represents the input Fourier features.
For integer translations, we sample the translation offsets $x_0$ and $x_1$ from $\mathcal{X} = \mathcal{U}[-\sfreq_N/8, \sfreq_N/8]$, where $\sfreq_N$ is the width of the image in pixels.

In practice, we estimate the expectation in Equation~\ref{eq:app_metrict} as an average over 50,000 random samples of $(\ww, \xx) \sim \WW \times \mathcal{X}^2$.
For given $\ww$ and $\xx$, we generate the reference image $I_\trans$ by running the synthesis network and translating the resulting image by $\xx$ pixels (operator $\Trans_{\xx}$).
We then obtain the approximate result image $K_\trans$ by translating the input Fourier features by the corresponding amount (operator $\trans_{\xx}$), as discussed in Appendix~\ref{app:implementation_arch}, and running the synthesis network again.
The mutually valid region of $I_\trans$ (translated by $(x_0, x_1)$) and $K_\trans$ (translated by $(0,0)$) is given by
\begin{equation}
  \begin{array}{l}
    \mathcal{V} =  \{ \max(x_0, 0), \ldots, \sfreq_N + \min(x_0, 0) - 1 \} \times \\[2.5mm]
    \hspace{7.1mm} \{ \max(x_1, 0), \ldots, \sfreq_N + \min(x_1, 0) - 1 \}.
  \end{array}
\end{equation}

\subsection{Fractional translation}
\label{app:metrics_frac}

\figBridgeTablesFrac{fig:BridgeTablesFrac} %

Our translation equivariance metric has the nice property that, for a perfectly equivariant generator, the value of \metrict{} converges to $\infty$ dB when the number of samples tends to infinity.
However, this comes at the cost of completely ignoring subpixel effects.
In fact, it is easy to imagine a generator that is perfectly equivariant to integer translation but fails with subpixel translation; in principle, this is true for any generator whose output is not properly bandlimited, including, e.g., implicit coordinate-based MLPs~\cite{Anokhin2020}.

To verify that our generators \emph{are} able to handle subpixel translation, we define an alternative translation equivariance metric, \metrict{}\textsubscript{frac}, where the translation offsets $x_0$ and $x_1$ are sampled from a continuous distribution $\mathcal{X} = \mathcal{U}(-\sfreq_N/8, \sfreq_N/8)$.
While the continuous operator $\trans_{\xx}$ readily supports this new definition with fractional offsets, extending the discrete $\Trans_{\xx}$ is slightly more tricky.

In practice, we define $\Trans_{\xx}$ via standard Lanczos resampling, by filtering the image produced by $\Synthesis$ using the prototypical Lanczos filter (Equation~\ref{eq:app_lanczos}) with $a = 3$, evaluated at integer tap locations offset by $\xx$.
We explicitly normalize the resulting discretized filter to enforce the partition of unity property.
We also shrink the mutually valid region to account for the spatial extent $a$ by redefining
\begin{equation}
  \begin{array}{l}
    \mathcal{V} =  \{ \max(x_0 + a, 0), \ldots, \sfreq_N + \min(x_0 - a, -1) \} \times \\[2.5mm]
    \hspace{7.1mm} \{ \max(x_1 + a, 0), \ldots, \sfreq_N + \min(x_1 - a, -1) \}.
  \end{array}
\end{equation}
Figure~\ref{fig:BridgeTablesFrac} compares the results of the two metrics, \metrict{} and \metrict{}\textsubscript{frac}, using the same training configurations as Figure~\refpaper{fig:BridgeTables} in the main paper.
The metrics agree reasonably well up until $\sim$40 dB, after which the fractional metric starts to saturate; it consistently fails to rise above 50 dB in our tests.
This is due to the fact that the definition of subpixel translation is inherently ambiguous.
The choice of the resampling filter represents a tradeoff between aliasing, ringing, and retention of high frequencies; there is no reason to assume that the generator would necessarily have to make the same tradeoff as the metric.
Based on the results, we conclude that our configs~\textsc{g--r} are essentially perfectly equivariant to subpixel translation within the limits of Lanczos resampling's accuracy.
However, due to its inherent limitations, we refrain from choosing \metrict{}\textsubscript{frac} as our primary metric.

\subsection{Rotation}

Measuring equivariance with respect to arbitrary rotations has the same fundamental limitation as our \metrict{}\textsubscript{frac} metric: the resampling operation is inherently ambiguous, so we cannot except the results to be perfectly accurate beyond $\sim$40 dB.
Arbitrary rotations also have the additional complication that the bandlimit of a discretely sampled image is not radially symmetric.

Consider rotating the continuous representation of a discretely sampled image by 45$^\circ$.
The original frequency content of the image is constrained within the rectangular bandlimit $\ff \in [-\sfreq_N/2, +\sfreq_N/2]^2$.
The frequency content of the rotated image, however, forms a diamond shape that extends all the way to $\Vert\ff\Vert = \sqrt{2}\sfreq_N/2$ along the main axes but only to $\Vert\ff\Vert = \sfreq_N/2$ along the diagonals.
In other words, it simultaneously has too much frequency content, but also too little.
This has two implications.
First, in order to obtain a valid discretized result image, we have to low-pass filter the image \emph{both} before \emph{and} after the rotation to completely eliminate aliasing.
Second, even if we are successful in eliminating the aliasing, the rotated image will still lack the highest representable diagonal frequencies.
The second point further implies that when computing PSNR, our reference image $I$ will inevitably lack some frequencies that are present in the output of $\Synthesis$.
To obtain the correct result, we must eliminate these extraneous frequencies\,---\,without modifying the output image in any other way.

Based on the above reasoning, we define our \metricr{} metric as follows:
\begin{align}
  \label{eq:app_metricr}
  \begin{array}{l}
    \metricr = \text{PSNR}_{\WW \times \mathcal{A} \times \mathcal{V} \times \mathcal{C}}(I_\rot, K_\rot), \\[2.5mm]
    I_\rot(\ww, \alpha, \pp, c) = \Rot_{\alpha} \big[ \Synthesis(\feat_0; \ww) \big] [\pp, c], \\[2.5mm]
    K_\rot(\ww, \alpha, \pp, c) = \pseudoRot_{\alpha} \big[ \Synthesis(\rot_{\alpha}[\feat_0]; \ww) \big] [\pp, c],
  \end{array}
\end{align}
where the random rotation angle $\alpha$ is drawn from $\mathcal{A} = \mathcal{U}(0^\circ, 360^\circ)$ and operator $\rot_{\alpha}$ corresponds to continuous rotation of the input Fourier features by $\alpha$ with respect to the center of the canvas $[0,1]^2$.
$\Rot_{\alpha}$ corresponds to high-quality rotation of the reference image, and $\pseudoRot_{\alpha}$ represents a \emph{pseudo-rotation} operator that modifies the frequency content of the image \emph{as if} it had undergone $\Rot_{\alpha}$\,---\,but without \emph{actually} rotating it.

The \emph{ideal} rotation operator $\hat{\Rot}$ is easily defined under our theoretical framework presented in Section~\refpaper{sec:theory_layers}:
\begin{equation}
  \hat{\Rot}_{\alpha}[\Feat] = \comb \mult \big( \ideallow \conv \rot_{\alpha}[\idealsq \conv \Feat] \big) = 1/\sfreq^2 \cdot \comb \mult \big( \ideallow \conv \rot_{\alpha}[\ideallow \conv \Feat] \big).
\end{equation}
In other words, we first convolve the discretely sampled input image $\Feat$ with $\idealsq$ to obtain the corresponding continuous representation.
We then rotate this continuous representation using $\rot_{\alpha}$, bandlimit the result by convolving with $\ideallow$, and finally extract the corresponding discrete representation by multiplying with $\comb$. To reduce notational clutter, we omit the subscripts denoting the sampling rate $\sfreq$.
We can swap the order of the rotation and a convolution in the above formula by rotating the kernel in the opposite direction to compensate:
\begin{align}
  \hat{\Rot}_{\alpha}[\Feat] &= 1/\sfreq^2 \cdot \comb \mult \rot_{\alpha} [\hat{h}_R \conv \Feat], &
  \hat{h}_R &= \rot_{-\alpha}[\ideallow] \conv \ideallow,
\end{align}
where $\hat{h}_R$ represents an ideal ``rotation filter'' that bandlimits the signal with respect to both the input and the output. Its spectrum is the eight-sided polygonal intersection of the original and the rotated rectangle.

In order to obtain a practical approximation $\Rot_{\alpha}$, we must replace $\hat{h}_R$ with an approximate filter $h_R$ that has finite support.
Given such a filter, we get $\Rot_{\alpha}[\Feat] = 1/\sfreq^2 \cdot \comb \mult \rot_{\alpha} [h_R \conv \Feat]$.
In practice, we implement this operation using two additional approximations.
First, we approximate $1 / \sfreq^2 \cdot h_R \conv \Feat$ by an upsampling operation to a higher temporary resolution, using $h_R$ as the upsampling filter and $m = 4$.
Second, we approximate $\comb \mult \rot_{\alpha}$ by performing a set of bilinear lookups from the temporary high-resolution image.

To obtain $h_R$, we again utilize the standard Lanczos window with $a = 3$:
\begin{equation}
  h_R = \big( \rot_{-\alpha}[\ideallow] \conv \ideallow \big) \mult (\rot_{-\alpha}[\wlanczossep] \conv \wlanczossep),
\end{equation}
where we apply the same rotation-convolution to both the filter and the window function.
$\wlanczossep$ corresponds the canonical separable Lanczos window, similar to the one used in Equation~\ref{eq:app_lanczos}:
\begin{equation}
  \wlanczossep(\xx) =
  \begin{cases}
    \sinc(x_0 / a) \cdot \sinc(x_1 / a), & \textrm{if } \max(|x_0|, |x_1|) < a, \\
    0,                                   & \textrm{if } \max(|x_0|, |x_1|) \ge a,
  \end{cases}
\end{equation}

We can now define the pseudo-rotation operator $\pseudoRot_{\alpha}[\Feat]$ as a simple convolution with another filter that resembles $h_R$:
\begin{equation}
  \begin{array}{l}
    \pseudoRot_{\alpha}[\Feat] =  1/\sfreq^2 \cdot \comb \mult (h^*_R \conv \Feat) = H^*_R \conv \Feat, \\[2.5mm]
    h^*_R = \big( \ideallow \conv \rot_{\alpha}[\ideallow] \big) \mult (\wlanczossep \conv \rot_{\alpha}[\wlanczossep]),
  \end{array}
\end{equation}
where the discrete version $H^*_R$ is obtained from $h^*_R$ using Equation~\ref{eq:app_discretization}.

Finally, we define the valid region $\mathcal{V}$ the same way as in Appendix~\ref{app:metrics_frac}: the set of pixels for which both filter footprints fall within the bounds of the corresponding original images.

\section{Implementation details}
\label{app:implementation}

We implemented our alias-free generator on top of the official PyTorch implementation of StyleGAN2-ADA, available at {\small\url{https://github.com/NVlabs/stylegan2-ada-pytorch}}.
We kept most of the details unchanged, including
	discriminator architecture~\cite{Karras2019},
	weight demodulation~\cite{Karras2019},
	equalized learning rate for all trainable parameters~\cite{Karras2017},
	minibatch standard deviation layer at the end of the discriminator~\cite{Karras2017},
	exponential moving average of generator weights~\cite{Karras2017},
  mixed-precision FP16/FP32 training~\cite{Karras2020},
	non-saturating logistic loss~\cite{Goodfellow2014},
  $R_1$ regularization~\cite{Mescheder2018},
	lazy regularization~\cite{Karras2019},
	and Adam optimizer~\cite{adam} with $\beta_1=0$, $\beta_2=0.99$, and $\epsilon=10^{-8}$.

We ran all experiments on NVIDIA DGX-1 with 8 Tesla V100 GPUs using PyTorch 1.7.1, CUDA 11.0, and cuDNN 8.0.5.
We computed FID between 50k generated images and all training images using the official pre-trained Inception network, available at {\small\url{http://download.tensorflow.org/models/image/imagenet/inception-2015-12-05.tgz}}

\FINAL{Our implementation and pre-trained models are available at \codepage}

\newcommand{\T}{\boldsymbol{t}}

\subsection{Generator architecture}
\label{app:implementation_arch}

\paragraph{Normalization (configs~\textsc{d--r})}
We have observed that eliminating the output skip connections in StyleGAN2~\cite{Karras2019} results in uncontrolled drift of signal magnitudes over the generator layers.
This does not necessarily lead to lower-quality results, but it generally increases the amount of random variation between training runs and may occasionally lead to numerical issues with mixed-precision training.
We eliminate the drift by tracking a long-term exponential moving average of the input signal magnitude on each layer and normalizing the feature maps accordingly.
We update the moving average once per training iteration, based on the mean of squares over the entire input tensor, and freeze its value after training.
We initialize the moving average to 1 and decay it at a constant rate, resulting in 50\% decay per 20k real images shown to the discriminator.
With this explicit normalization in place, we have found it beneficial to slightly adjust the dynamic range of the output RGB colors.
StyleGAN2 uses $-1$ and $+1$ to represent black and white, respectively; we change these values to $-4$ and $+4$ starting from config~\textsc{d} and, for consistency with the original generator, divide the color channels by 4 afterwards.

\paragraph{Transformed Fourier features (configs~\textsc{h--r})}
We enable the orientation of the input features $\feat_0$ to vary on a per-image basis by introducing an additional affine layer (Figure~\refpaper{fig:practice}b) and applying a geometric transformation based on its output.
The affine layer produces a four-dimensional vector $\T = (r_c, r_s, t_x, t_y)$ based on $\ww$.
We initialize its weights so that $\T = (1, 0, 0, 0)$ at the beginning, but allow them to change freely over the course of training.
To interpret $\T$ as a geometric transformation, we first normalize its value based on the first two components, i.e., $\T' = (r_c', r_s', t_x', t_y') = \T \big/ \sqrt{r_c^2 + r_s^2}$.
This makes the transformation independent of the magnitude of $\ww$, similar to the weight modulation and demodulation~\cite{Karras2019} on the other layers.
We then interpret the first two components as rotation around the center of the canvas $[0,1]^2$, with the rotation angle $\alpha$ defined by $r_c' = \cos \alpha$ and $r_s' = \sin \alpha$.
Finally, we interpret the remaining two components as translation by $(t_x', t_y')$ units, so that the translation is performed after the rotation.
In practice, we implement the resulting geometric transformation by modifying the phases and two-dimensional frequencies of the Fourier features, which is equivalent to applying the same transformation to the continuous representation of $\feat_0$ analytically.

\paragraph{Flexible layer specifications}
In configs~\textsc{t} and~\textsc{r}, we define the per-layer filter parameters (Figure~\refpaper{fig:practice}c) as follows.
The cutoff frequency $\freqc$ and the minimum acceptable stopband frequency $\freqt$ obey geometric progression until the first critically sampled layer:
\begin{align}
  \freqc[i] &= \freq_{c,0} \cdot (\freq_{c,N} / \freq_{c,0}) ^ {\min(i / (N - N_\text{crit}), 1)}, &
  \freqt[i] &= \freq_{t,0} \cdot (\freq_{t,N} / \freq_{t,0}) ^ {\min(i / (N - N_\text{crit}), 1)},
\end{align}
where $N = 14$ is the total number of layers, $N_\text{crit} = 2$ is the number of critically sampled layers at the end, $\freq_{c,0} = 2$ corresponds to the frequency content of the input Fourier features, and $\freq_{c,N} = \sfreq_N / 2$ is defined by the output resolution.
$\freq_{t,0}$ and $\freq_{t,N}$ are free parameters; we use $\freq_{t,0} = 2^{2.1}$ and $\freq_{t,N} = \freq_{c,N} \cdot 2^{0.3}$ in most of our tests.
Given the values of $\freqc[i]$ and $\freqt[i]$, the sampling rate $\sfreq[i]$ and transition band half-width $\freqh[i]$ are then determined by
\begin{align}
  \sfreq[i] &= \exp_2 \left\lceil \log_2 \big(\min(2 \cdot \freqt[i], \sfreq_N) \big) \right\rceil, &
  \freqh[i] &= \max(\freqt[i], \sfreq[i] / 2) - \freqc[i].
\end{align}
The sampling rate is rounded up to the nearest power of two that satisfies $\sfreq[i] \ge 2 \freqt[i]$, but it is not allowed to exceed the output resolution.
The transition band half-width is selected to satisfy either $\freqc[i] + \freqh[i] = \freqt[i]$ or $\freqc[i] + \freqh[i] = \sfreq[i] / 2$, whichever yields a higher value.

We consider $\freqc[i]$ to represent the output frequency content of layer $i$, for $i \in \{0, 1, \ldots, N-1\}$, whereas the input is represented by $\freqc[\max(i - 1, 0)]$.
Thus, we construct the corresponding upsampling filter according to $\freqc[\max(i - 1, 0)]$ and $\freqh[\max(i - 1, 0)]$ and the downsampling filter according to $\freqc[i]$ and $\freqh[i]$.
The nonlinearity is evaluated at a temporary sampling rate $\sfrequp = \max(\sfreq[i], \sfreq[\max(i - 1, 0)]) \cdot m$, where $m$ is the upsampling parameter discussed in Section~\refpaper{sec:practice_additions} that we set to 2 in most of our tests.

\subsection{Hyperparameters and training configurations}

\figHyperparams{fig:Hyperparams} %
We used 8 GPUs for all our training runs and continued the training until the discriminator had seen a total of 25M real images when training from scratch, or 5M images when using transfer learning.
Figure~\ref{fig:Hyperparams} shows the hyperparameters used in each experiment.
\FINAL{We performed the baseline runs (configs~\textsc{a--c}) using the corresponding standard configurations:
StyleGAN2 config~F~\cite{Karras2019} for the high-resolution datasets in Figure~\refpaper{fig:ResultTables}, left,
and ADA 256$\times$256 baseline config~\cite{Karras2020} for the ablations in Figure~\refpaper{fig:BridgeTables} and Figure~\refpaper{fig:ResultTables}, right.}

Many of our hyperparameters, including discriminator capacity and learning rate, batch size, and generator moving average decay, are inherited directly from the baseline configurations, and kept unchanged in all experiments.
In configs~\textsc{c} and~\textsc{d}, we disable noise inputs~\cite{Karras2018}, path length regularization~\cite{Karras2019}, and mixing regularization~\cite{Karras2018}.
In config~\textsc{d}, we also decrease the mapping network depth to 2 and set the minibatch standard deviation group size to 4 as recommended in the StyleGAN2-ADA documentation.
The introduction of explicit normalization in config~\textsc{d} allows us to use \FINAL{the same generator learning rate, 0.0025, for all output resolutions.}
In Figure~\refpaper{fig:ResultTables}, right, we show results for path length regularization with weight 0.5 and mixing regularization with probability 0.5.

\paragraph{Augmentation}
Since our datasets are horizontally symmetric in nature, we enable dataset $x$-flip augmentation in all our experiments.
To prevent the discriminator from overfitting, we enable adaptive discriminator augmentation (ADA)~\cite{Karras2020} with default settings for \textsc{MetFaces}, \textsc{MetFaces-U}, \textsc{AFHQv2}, and \textsc{Beaches}, but disable it for \textsc{FFHQ} and \textsc{FFHQ-U}.
Furthermore, we train \textsc{MetFaces} and \textsc{MetFaces-U} using transfer learning from the corresponding \textsc{FFHQ} or \textsc{FFHQ-U} snapshot with the lowest FID, similar to Karras~et~al.~\cite{Karras2020}, but start the training from scratch in all other experiments.

\paragraph{Generator capacity}
StyleGAN2 defines the number of feature maps on a given layer to be inversely proportional to its resolution, i.e., $C[i] = C(\sfreq[i]) = \min(\round(C_\text{base} / \sfreq[i]), C_\text{max})$, where $\sfreq[i]$ is the output resolution of layer $i$.
Parameters $C_\text{base}$ and $C_\text{max}$ control the overall capacity of the generator; our baseline configurations use $C_\text{max} = 512$ and $C_\text{base} = 2^{14}$ or $2^{15}$ depending on the output resolution.
Since StyleGAN2 can be considered to employ critical sampling on all layers, i.e., $\freqc[i] = \sfreq[i] / 2$, we can equally well define the number of feature maps as $C[i] = C(2 \freqc[i])$.
These two definitions are equivalent for configs~\textsc{a--f}, but in configs~\textsc{g--r} we explicitly set $\freqc[i] \le \sfreq[i] / 2$, which necessitates using the latter definition.
In config~\textsc{r}, we double the value of both $C_\text{base}$ and $C_\text{max}$ to compensate for the reduced capacity of the 1$\times$1 convolutions.
In Figure~\refpaper{fig:ResultTables}, right, we sweep the capacity by multiplying both parameters by 0.5, 1.0, and 2.0.

\paragraph{\emph{R}\textsubscript{1} regularization}
The optimal choice for the $R_1$ regularization weight $\gamma$ is highly dependent on the dataset, necessitating a grid search~\cite{Karras2019,Karras2020}.
For the baseline config~\textsc{b}, we tested $\gamma \in \{1, 2, 5, 10, 20\}$ and selected the value that gave the best FID for each dataset.
For our configs~\textsc{t} and~\textsc{r}, we followed the recommendation of Karras~et~al.~\cite{Karras2020} to define $\gamma = \gamma_0 \cdot N / M$, where $N = \sfreq_N^2$ is the number of output pixels and $M$ is the batch size, and performed a grid search over $\gamma_0 \in \{0.0002, 0.0005, 0.0010, 0.0020, 0.0050\}$.
For the low-resolution ablations, we chose to use a fixed value $\gamma = 1$ for simplicity.
The resulting values of $\gamma$ are shown in Figure~\ref{fig:Hyperparams}, right.

\FINAL{
\paragraph{Training of config \textsc{r}}
In this configuration, we blur all images the discriminator sees in the beginning of the training. This Gaussian blur is executed just before the ADA augmentation. We start with $\sigma=10$ pixels, which we ramp to zero over the first 200k images. This prevents the discriminator from focusing too heavily on high frequencies early on. It seems that in this configuration the generator sometimes learns to produce high frequencies with a small delay, allowing the discriminator to trivially tell training data from the generated images without providing useful feedback to the generator. As such, config~\textsc{r} is prone to random training failures in the beginning of the training without this trick. The other configurations do not have this issue.}

\subsection{G-CNN comparison}

In Figure~\refpaper{fig:ResultTables}, bottom, we compare our config~\textsc{r} with config~\textsc{t} extended with $p4$-symmetric group convolutions~\cite{Cohen2016,Dey2021}.
$p4$ symmetry makes the generator equivariant to 0$^\circ$, 90$^\circ$, 180$^\circ$, and 270$^\circ$ rotations, but not to arbitrary rotation angles.
In practice, we implement the group convolutions by extending all intermediate activation tensors in the synthesis network with an additional group dimension of size 4 and introducing appropriate redundancy in the convolution weights.
We keep the input layer unchanged and introduce the group dimension by replicating each element of $z_0$ four times.
Similarly, we eliminate the group dimension after the last layer by computing an average of the four elements.
$p4$-symmetric group convolutions have 4$\times$ as many trainable parameters as the corresponding regular convolutions.
To enable an apples-to-apples comparison, we compensate for this increase by halving the values of $C_\text{base}$ and $C_\text{max}$, which brings the number of parameters back to the original level.

\section{Energy consumption}
\label{app:energy}

\figEnergy{fig:Energy} %

Computation is an essential resource in machine learning projects: its availability and cost, as well as the associated energy consumption, are key factors in both choosing research directions and practical adoption. We provide a detailed breakdown for our entire project in Table~\ref{fig:Energy} in terms of both GPU time and electricity consumption.
We report expended computational effort as single-GPU years (Volta class GPU). We used a varying number of NVIDIA DGX-1s for different stages of the project, and converted each run to single-GPU equivalents by simply scaling by the number of GPUs used.

We followed the Green500 power measurements guidelines \cite{Ge2020}. 
The entire project consumed approximately 225 megawatt hours (MWh) of electricity. 
Approximately 70\% of it was used for exploratory runs, where we gradually built the new configurations; first in an unstructured manner and then specifically ironing out the new \FINAL{StyleGAN3-T and StyleGAN3-R} configurations. 
Setting up the intermediate configurations between StyleGAN2 and our generators, as well as, the key parameter ablations was also quite expensive at $\sim$15\%.
Training a single instance of \FINAL{StyleGAN3-R} at 1024$\times$1024 is only slightly more expensive (0.9MWh) than training StyleGAN2 (0.7MWh)~\cite{Karras2019}.

\fi

\end{document}